\documentclass[11pt,twocolumn]{article}

\usepackage[margin=1in]{geometry}
\usepackage[T1]{fontenc}
\usepackage[utf8]{inputenc}
\usepackage{lmodern}

\usepackage[authoryear,longnamesfirst]{natbib}

\usepackage{amsmath}
\usepackage{amssymb}

\usepackage{graphicx}
\usepackage{tabularx}
\usepackage{array}
\usepackage{booktabs}
\usepackage{float}
\usepackage{placeins}

\usepackage{listings}
\usepackage[most,breakable]{tcolorbox}
\usepackage{xcolor}

\usepackage{algorithm}
\usepackage{algpseudocode}

\usepackage[colorlinks=true,
            linkcolor=blue!70!black,
            citecolor=blue!70!black,
            urlcolor=blue!70!black,
            breaklinks=true]{hyperref}

\usepackage{fancyhdr}
\usepackage{abstract}

\tcbuselibrary{skins,breakable}
\tcbset{
  promptbox/.style={
    colback=white, colframe=black,
    boxrule=0.8pt, arc=3pt,
    left=6pt, right=6pt, top=6pt, bottom=6pt
  },
  systembox/.style={
    colback=black, colframe=black, coltext=white,
    boxrule=0.8pt, arc=3pt
  },
  userbox/.style={
    colback=blue!5, colframe=blue!60!black,
    boxrule=0.8pt, arc=3pt
  }
}

\begin{document}
\def\floatpagefraction{1}
\def\textfraction{.001}
\twocolumn[{%
\begin{@twocolumnfalse}

\begin{center}
{\LARGE\bfseries
Empirical Evidence of Complexity-Induced Limits in\\
Large Language Models on Finite Discrete State-Space\\
Problems with Explicit Validity Constraints\par}

\vspace{1.2em}

{\large
Md.\ Fahad Ullah Utsho$^{1}$ \quad
Mohd.\ Ruhul Ameen$^{2}$ \quad
Akif Islam$^{3}$ \quad
Md.\ Golam Rashed$^{1}$ \quad
Dipankar Das$^{1}$ \par}

\vspace{0.8em}

{\small
$^{1}$Department of Information and Communication Engineering, University of Rajshahi, Bangladesh\par
\texttt{fahadullahutsho@gmail.com}, \texttt{golamrashed@ru.ac.bd}, \texttt{dipankar@ru.ac.bd}\par
\vspace{0.4em}
$^{2}$College of Engineering and Computer Sciences, Marshall University, Huntington, WV, USA\par
\texttt{ameen@marshall.edu}\par
\vspace{0.4em}
$^{3}$Department of Computer Science and Engineering, University of Rajshahi, Bangladesh\par
\texttt{iamakifislam@gmail.com}\par
}

\end{center}

\vspace{1em}

\begin{abstract}
\begin{abstract}
Large Language Models (LLMs) are increasingly described as possessing strong reasoning capabilities, supported by high performance on mathematical, logical, and planning benchmarks. However, most existing evaluations rely on aggregate accuracy over fixed datasets, obscuring how reasoning behavior evolves as task complexity increases. In this work, we introduce a controlled benchmarking framework to systematically evaluate the robustness of reasoning in Large Reasoning Models (LRMs) under progressively increasing problem complexity. We construct a suite of nine classical reasoning tasks—Boolean Satisfiability, Cryptarithmetic, Graph Coloring, River Crossing, Tower of Hanoi, Water Jug, Checker Jumping, Sudoku, and Rubik’s Cube—each parameterized to enable precise control over complexity while preserving underlying semantics. Using deterministic validators, we evaluate multiple open and proprietary LRMs across low-, intermediate-, and high-complexity regimes, ensuring that only fully valid solutions are accepted. Our results reveal a consistent phase-transition-like behavior: models achieve high accuracy at low complexity but experience a rapid and often abrupt degradation beyond task-specific complexity thresholds. We formalize this phenomenon as \emph{reasoning collapse}, defined as a sharp drop in success probability with respect to increasing complexity. Across tasks, we observe substantial accuracy declines (often exceeding 50\%) beyond these thresholds, accompanied by failure modes such as inconsistent reasoning traces, violation of constraints, loss of state tracking, and confidently incorrect outputs. Importantly, increased reasoning length does not reliably improve correctness, and performance gains in one problem family do not generalize to others. These findings suggest that contemporary LRMs rely on brittle, heuristic reasoning strategies that fail to scale with problem complexity. Our work highlights the need for evaluation methodologies that move beyond static benchmarks and explicitly measure reasoning robustness under controlled complexity, providing a more reliable assessment of true reasoning capability.
\end{abstract}
\end{abstract}

\vspace{1em}
\end{@twocolumnfalse}
}]


\section{Introduction}
Over the last few years, large language models (LLMs) have evolved from primarily \emph{next-token predictors} into systems capable of solving multi-step problems, generating programs, and producing natural-language justifications. A particularly influential line of work demonstrated that explicit \emph{chain-of-thought} (CoT) prompting can elicit intermediate reasoning steps and substantially improve performance on tasks requiring compositional and arithmetic structure \cite{wei2022chainofthought}. Subsequent methods, including zero-shot CoT and self-consistency, further enhanced performance by encouraging models to generate and aggregate multiple reasoning trajectories \cite{kojima2022zeroshot,wanga2022selfconsistency}. In parallel, the community has developed large-scale evaluation suites and specialized reasoning benchmarks—spanning multi-task challenge collections and mathematical problem sets—to track progress and compare models \cite{srivastava2022bigbench,cobbe2021gsm8k,hendrycks2021math}.

Despite these advances, a fundamental limitation remains: most existing evaluations rely on aggregate accuracy over fixed datasets, which obscures how reasoning behavior evolves as task complexity increases. In particular, such benchmarks provide limited insight into whether models perform \emph{algorithmic reasoning}—maintaining consistent intermediate states under constraints—or rely on surface-level pattern matching learned from training data. Moreover, benchmark-based evaluation is susceptible to data contamination and memorization effects, which can inflate apparent reasoning ability when test instances (or close variants) are present in training corpora \cite{oren2023contamination}.

Recent developments have introduced \emph{Large Reasoning Models} (LRMs), which generate extended “thinking” traces prior to producing final answers. While these traces create a compelling narrative of deliberation—suggesting exploration, hypothesis revision, and convergence—they also introduce a critical evaluation blind spot. Current metrics typically assess only final correctness, ignoring whether intermediate reasoning steps are internally consistent, constraint-valid, or algorithmically meaningful. As a result, models can produce fluent but invalid reasoning trajectories that nonetheless appear convincing.

A growing body of empirical evidence suggests that such reasoning behavior does not scale uniformly with problem difficulty. Notably, recent work (e.g., Apple’s \emph{The Illusion of Thinking}) demonstrates that LRMs exhibit abrupt performance degradation beyond task-specific complexity thresholds, often accompanied by reduced reasoning effort despite ample token budgets \cite{shojaee2025illusion}. These findings challenge a widely held assumption: that increasing inference-time computation (e.g., longer reasoning traces) necessarily leads to improved reasoning capability.

In this work, we formalize and systematically investigate this phenomenon by studying reasoning as a function of \emph{controlled problem complexity}. We define reasoning as the ability to maintain \emph{state-consistent, constraint-valid transformations} over a sequence of steps, and introduce the notion of \emph{reasoning collapse}: a sharp degradation in success probability beyond a complexity threshold. To probe this behavior, we construct a suite of parametrized puzzle environments in which complexity can be increased while preserving underlying semantics. This enables precise measurement of when reasoning fails, how failure manifests in reasoning traces, and whether capabilities generalize across problem domains.

Our central hypothesis is as follows: if LRMs perform genuine algorithmic reasoning, performance should degrade gradually and maintain structural consistency as complexity increases; if they rely on heuristic pattern matching, performance should exhibit abrupt transitions, inconsistent intermediate states, and poor cross-task generalization. By evaluating models under controlled complexity scaling, we aim to distinguish between these regimes and provide a more robust characterization of reasoning in modern LLMs.

\subsection{Contributions}
We make the following contributions:
\begin{sloppypar}
\begin{enumerate}
    \item \textbf{Complexity-collapse profiling for LRMs:} We empirically estimate per-model reasoning collapse thresholds across multiple parametrized puzzle environments, enabling fine-grained analysis beyond aggregate accuracy.
    
    \item \textbf{Quantitative analysis of reasoning degradation:} We characterize how reasoning traces degrade with increasing complexity, identifying failure modes such as inconsistency, premature termination, and confidently incorrect trajectories.
    
    \item \textbf{Regime-aware evaluation of reasoning performance:} We analyze model behavior across low-, intermediate-, and high-complexity regimes to determine when extended reasoning improves performance and when it fails, extending recent observations on reasoning limits \cite{shojaee2025illusion}.
    
    \item \textbf{Cross-domain generalization analysis:} We evaluate whether performance improvements in one problem family transfer to structurally distinct tasks, providing evidence on the extent of genuine reasoning versus task-specific heuristics.
    
    \item \textbf{Reproducible and deterministic evaluation framework:} We introduce a benchmark suite with explicit complexity control and deterministic validators, ensuring that only fully valid solutions are accepted and enabling precise measurement of reasoning robustness.
\end{enumerate}
\end{sloppypar}

\subsection{Formal Definitions and Problem Setting}

To enable precise analysis, we formalize the notion of reasoning and its failure under increasing complexity.

\paragraph{Reasoning.}
We define reasoning as the ability of a model to produce a sequence of \emph{state-consistent and constraint-valid transformations} that lead from an initial state to a goal state. Formally, given a problem instance characterized by a state space $\mathcal{S}$, action space $\mathcal{A}$, transition function $\mathcal{T}$, and validity predicate $\mathcal{V}$, a reasoning trajectory $\pi = (a_1, \dots, a_T)$ is valid if:
\[
\mathcal{V}(s_t, a_t) = 1 \quad \forall t \in \{1, \dots, T\}, \quad \text{and} \quad s_T \in \mathcal{G},
\]
where $s_{t+1} = \mathcal{T}(s_t, a_t)$ and $\mathcal{G}$ denotes the goal set. Under this definition, correct reasoning requires maintaining consistency across all intermediate states, not merely producing a correct final answer.

\paragraph{Problem Complexity.}
We define problem complexity as a parameterized quantity $\lambda$ that controls structural difficulty while preserving task semantics. Depending on the task, $\lambda$ may correspond to recursion depth (e.g., number of disks in Tower of Hanoi), constraint density (e.g., clause-to-variable ratio in SAT), or planning horizon (e.g., scramble length in Rubik’s Cube). Increasing $\lambda$ induces growth in state space size, solution length, or constraint interactions.

\paragraph{Reasoning Collapse.}
We define \emph{reasoning collapse} as a sharp degradation in model performance as a function of increasing complexity. Let $\text{Acc}(\lambda)$ denote the success rate at complexity level $\lambda$. We characterize the collapse threshold $\lambda^*$ as:
\[
\lambda^* = \arg\max_{\lambda} \left| \frac{d \, \text{Acc}(\lambda)}{d\lambda} \right|,
\]
i.e., the point at which performance exhibits the steepest decline. Empirically, this corresponds to a phase-transition-like behavior where models shift from high accuracy to near-random or systematically invalid outputs.

\paragraph{Failure Modes.}
Beyond aggregate accuracy, we characterize reasoning failure through violations of intermediate validity. Common failure modes include:
\begin{itemize}
    \item \textbf{State inconsistency:} intermediate states contradict prior steps,
    \item \textbf{Constraint violation:} outputs violate task-specific rules,
    \item \textbf{Incomplete reasoning:} trajectories terminate prematurely,
    \item \textbf{Confident errors:} fluent but invalid solutions.
\end{itemize}

\paragraph{Hypothesis.}
Our central hypothesis is that if LRMs perform genuine algorithmic reasoning, performance should degrade gradually and maintain structural consistency as $\lambda$ increases. Conversely, if reasoning is driven by heuristic pattern matching, performance should exhibit abrupt collapse, inconsistent intermediate states, and poor generalization across tasks.

This formalization enables us to move beyond static benchmark accuracy and evaluate reasoning as a function of controlled complexity.

\section{Related Work}

A substantial body of research has focused on improving and evaluating reasoning capabilities in large language models (LLMs). One of the most influential directions is \emph{Chain-of-Thought} (CoT) prompting, which encourages models to generate intermediate reasoning steps prior to producing a final answer. \citet{wei2022chainofthought} showed that CoT can substantially improve multi-step problem solving across arithmetic, commonsense, and symbolic reasoning tasks. Subsequent extensions such as zero-shot CoT and self-consistency further enhanced performance by eliciting or aggregating multiple reasoning trajectories \citep{kojima2022zeroshot,wang2022selfconsistency}. While these methods yield empirical gains, they do not, by themselves, establish whether models execute robust algorithmic reasoning or instead rely on heuristic pattern completion shaped by training distributions.

Beyond prompting strategies, the community has developed a wide range of benchmarks for assessing reasoning performance. Large-scale suites such as BIG-Bench, GSM8K, and MATH provide standardized datasets spanning diverse reasoning domains \citep{srivastava2022bigbench,cobbe2021gsm8k,hendrycks2021math}. However, these benchmarks are primarily \emph{static}: difficulty is largely determined by dataset construction and latent distributional properties rather than by explicit, controllable parameters. As a result, they offer limited insight into how reasoning behavior evolves as task complexity increases. In addition, benchmark outcomes can be confounded by data contamination, memorization, or distributional overlap with pretraining corpora, complicating interpretation of observed performance improvements \citep{oren2023contamination,ni2025surveybenchmarks}.

A parallel line of work has examined the validity of current evaluation methodologies and argued that accuracy-centric reporting is insufficient to capture reasoning quality. Survey and review papers increasingly advocate for complementary dimensions such as coherence, consistency, and structural validity of reasoning traces \citep{chang2023surveyEvalLLM,bean2025constructValidity}. These critiques emphasize that models may produce fluent, persuasive explanations while still making logical errors or violating task constraints, motivating evaluations that probe reasoning processes rather than only final answers.

To mitigate these issues, recent benchmarks and protocols have begun incorporating more dynamic or fine-grained evaluation designs. ReasonBENCH introduces multi-run evaluation to quantify instability and reproducibility of reasoning behavior across decoding runs \citep{reasonbench2025}. DyVal proposes dynamically generated evaluation sets with varying difficulty levels to study performance under changing task demands \citep{zhu2024dyval}. Similarly, frameworks such as LogicGame and related environments emphasize structured execution and rule-following in controlled simulated problem spaces \citep{gui2025logicgame}. While these efforts move beyond static test sets, they often focus on one dimension at a time (e.g., variance, correctness, or execution fidelity) and do not provide a unified framework for characterizing how reasoning traces degrade across systematically controlled complexity regimes.

In parallel, \emph{LLM-as-a-Judge} approaches have gained traction, using models to evaluate the reasoning quality or correctness of other model outputs \citep{saha2025evalplanner}. Although these methods can scale evaluation when deterministic checking is unavailable, they introduce additional uncertainty because judgments depend on the reliability and calibration of the evaluator model. Moreover, judge-based approaches do not directly isolate complexity-driven failure dynamics when correctness can be determined objectively.

Despite rapid progress, a fundamental gap remains: much of the literature still treats reasoning as a static capability summarized by aggregate accuracy, rather than as a process that may fail in predictable ways as cognitive load increases. In particular, there is limited work that (i) \emph{explicitly parameterizes} problem complexity, (ii) measures performance as a \emph{continuous function} of that complexity, and (iii) analyzes the \emph{failure dynamics} of reasoning traces under increasing constraint pressure and planning depth \citep{reasonbench2025,zhu2024dyval,gui2025logicgame,oren2023contamination,ni2025surveybenchmarks}.

Our work addresses this gap by introducing a controlled-complexity evaluation framework based on parametrized puzzle environments with deterministic validation. Unlike prior benchmarks that rely on fixed datasets or only partially controlled difficulty, our approach enables precise measurement of reasoning behavior as complexity increases, allowing us to estimate reasoning-collapse thresholds, characterize trace-level failure modes, and evaluate cross-task generalization under strict validity constraints. In doing so, we move beyond static accuracy reporting toward a more principled evaluation of reasoning robustness.
\section{Benchmark Tasks and Complexity Design}

Recent advancements in Large Reasoning Models (LRMs) have introduced an important interpretive challenge: when models achieve high accuracy on complex benchmarks, it remains unclear whether such performance reflects genuine reasoning ability, increased exposure to benchmark-adjacent training data, or the availability of larger inference budgets that enable extended reasoning traces. Existing evaluations largely rely on static benchmark datasets and aggregate accuracy metrics, making it difficult to disentangle these confounding factors or to characterize how reasoning behavior evolves as task demands increase.

Several recent studies have attempted to mitigate these issues by comparing ``thinking-style’’ models with their non-deliberative counterparts under controlled inference budgets, typically using mathematical datasets such as GSM8K, MATH, or AIME-style problems. While informative, these analyses are constrained by two limitations. First, mathematical benchmarks offer little control over fine-grained problem complexity: difficulty is inherently dataset-specific and tied to the original problem distribution rather than defined through explicit structural parameters. Second, contamination and repeated exposure across model generations complicate interpretation, as improvements on older datasets may reflect memorization or indirect pretraining signal rather than progress in reasoning capability.

To address these limitations, we adopt a controlled complexity paradigm inspired by recent work on puzzle-like reasoning environments. Instead of treating reasoning ability as a monolithic attribute, we evaluate LRMs using task families whose underlying semantics remain fixed while complexity can be systematically and transparently varied. This design enables us to probe reasoning robustness along well-defined dimensions, identify the onset and progression of collapse phenomena, and differentiate genuine algorithmic reasoning from brittle heuristic or pattern-based strategies.

\begin{table*}[!htbp]
\caption{Context window size and total parameter count of evaluated models.}
\label{tab:model_specs}
\renewcommand{\arraystretch}{1.1}
\begin{tabular*}{\textwidth}{@{\extracolsep{\fill}} lcc @{}}
\toprule
\textbf{Model} & \textbf{Context Length (Tokens)} & \textbf{Parameters} \\
\midrule
Qwen Plus 0728 (Thinking) \cite{alibaba2025qwen} & 1{,}000{,}000 & 3.6B \\
Kimi K2 (Thinking) \cite{moonshot2025kimi} & 128{,}000 & 1T (32B active) \\
Claude 3.7 Sonnet (Thinking) \cite{anthropic2025claude37} & 200{,}000 & approx. 175B \\
Gemini 3 Pro Preview \cite{google2026gemini} & 1{,}048{,}576 & >1T \\
GPT-5 Pro \cite{openai2026gpt5} & 400{,}000 & approx. 1.8T \\
DeepSeek V3.2 \cite{deepseek2025v32} & 163{,}840 & 671B (37B active) \\
\bottomrule
\end{tabular*}
\end{table*}

\subsection{Defining the Problem Families}

We evaluate LRM reasoning using nine controllable puzzle environments spanning recursion, sequential arithmetic, constraint satisfaction, combinatorial search, and spatial planning. Each puzzle admits a clear definition of valid states, legal transitions, and goal conditions, enabling unambiguous verification of correctness. Crucially, the difficulty of each puzzle can be adjusted through explicit parameters, allowing systematic control of problem complexity while preserving the underlying task semantics.

\subsubsection{Tower of Hanoi}

The Tower of Hanoi is a classical recursive planning task defined over a finite discrete state space with explicit move-validity constraints. Its fixed transition rules and exponentially increasing optimal solution length make it an ideal testbed for evaluating long-horizon reasoning robustness.

\begin{figure} 
  \centering
  \includegraphics[width=0.9\columnwidth]{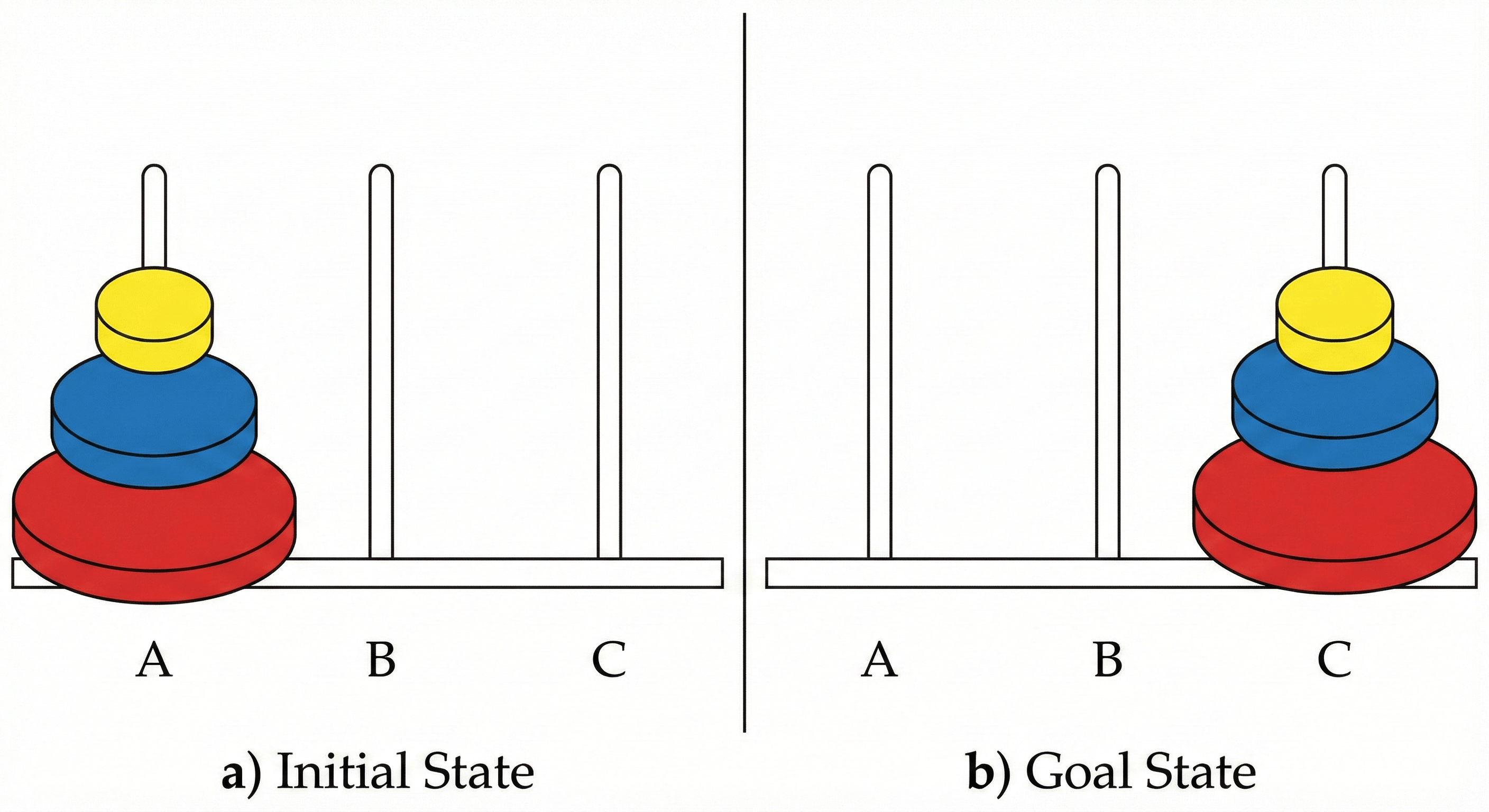}
  \caption{Tower of Hanoi setup with three pegs ($A,B,C$) and $n$ disks; only the top disk may be moved, and no larger disk may be placed on a smaller disk.}
  \label{fig:tower_of_hanoi}
\end{figure}

\paragraph{Formal Structure.}
For a complexity parameter $n \in \mathbb{N}$, we define:
\[
\mathcal{H}_n = (S, A, T, V, G)
\]
where $S$ is the state space, $A$ the action set, $T$ the transition function, $V$ the validity predicate, and $G$ the goal set.

\paragraph{State Space.}
Three pegs $\{A,B,C\}$ hold disks $D=\{1,\dots,n\}$ (1 = smallest).  
A state $s\in S$ is:
\[
s = (P_A, P_B, P_C)
\]
with each $P_i$ an ordered stack of disks respecting size order.  
Ignoring ordering legality, the combinatorial state space size is $|S|=3^n$.

\paragraph{Actions and Transitions.}
Actions move the top disk of one peg to another:
\[
A = \{\text{Move}(d, p_i \rightarrow p_j)\mid p_i\neq p_j\}.
\]
A transition $T(s_t,a_t)$ is defined iff:
\begin{enumerate}
    \item $d$ is the top disk on $p_i$,
    \item $p_j$ is empty or its top disk is larger than $d$.
\end{enumerate}

\paragraph{Validity.}
A move is valid when ordering constraints are respected:
\[
V(s_t,a_t) =
\begin{cases}
1 & \text{valid}\\
0 & \text{invalid}.
\end{cases}
\]
A trajectory $\pi=(a_1,\dots,a_T)$ is valid if $V(s_t,a_t)=1$ for all $t$.

\paragraph{Goal.}
The target configuration is all disks stacked on peg $C$ in correct order:
\[
G = \{(\emptyset,\emptyset,[n,n\!-\!1,\dots,1])\}.
\]

\paragraph{Complexity Parameterization.}
The complexity is directly controlled by:
\[
\lambda = n.
\]
The optimal solution length grows exponentially:
\[
L_{\min}(n)=2^n - 1.
\]
Thus, increasing $n$ extends the required planning horizon without altering the task’s structural rules.

\paragraph{Implications for Reasoning Evaluation.}
Because each additional disk doubles the optimal trajectory length, local reasoning errors compound rapidly. If each step is correct with probability $p$, the success probability scales as:
\[
P_{\text{success}} \approx p^{L_{\min}(n)},
\]
leading to sharp collapse once the required horizon exceeds the model’s state-tracking capacity.  
Tower of Hanoi therefore provides a clean axis for probing long-range dependency maintenance, recursive structure reproduction, and the emergence of complexity-induced reasoning failure.

\subsubsection{Checker Jumping}

\begin{figure}
  \centering
  \includegraphics[width=0.9\columnwidth]{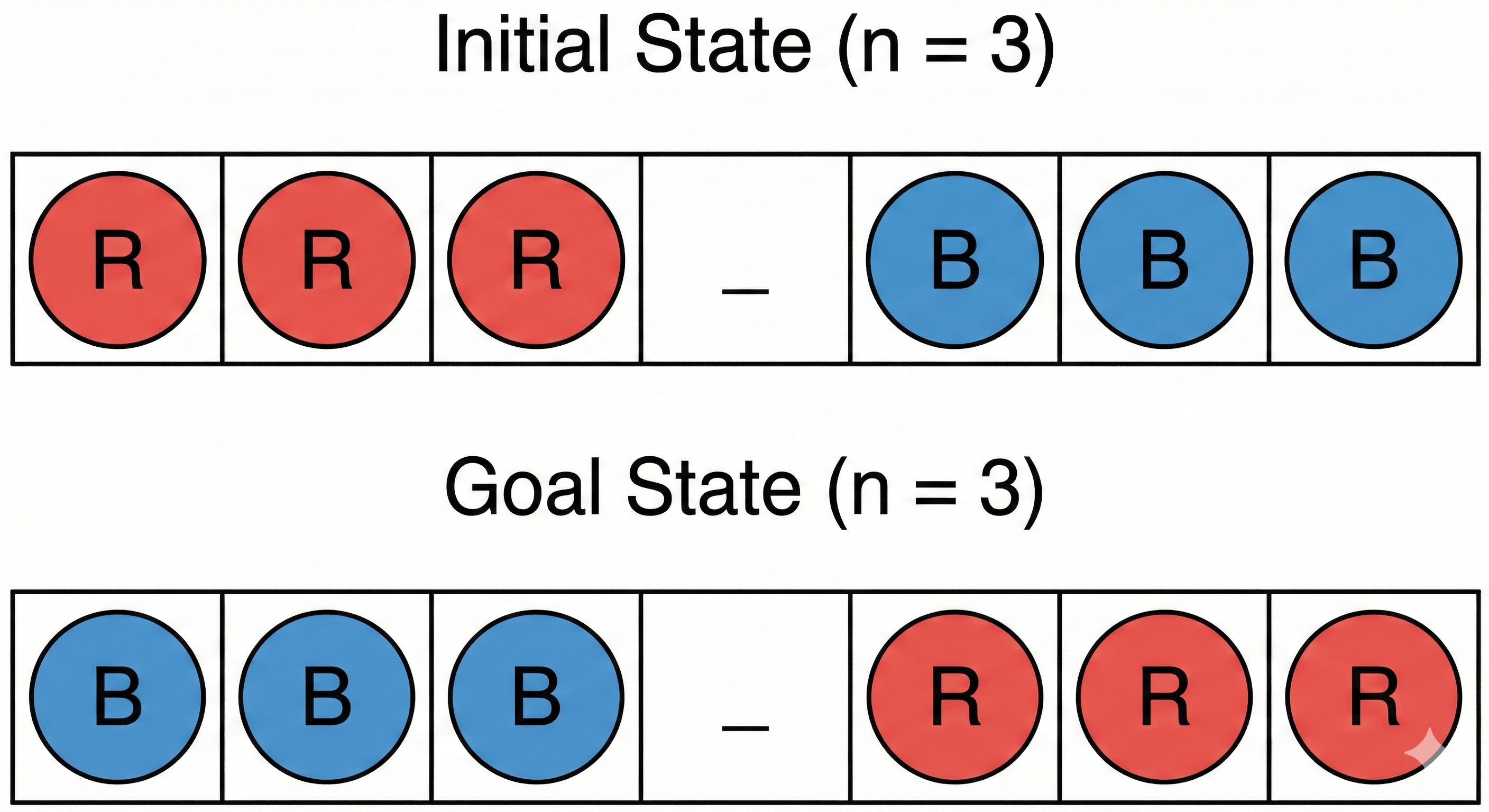}
  \caption{Checker Jumping initial configuration for $n$ red and $n$ blue checkers on a $(2n+1)$-cell board with a single empty space.}
  \label{fig:checker_jumping_setup}
\end{figure}

\begin{figure}
  \centering
  \includegraphics[width=.9\columnwidth]{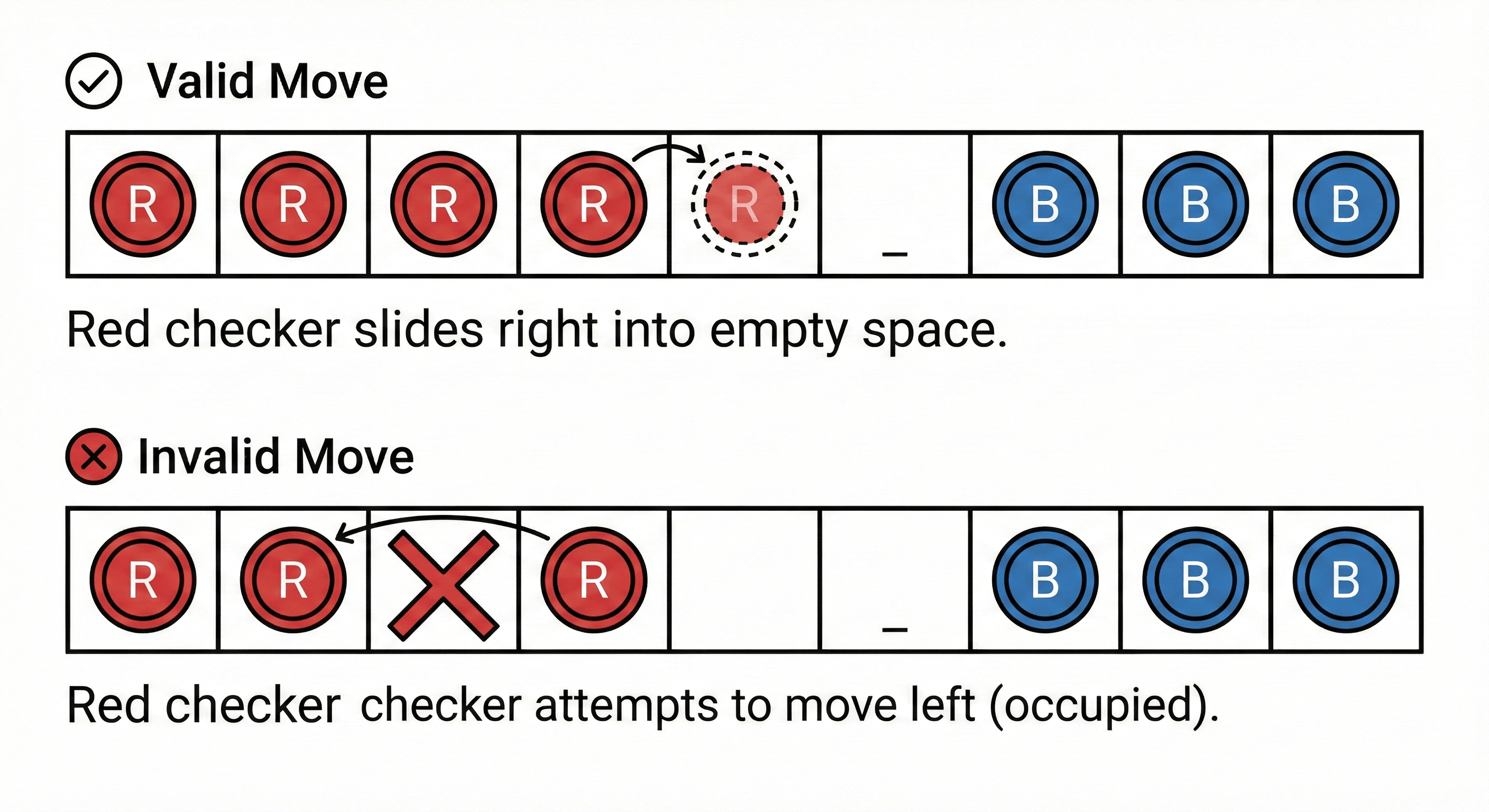}
  \caption{Valid local moves in Checker Jumping: (i) slide into adjacent empty space; (ii) jump over one opposing checker into the empty space.}
  \label{fig:checker_jumping_moves}
\end{figure}

Checker Jumping is a one-dimensional sequential planning problem defined over a finite discrete state space with strict local movement constraints. Its symmetric interactions and reversible local dependencies make it a useful testbed for assessing mid-horizon planning, conflict resolution, and consistency maintenance.

\paragraph{Formal Structure.}
For $n$ red and $n$ blue checkers, the instance is:
\[
\mathcal{C}_n = (S, A, T, V, G).
\]

\paragraph{State Space.}
The board has $2n+1$ linear positions. A state is:
\[
s \in \{R, B, \_\}^{2n+1}, \qquad \#R=n,\;\#B=n,\;\#\_=1.
\]
The state-space cardinality is:
\[
|S| = \frac{(2n+1)!}{n!\,n!\,1!}.
\]
The initial and goal configurations are mirror images:
\[
(R^n,\,\_,\,B^n) \quad\rightarrow\quad (B^n,\,\_,\,R^n).
\]

\paragraph{Actions.}
Two moves are permitted:
\begin{enumerate}
    \item \textbf{Slide:} move a checker into an adjacent empty cell.
    \item \textbf{Jump:} jump over exactly one opposing checker into the empty cell.
\end{enumerate}
Red moves right; blue moves left.

\paragraph{Transitions and Validity.}
A transition $T(s_t,a_t)$ is defined only if slide/jump and directional constraints are satisfied.  
Validity is:
\[
V(s_t,a_t)=1 \text{ if the move obeys rules, else } 0.
\]

\paragraph{Goal.}
A trajectory is accepted if $s_T \in G = \{(B^n,\,\_,\,R^n)\}$.

\paragraph{Example ($n=1$).}
\[
(R,\,\_,\,B)\rightarrow(\_,R,B)\rightarrow(B,R,\_).
\]
Moving a red checker left is illegal and yields $V(s,a)=0$.

\paragraph{Complexity Parameterization.}
The complexity parameter is $\lambda=n$.  
The minimum solution length is:
\[
L_{\min}(n) = (n+1)^2 - 1,
\]
growing quadratically with $n$, while the state space grows combinatorially.

\paragraph{Complexity and Cognitive Load.}
Although the horizon increases polynomially, long trajectories amplify local errors. With per-step accuracy $p$, the success probability scales as:
\[
P_{\text{success}}\approx p^{L_{\min}(n)}.
\]
As $n$ grows, maintaining consistent board states becomes increasingly difficult due to symmetric interactions and reversible conflicts, making error propagation and deadlocks more likely.

\paragraph{Relevance for Reasoning Evaluation.}
Checker Jumping stresses sequential consistency rather than recursive structure. Increasing $n$ requires LRMs to maintain valid local interactions over longer horizons while managing global configuration shifts, providing a controlled axis for evaluating mid-range planning fidelity and susceptibility to cumulative reasoning errors.

\subsubsection{River Crossing}

\begin{figure}
  \centering
  \includegraphics[width=0.9\columnwidth]{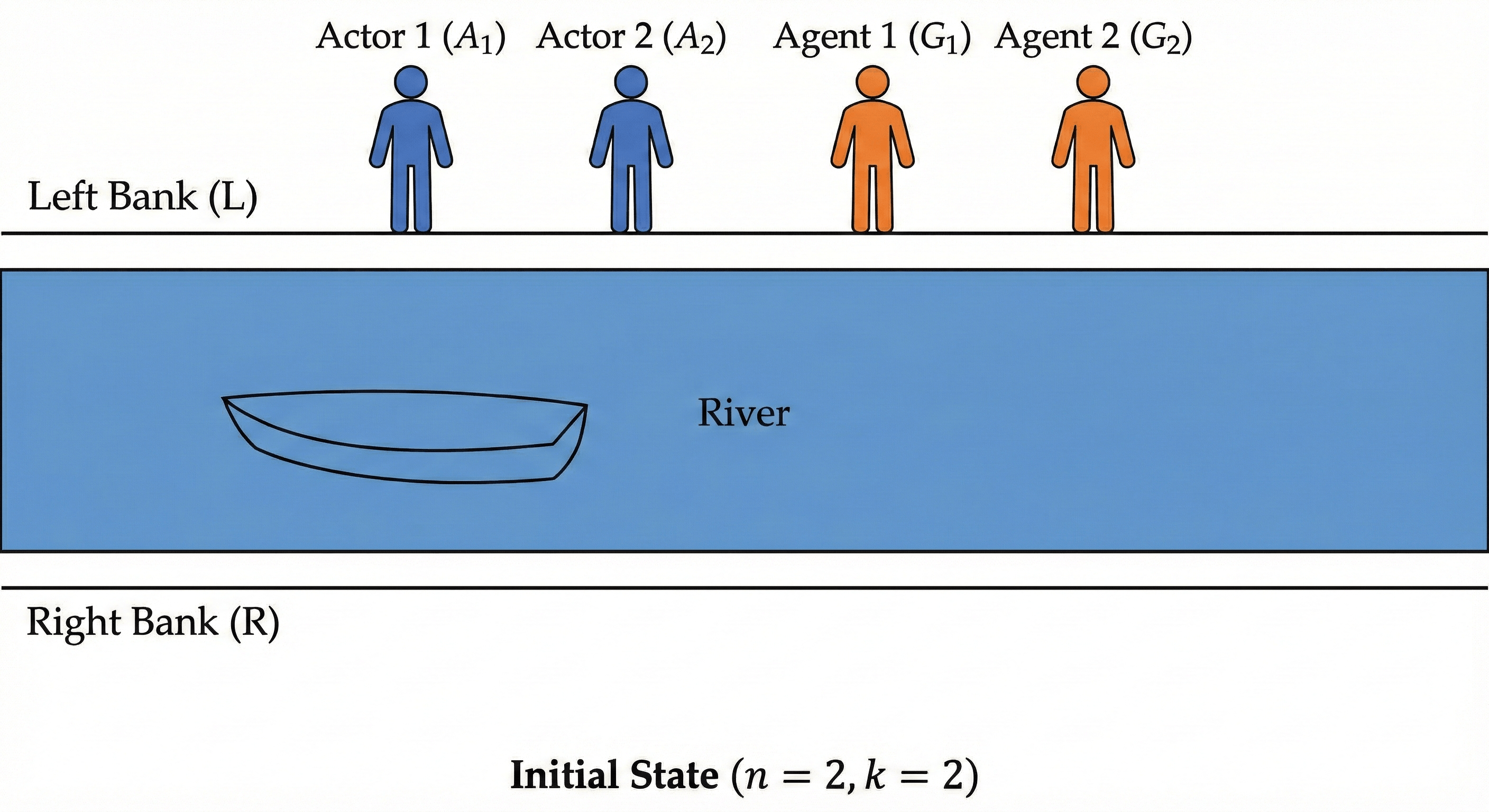}
  \caption{River Crossing: example goal configuration where all actors and agents are safely transported to the right bank.}
  \label{fig:river_crossing_goal}
\end{figure}

\begin{figure}
  \centering
  \includegraphics[width=0.9\columnwidth]{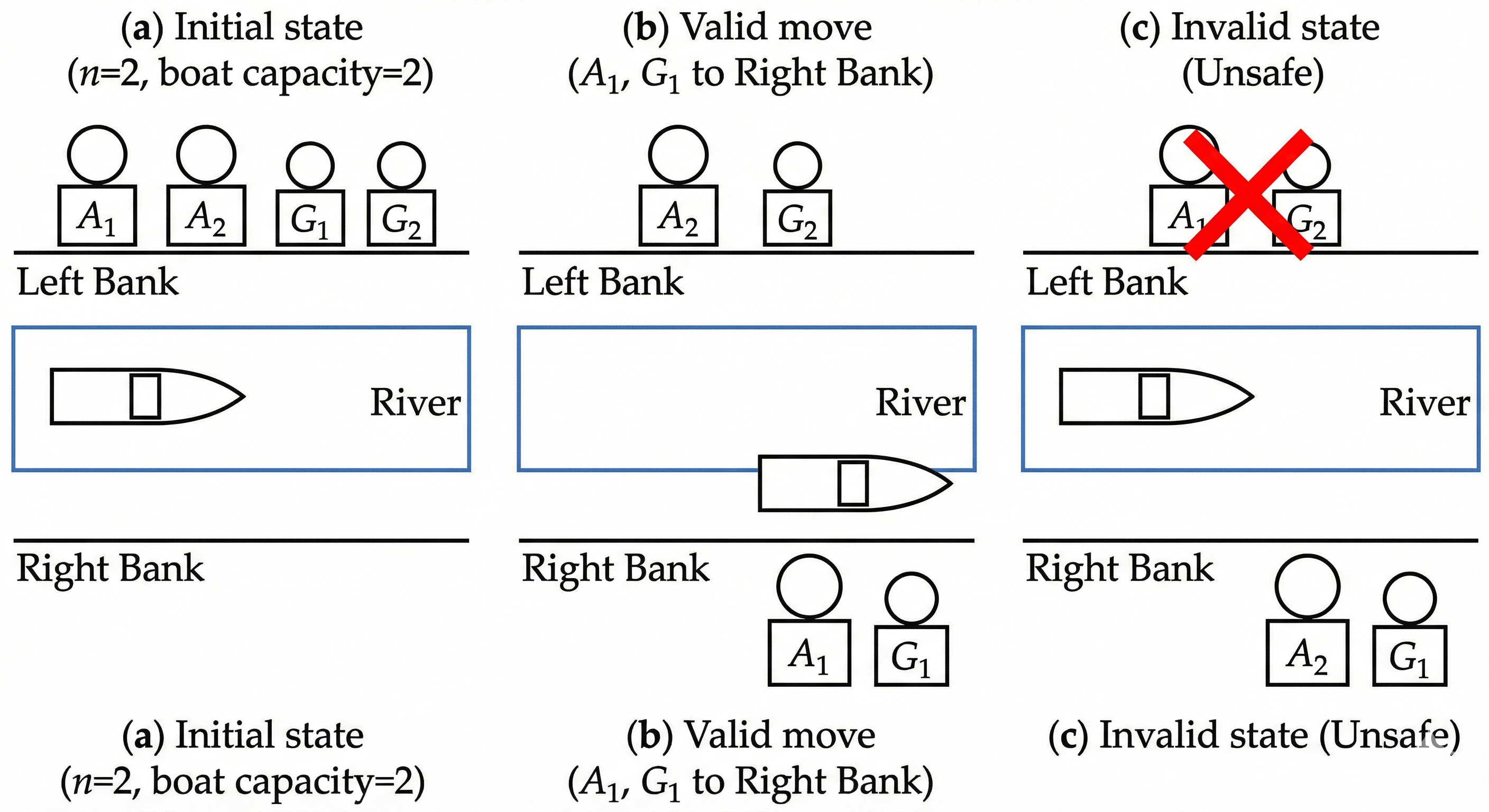}
  \caption{Valid boat moves in River Crossing: the boat carries a non-empty subset of entities (up to capacity $k$) between banks while maintaining safety constraints.}
  \label{fig:river_crossing_moves}
\end{figure}

River Crossing is a safety-constrained sequential planning problem where each intermediate state must satisfy strict validity conditions. Unlike purely structural puzzles (e.g., Tower of Hanoi), River Crossing requires maintaining global \emph{safety invariants} at every step; a single unsafe configuration invalidates the entire trajectory. This makes the task well-suited for evaluating whether models can maintain explicit constraints under increasing complexity.

\paragraph{Formal Structure.}
For $n$ actor–agent pairs and boat capacity $k$, an instance is:
\[
\mathcal{R}_{n,k} = (S, A, T, V, G).
\]

\paragraph{State Representation.}
Actors $A^\star=\{a_1,\dots,a_n\}$ and their protecting agents $G^\star=\{g_1,\dots,g_n\}$ are placed on left or right banks, along with the boat:
\[
s=(L,R,b).
\]
The raw state count is:
\[
|S| = 2^{2n+1},
\]
though many states violate safety rules.

\paragraph{Safety Constraint.}
A bank $B$ is unsafe if an actor $a_i\in B$ is present without its agent $g_i$, while at least one non-matching agent is also present:
\[
\exists i:\; a_i\in B,\ g_i\notin B,\ \exists j\neq i:\ g_j\in B.
\]
The predicate $V_{\text{safe}}(s)=1$ iff both banks are safe.

\paragraph{Actions.}
The boat carries a non-empty subset $X$ of entities on its current bank, subject to:
\[
1 \le |X| \le k.
\]
An action $\text{Cross}(X)$ moves $X$ to the opposite bank and flips the boat position.

\paragraph{Transitions and Validity.}
A move is valid only if the subset is legal and the resulting state remains safe:
\[
V(s_t,a_t)=1 \iff 1\le |X|\le k,\ X \subseteq B(s_t),\ V_{\text{safe}}(s_{t+1})=1.
\]

\paragraph{Goal.}
All entities must reach the right bank:
\[
G=\{(\emptyset,\ A^\star\cup G^\star,\ R)\}.
\]

\paragraph{Complexity Parameterization.}
The complexity parameter is:
\[
\lambda=(n,k).
\]
Increasing $n$ raises safety-constraint density; decreasing $k$ increases planning depth.  
From a state with $m$ entities on the boat bank:
\[
|A(s)|=\sum_{i=1}^{\min(k,m)} \binom{m}{i},
\]
which grows rapidly with $m$ and $k$.

\paragraph{Cognitive Load and Failure Modes.}
River Crossing difficulty arises from:
\begin{enumerate}
    \item \textbf{Global safety dependence}: each action requires evaluating the entire next configuration.
    \item \textbf{Long-horizon planning}: invalid states are often irreversible.
    \item \textbf{Branching under constraints}: many syntactically possible boat selections are unsafe.
\end{enumerate}

For a minimum-length plan $L_{\min}(n,k)$ and per-step correctness probability $p$:
\[
P_{\text{success}} \approx p^{L_{\min}(n,k)},
\]
showing rapid degradation as complexity grows.

\paragraph{Relevance for Reasoning Evaluation.}
River Crossing stresses models’ ability to maintain explicit constraint satisfaction throughout planning. Because a single unsafe intermediate state invalidates the solution, the task exposes common reasoning failures such as neglected constraints, incorrect global-state tracking, and confident-but-invalid step sequences under increasing complexity.

\subsubsection{Boolean Satisfiability (SAT)}

\begin{figure}
  \centering
  \includegraphics[width=0.9\columnwidth]{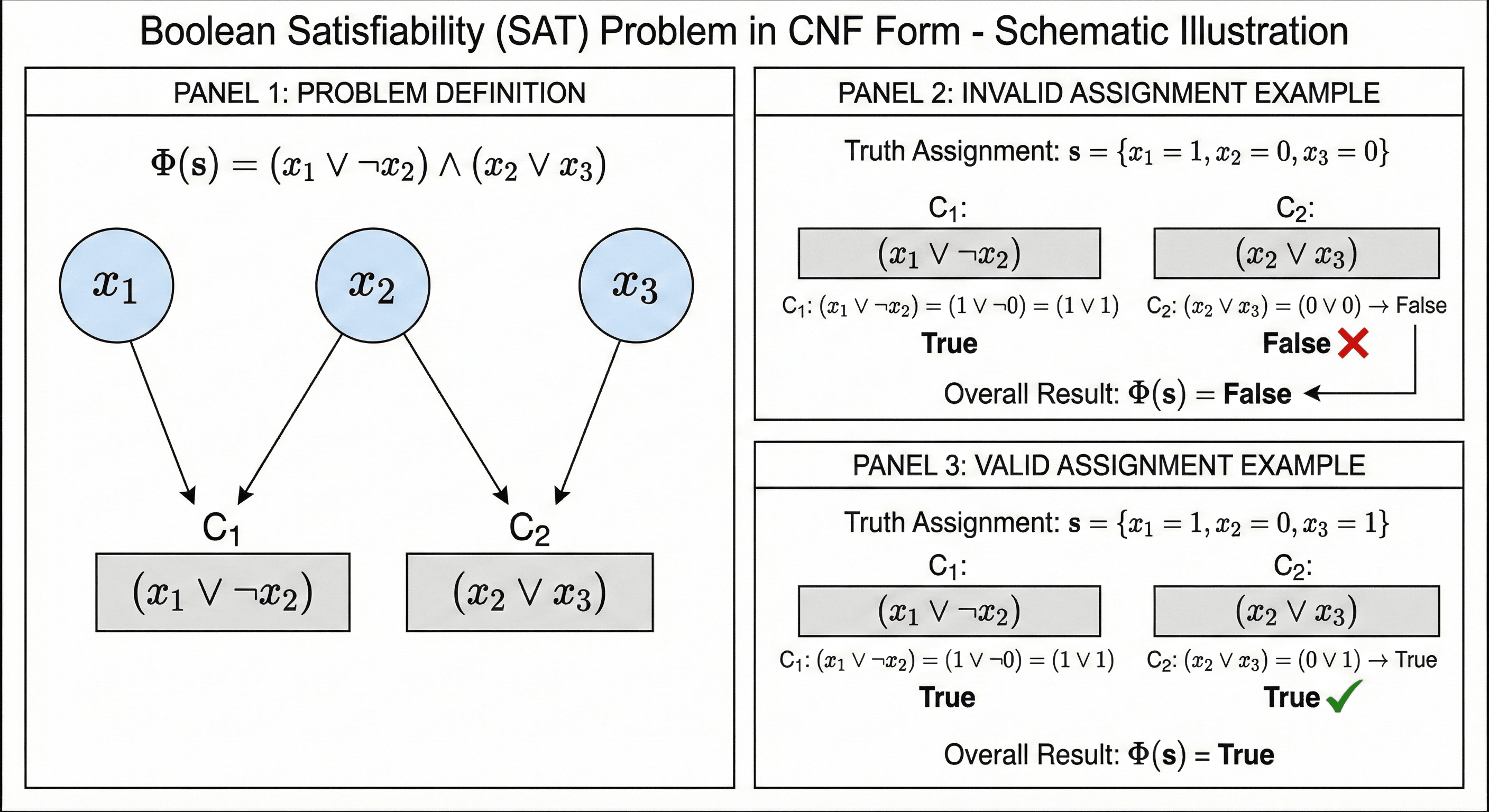}
  \caption{Boolean Satisfiability (SAT): CNF formula structure with variables, literals, and clause interactions.}
  \label{fig:boolean_sat_diagram}
\end{figure}

Boolean Satisfiability (SAT) is a canonical global constraint satisfaction problem whose solution requires assigning truth values to variables such that all clauses in a conjunctive normal form (CNF) formula evaluate to true. Unlike trajectory-based puzzles, SAT stresses global logical consistency across interacting constraints, making it a principled axis for probing reasoning robustness as complexity scales.

\paragraph{Formal Structure.}
Let $X=\{x_1,\dots,x_n\}$ be Boolean variables, and let a CNF formula be:
\[
\Phi = \bigwedge_{i=1}^{m} C_i,
\]
where each clause $C_i$ is a disjunction of literals ($x_j$ or $\lnot x_j$).  
An instance is defined as:
\[
\mathcal{S}_{n,m} = (S, A, T, V, G).
\]

\paragraph{State Space.}
A state is a complete assignment:
\[
s : X \rightarrow \{0,1\}.
\]
The assignment space grows exponentially:
\[
|S| = 2^n.
\]

\paragraph{Actions and Transitions.}
Although SAT is not inherently sequential, it may be interpreted as assignment-space search.  
A move flips a variable:
\[
a=\text{Flip}(x_j),\quad T(s,a)=s',
\]
where only $x_j$ is toggled.

\paragraph{Validity Constraint.}
A state is valid iff all clauses evaluate to true:
\[
V(s)=1 \iff \Phi(s)=1 = \bigwedge_{i=1}^{m} C_i(s).
\]

\paragraph{Goal.}
The solution set is:
\[
G=\{s\in S\mid \Phi(s)=1\}.
\]

\paragraph{Example.}
For:
\[
\Phi=(x_1\lor \lnot x_2)\land(x_2\lor x_3),
\]
the assignment $(1,0,0)$ violates the second clause, while $(1,0,1)$ satisfies both clauses and is therefore in $G$.

\paragraph{Failure Modes.}
Incorrect solutions typically exhibit:
\begin{itemize}
    \item incomplete assignments,
    \item contradictory values,
    \item missed clause violations,
    \item confident but unsatisfied CNF evaluations.
\end{itemize}

\paragraph{Complexity Parameterization.}
We define:
\[
\lambda=(n,m),
\]
where $n$ controls assignment-space growth, and $m$ governs constraint density.  
Clause-to-variable ratio:
\[
\alpha=\frac{m}{n}
\]
is a key complexity axis.  
For random 3-SAT, hardness peaks near $\alpha \approx 4.26$ (phase transition).

\paragraph{Complexity and Cognitive Load.}
As $n$ increases, the search space doubles per variable.  
As clause density grows, variable–clause dependencies become denser.  
If each clause is independently satisfied with probability $p_c$:
\[
P(\text{all clauses satisfied})\approx p_c^{\,m},
\]
which decays exponentially in $m$.  

SAT therefore stresses:
\begin{itemize}
    \item global logical consistency,
    \item simultaneous multi-clause satisfaction,
    \item non-local reasoning dependencies.
\end{itemize}

\paragraph{Relevance for Reasoning Evaluation.}
SAT isolates global constraint handling without relying on trajectory structure. Increasing $(n,\alpha)$ amplifies:
\begin{itemize}
    \item assignment-space explosion,
    \item clause-interaction density,
    \item likelihood of subtle but catastrophic violations.
\end{itemize}
Failures at higher complexity typically manifest as incomplete assignments, overlooked clause violations, or internally inconsistent reasoning traces, making SAT a direct probe of reasoning stability under global constraint pressure.

\subsubsection{Cryptarithmetic}

\begin{figure}
  \centering
  \includegraphics[width=0.9\columnwidth]{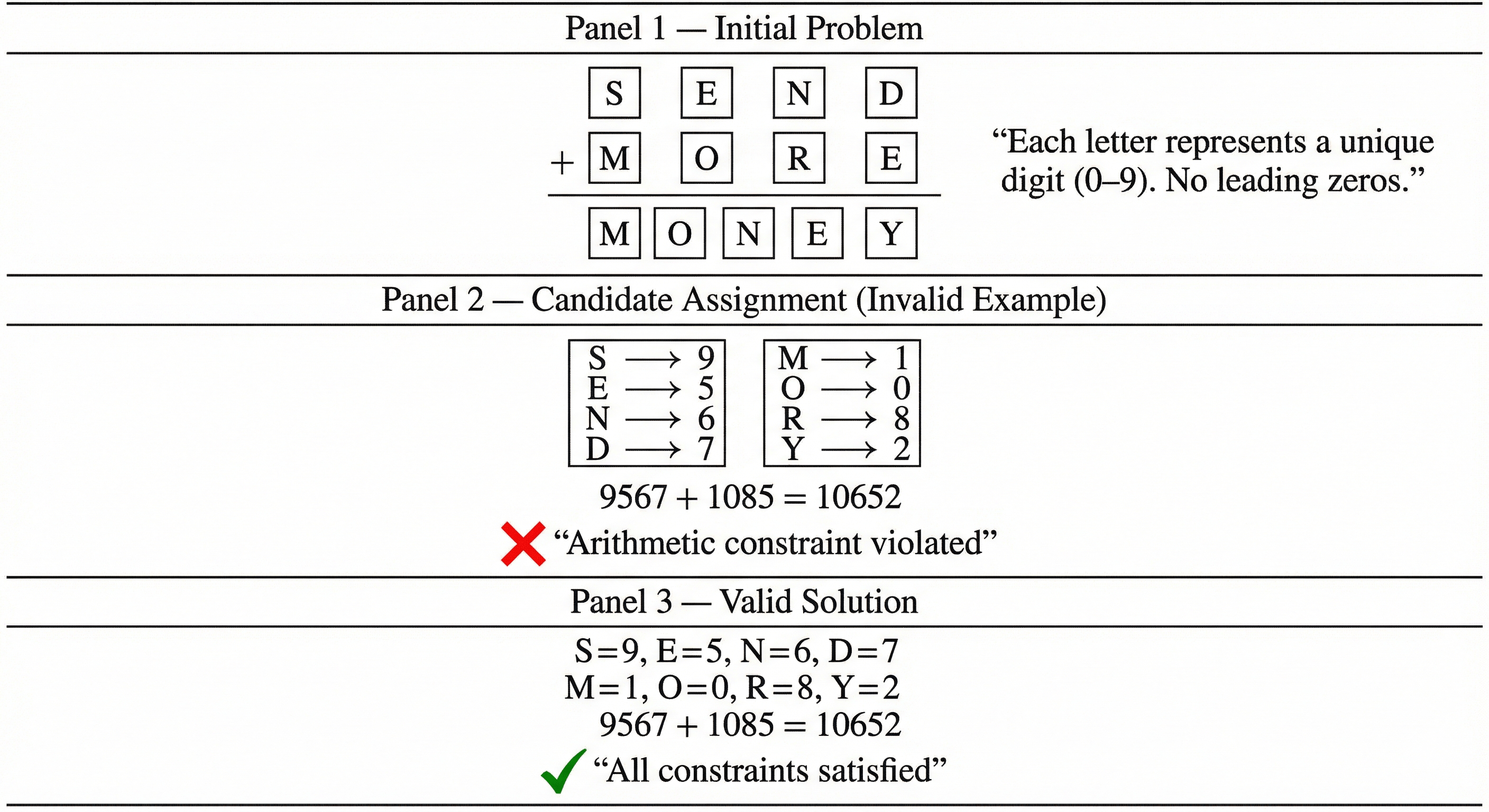}
  \caption{Cryptarithmetic (alphametic) puzzle: each letter must map to a unique digit, and the decoded arithmetic expression must hold exactly.}
  \label{fig:cryptarithmetic}
\end{figure}

Cryptarithmetic is a symbolic constraint satisfaction problem in which each letter maps to a unique digit and an arithmetic equation must hold under this mapping. Because valid solutions require enforcing injectivity, preventing leading zeros, and satisfying coupled column-wise addition constraints with carries, cryptarithmetic provides a controlled setting for assessing global consistency maintenance under increasing symbolic and arithmetic complexity.

\paragraph{Formal Structure.}
Let $\Sigma=\{\ell_1,\dots,\ell_k\}$ be the distinct letters in the puzzle.  
A solution is an injective function:
\[
f:\Sigma \rightarrow \{0,1,\dots,9\}.
\]
We define an instance as:
\[
\mathcal{K} = (S, A, T, V, G),
\]
where $S$ is the assignment space and $V$ enforces global validity.

\paragraph{State Space.}
A state $s\in S$ is a complete digit assignment for all $k$ letters.  
The number of injective assignments is:
\[
|S| = P(10,k)=\frac{10!}{(10-k)!}.
\]

\paragraph{Actions.}
For search-style interpretations, a move assigns an unused digit to an unassigned letter:
\[
a=\text{Assign}(\ell \leftarrow d).
\]
The transition extends the partial mapping with this assignment.

\paragraph{Validity Constraints.}
A full assignment $f$ is valid iff all constraints hold:
\begin{enumerate}
    \item \textbf{Injectivity:} $f(\ell_i)\neq f(\ell_j)$ for all $\ell_i\neq\ell_j$.
    \item \textbf{No leading zeros:} any leading letter must satisfy $f(\ell)\neq 0$.
    \item \textbf{Arithmetic consistency:} decoding under $f$ must satisfy the equation exactly.
\end{enumerate}
Thus,
\[
V(f)=1 \iff \text{all constraints hold}.
\]

\paragraph{Goal.}
The solution set is:
\[
G=\{f\in S\mid V(f)=1\}.
\]

\paragraph{Example.}
In \texttt{SEND + MORE = MONEY}, the mapping  
\[
S{=}9,\,E{=}5,\,N{=}6,\,D{=}7,\,M{=}1,\,O{=}0,\,R{=}8,\,Y{=}2
\]
yields  
\[
9567 +1085 = 10652,
\]
so $f\in G$.  
Invalid assignments typically violate injectivity, assign zero to a leading letter, or break column-wise arithmetic.

\paragraph{Column Constraints.}
For column $t$ with carry $c_t$:
\[
f(\ell^{(1)}_t) + f(\ell^{(2)}_t) + c_t \equiv f(\ell^{(3)}_t) \pmod{10},
\]
\[
c_{t+1}=\left\lfloor \frac{f(\ell^{(1)}_t) + f(\ell^{(2)}_t) + c_t}{10} \right\rfloor.
\]
Assignments forcing violation in any column are invalid.

\paragraph{Complexity Parameterization.}
We define:
\[
\lambda=(k, L, \kappa),
\]
where:
\begin{itemize}
    \item $k$ = number of distinct letters,
    \item $L$ = maximum column depth (number length),
    \item $\kappa$ = carry-chain coupling depth.
\end{itemize}
The assignment space grows as $P(10,k)$ and carry chains introduce long-range dependencies.

\paragraph{Complexity and Cognitive Load.}
Difficulty increases with:
\begin{enumerate}
    \item \textbf{Larger symbol sets ($k\uparrow$)}: expands injective search space.
    \item \textbf{Longer carry chains ($\kappa\uparrow$)}: strengthens cross-column coupling.
    \item \textbf{More columns ($L\uparrow$)}: increases the number of simultaneous digit-sum constraints.
\end{enumerate}
If each column is independently satisfied with probability $p_{\text{col}}$, then:
\[
P(\text{all columns satisfied}) \approx p_{\text{col}}^{\,L},
\]
showing exponential decay with $L$.

\paragraph{Relevance for Reasoning Evaluation.}
Cryptarithmetic isolates symbolic–numeric reasoning under explicit global constraints. Increasing $k$, $L$, or $\kappa$ stresses requirement for globally consistent mappings, making this task particularly revealing of reasoning failures such as digit reuse, overlooked leading-zero violations, incorrect carry propagation, and confident but inconsistent arithmetic.

\subsubsection{Graph Coloring}

\begin{figure}
  \centering
  \includegraphics[width=0.9\columnwidth]{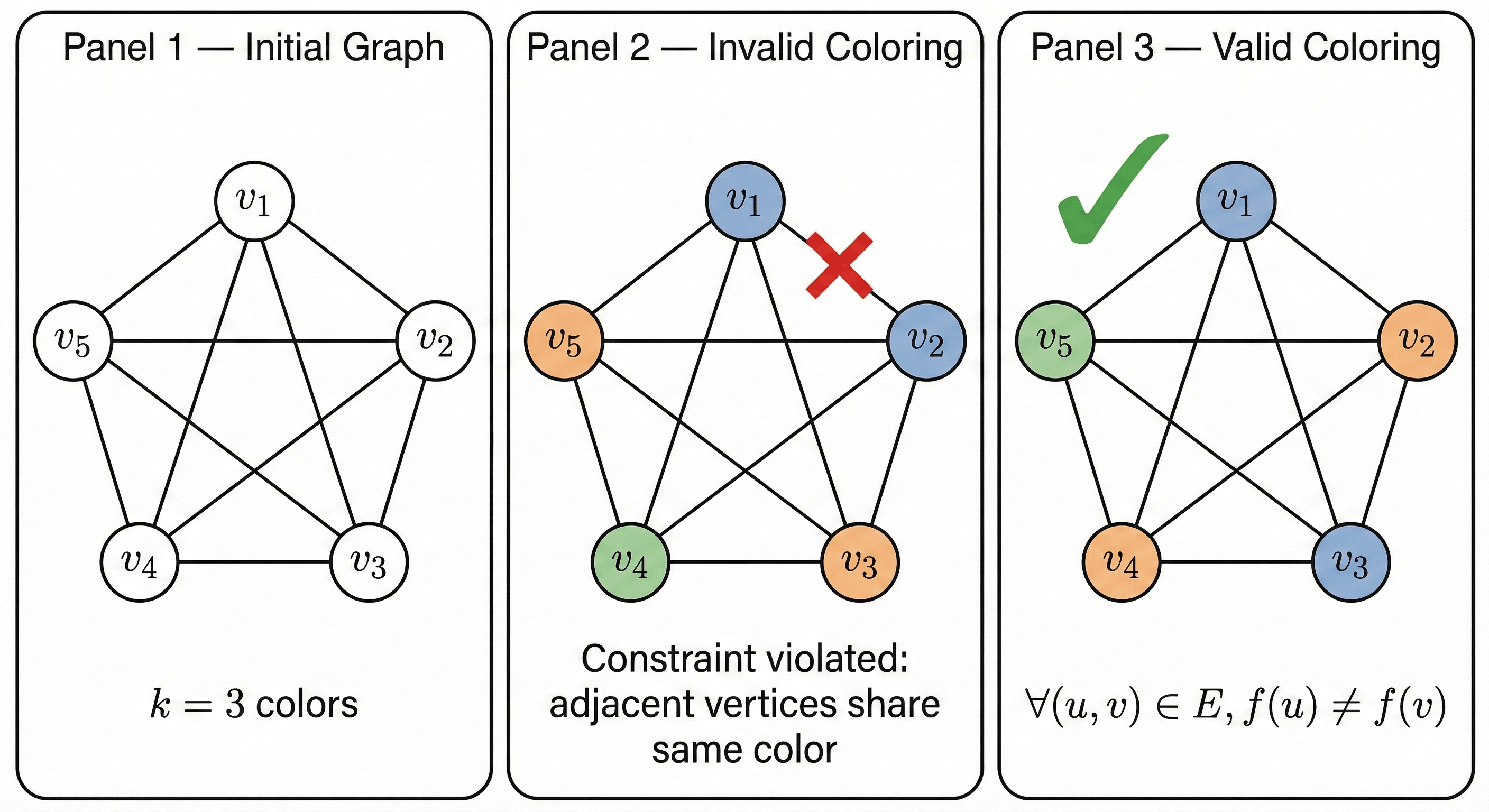}
  \caption{Graph Coloring example: each vertex receives a color from a fixed palette such that adjacent vertices do not share the same color.}
  \label{fig:graph_coloring_diagram}
\end{figure}

Graph Coloring is a classical combinatorial constraint satisfaction problem requiring assignment of colors to vertices of a graph such that adjacent vertices receive different colors. Because difficulty can be controlled by graph size and edge density—and correctness depends on enforcing global adjacency constraints—Graph Coloring serves as a principled benchmark for evaluating reasoning under increasing constraint interaction.

\paragraph{Formal Structure.}
Let $G=(V,E)$ be an undirected graph with $|V|=n$ vertices and $|E|=m$ edges.  
Given $k$ colors:
\[
\mathcal{C}=\{1,\dots,k\},
\]
a complete coloring is:
\[
f:V\rightarrow\mathcal{C}.
\]
An instance is defined as:
\[
\mathcal{G}_{n,m,k}=(S,A,T,V,G^\star),
\]
where $S$ is the coloring space and $V$ is the global validity predicate.

\paragraph{State Space.}
A state $s\in S$ is a complete assignment of colors to vertices:
\[
|S| = k^n.
\]

\paragraph{Actions.}
For search-style interpretations, an action assigns a color to an uncolored vertex:
\[
a = \text{Color}(v \leftarrow c).
\]
The transition extends the current partial mapping by setting $f(v)=c$.

\paragraph{Validity Constraint.}
A coloring is valid iff no adjacent vertices share a color:
\[
V(f)=1 \iff \forall (u,v)\in E,\ f(u)\neq f(v).
\]
Equivalently, the conflict count:
\[
\text{Conf}(f)=\sum_{(u,v)\in E}\mathbf{1}[f(u)=f(v)]
\]
is zero for valid colorings.

\paragraph{Goal.}
The solution set is:
\[
G^\star = \{f\in S\mid V(f)=1\}.
\]

\paragraph{Example.}
For a cycle $C_4$ with $k=2$ colors, the assignment
\[
f(1)=1,\ f(2)=2,\ f(3)=1,\ f(4)=2
\]
is valid.  
Assigning $f(1)=1, f(2)=1$ produces a conflict on edge $(1,2)$ and is invalid.

\paragraph{Right and Wrong Moves.}
A right move assigns a color that avoids immediate conflicts with already-colored neighbors:
\[
\forall u\in N(v),\ f(u)\neq c.
\]
A wrong move creates a conflict with at least one neighbor. Many locally valid moves may still lead to future dead-ends under small $k$, highlighting the need for global consistency.

\paragraph{Complexity Parameterization.}
We define complexity as:
\[
\lambda=(n,m,k),
\]
where $n$ controls assignment dimensionality, $m$ controls constraint density, and $k$ controls feasibility.

Average degree:
\[
\bar{d}=\frac{2m}{n}, 
\qquad
\rho=\frac{2m}{n(n-1)}
\]
provide interpretable measures of constraint density.

\paragraph{Complexity and Cognitive Load.}
For random colorings, the probability of avoiding a conflict on any edge is approximately:
\[
P(V(f)=1)\approx \left(1-\frac{1}{k}\right)^m,
\]
which decays exponentially as $m$ increases.  
Graphs near their chromatic threshold (i.e., $k\approx \chi(G)$) exhibit the strongest interaction constraints, making them hardest for LRMs.

\paragraph{Relevance for Reasoning Evaluation.}
Graph Coloring stresses global dependency maintenance over a combinatorially large assignment space. As $(n,m)$ grow, LRMs must avoid adjacency conflicts across increasingly dense constraint networks. Typical failures include overlooked conflicts, inconsistent partial assignments, or confident-but-invalid final colorings, making this task a direct probe of reasoning stability under combinatorial constraint interaction.

\subsubsection{Water Jug Problem}

\begin{figure}
  \centering
  \includegraphics[width=0.9\columnwidth]{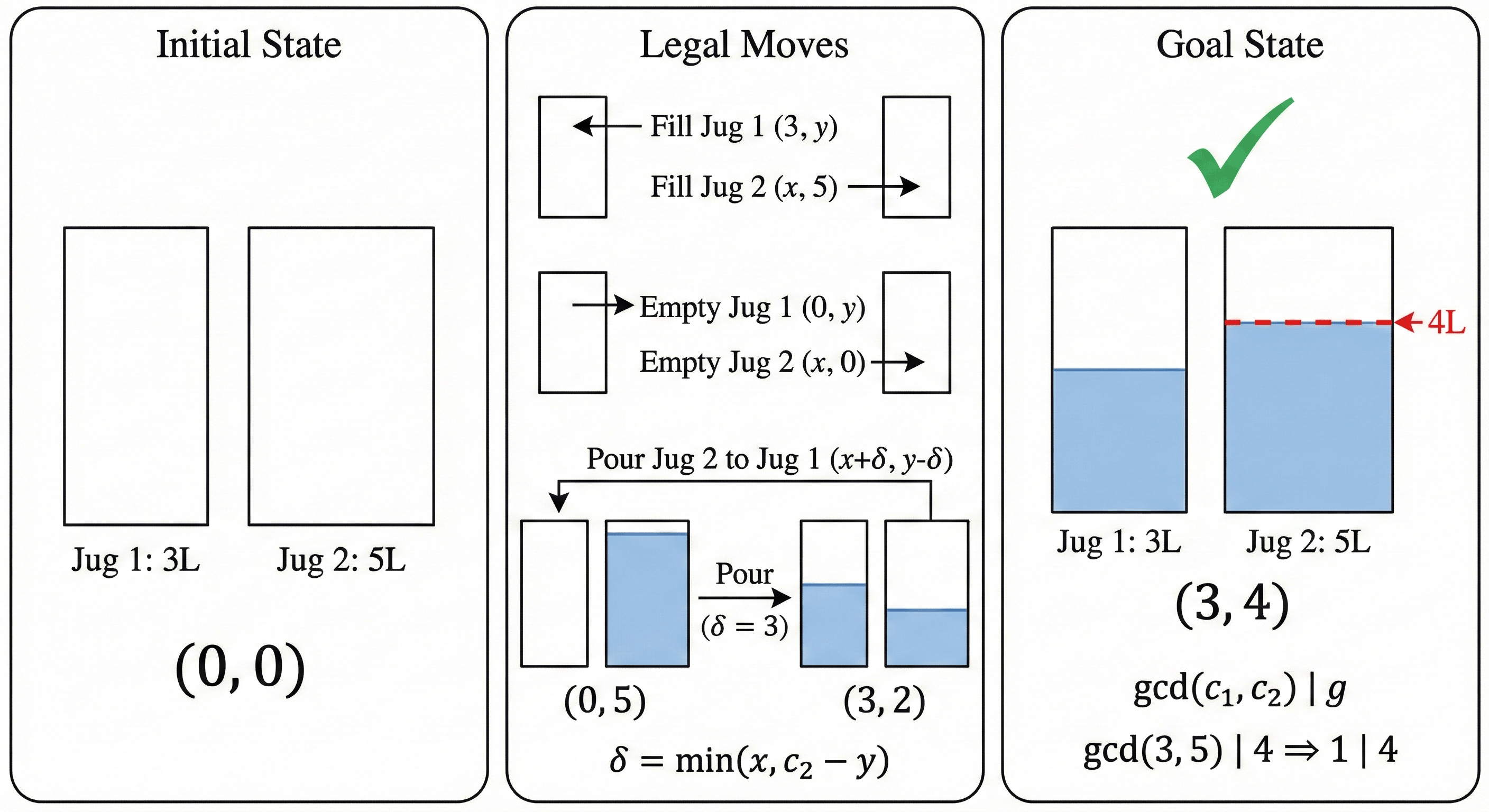}
  \caption{Water Jug setup with two jugs of capacities $c_1$ and $c_2$, illustrating permissible operations (fill, empty, pour).}
  \label{fig:water_jug_diagram}
\end{figure}

The Water Jug problem is a deterministic arithmetic planning task defined over a finite discrete state space. Given two jugs of fixed capacities and an unlimited water source, the goal is to obtain an exact target volume using only capacity-respecting operations. Because correctness relies on accurate arithmetic transitions and long-horizon sequential reasoning, the task provides a clean setting for evaluating numeric state tracking and operation-validity maintenance.

\paragraph{Formal Structure.}
For jug capacities $c_1, c_2 \in \mathbb{N}$ and target $g \in \mathbb{N}$, an instance is:
\[
\mathcal{W}_{c_1,c_2,g} = (S, A, T, V, G).
\]

\paragraph{State Space.}
A state records current water volumes:
\[
s = (x,y), \qquad 0 \le x \le c_1,\; 0 \le y \le c_2.
\]
Thus:
\[
|S| = (c_1+1)(c_2+1).
\]

\paragraph{Actions.}
Allowed operations are:
\begin{enumerate}
    \item Fill Jug 1: $(x,y)\rightarrow(c_1,y)$
    \item Fill Jug 2: $(x,y)\rightarrow(x,c_2)$
    \item Empty Jug 1: $(x,y)\rightarrow(0,y)$
    \item Empty Jug 2: $(x,y)\rightarrow(x,0)$
    \item Pour $1\rightarrow 2$: $(x,y)\rightarrow(x-\delta, y+\delta)$, $\delta=\min(x,c_2-y)$
    \item Pour $2\rightarrow 1$: $(x,y)\rightarrow(x+\delta, y-\delta)$, $\delta=\min(y,c_1-x)$
\end{enumerate}

\paragraph{Transition and Validity.}
Transitions follow deterministic arithmetic rules.  
A move is valid iff the resulting state satisfies capacity constraints:
\[
V(s_t,a_t)=1 \iff 0 \le x \le c_1,\; 0 \le y \le c_2.
\]

\paragraph{Goal.}
A state is a solution if:
\[
G=\{(x,y)\mid x=g\ \text{or}\ y=g\}.
\]

\paragraph{Example.}
For $c_1=3$, $c_2=5$, $g=4$, one valid sequence is:
\[
(0,0)\rightarrow(0,5)\rightarrow(3,2)\rightarrow(0,2)\rightarrow(2,0)\rightarrow(2,5)\rightarrow(3,4),
\]
where the final state satisfies $y=4$.

\paragraph{Invalid Moves.}
Examples include:
\begin{itemize}
    \item Overfilling a jug beyond capacity,
    \item Producing negative volumes,
    \item Incorrect pour amounts (not using $\delta=\min$ rule),
\end{itemize}
e.g., from $(3,2)$, attempting:
\[
(3,2)\rightarrow(1,5)
\]
is invalid since $\delta=\min(3,3)=3$, not $2$.

\paragraph{Feasibility.}
A classical solvability condition is:
\[
g \le \max(c_1,c_2) \quad\text{and}\quad \gcd(c_1,c_2)\mid g.
\]

\paragraph{Complexity Parameterization.}
We define:
\[
\lambda=(c_1,c_2,g),
\]
with state-space size:
\[
|S|=(c_1+1)(c_2+1).
\]
Although polynomial in capacity, minimal solution length can be large and depends on the Diophantine structure of $(c_1,c_2,g)$.

\paragraph{Complexity and Cognitive Load.}
Difficulty increases with:
\begin{enumerate}
    \item Larger capacities (more states),
    \item Targets requiring long operational cycles,
    \item Small $\gcd(c_1,c_2)$ relative to capacities (longer required sequences).
\end{enumerate}

If each step is correct with probability $p$, then:
\[
P_{\text{success}}\approx p^{L_{\min}},
\]
so long sequences amplify local arithmetic errors.

\paragraph{Relevance for Reasoning Evaluation.}
Water Jug exposes reasoning failures in arithmetic state tracking and operation validity. Increasing capacity enlarges the reachable state graph, while harder targets require longer sequences. Common LRM failure modes include incorrect pour arithmetic, unnoticed capacity violations, cyclic loops, or confident-but-infeasible solutions, making this task a targeted probe of numeric reasoning robustness.

\subsubsection{Sudoku}

\begin{figure}
  \centering
  \includegraphics[width=0.9\columnwidth]{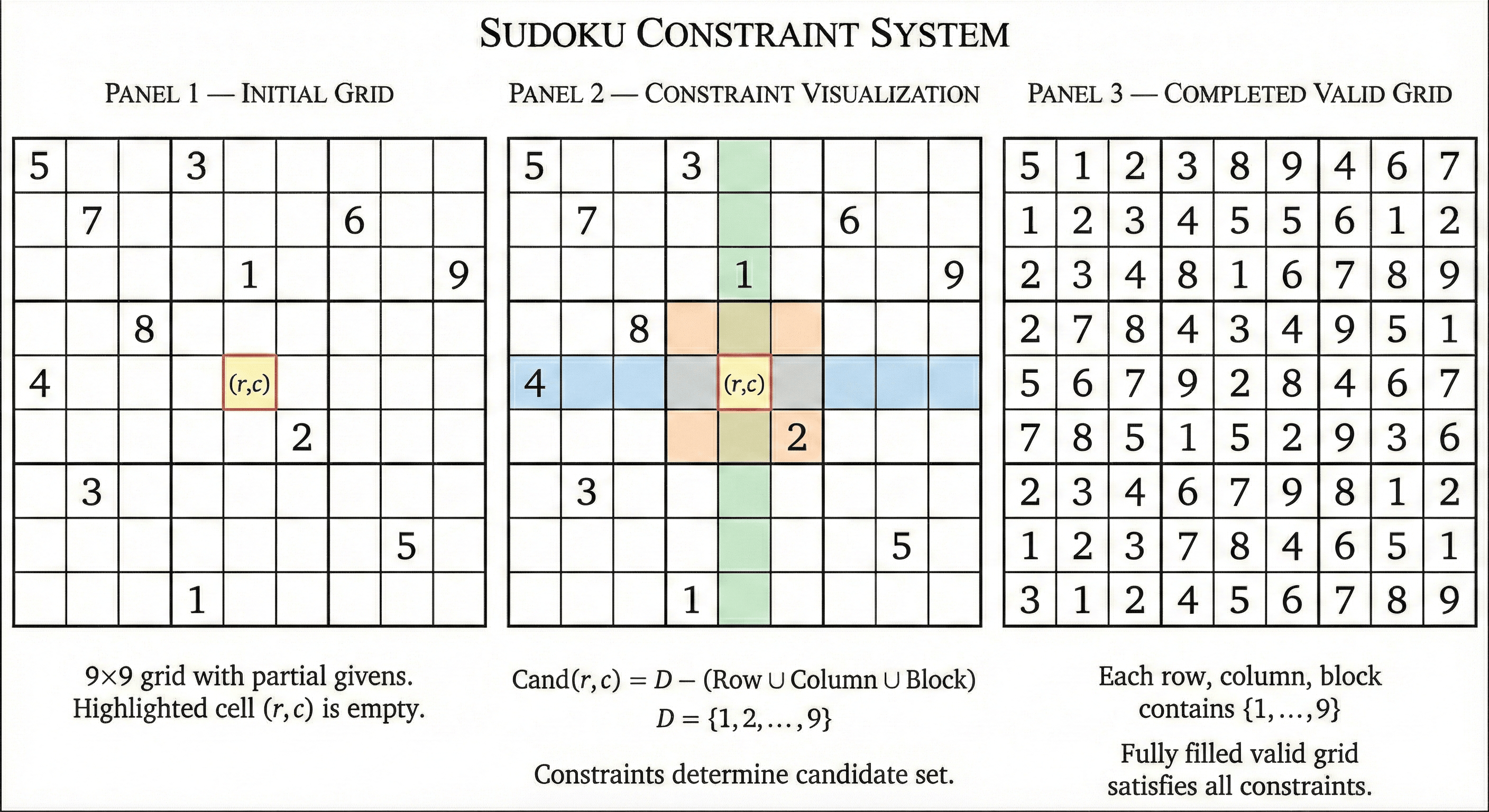}
  \caption{Sudoku grid structure: digits must satisfy row, column, and $3\times 3$ subgrid uniqueness constraints.}
  \label{fig:sudoku_diagram}
\end{figure}

Sudoku is a dense global constraint satisfaction problem defined over a $9\times 9$ grid. The task is to complete a partially filled grid so that each row, column, and $3\times 3$ subgrid contains the digits $\{1,\dots,9\}$ exactly once. Difficulty can be controlled by reducing the number of givens and constructing instances with deep backtracking requirements, making Sudoku a strong testbed for evaluating reasoning stability under highly overlapping global constraints.

\paragraph{Formal Structure.}
Let
\[
\mathcal{I}=\{1,\dots,9\}\times\{1,\dots,9\}
\]
index the $81$ cells, and define the digit set $\mathcal{D}=\{1,\dots,9\}$.  
A given puzzle specifies a partial assignment \(g:\mathcal{I}\rightharpoonup\mathcal{D}\).  
A complete solution is a function
\[
f:\mathcal{I}\rightarrow\mathcal{D}
\]
that extends \(g\) and satisfies all Sudoku constraints.  
We define an instance:
\[
\mathcal{U}=(S,A,T,V,G^\star),
\]
where \(S\) is the set of complete grids and \(V\) is the global validity predicate.

\paragraph{State Space.}
A complete grid has:
\[
|S| = 9^{81},
\]
though only a vanishingly small fraction satisfy all constraints and match the givens.

\paragraph{Actions.}
In constructive search, an action assigns a digit to an empty cell:
\[
a=\text{Assign}((r,c)\leftarrow d),\quad d\in\mathcal{D}.
\]
The transition extends the partial grid by setting \(f(r,c)=d\).

\paragraph{Validity Constraints.}
A complete assignment \(f\) is valid iff:
\begin{enumerate}
    \item \textbf{Row uniqueness:} each row contains digits $\{1,\dots,9\}$ once.
    \item \textbf{Column uniqueness:} each column contains digits $\{1,\dots,9\}$ once.
    \item \textbf{Subgrid uniqueness:} each $3\times 3$ block contains digits $\{1,\dots,9\}$ once.
    \item \textbf{Agreement with givens:} \(f(r,c)=g(r,c)\) for all given cells.
\end{enumerate}
Thus:
\[
V(f)=1 \iff f \text{ satisfies all constraints}.
\]

\paragraph{Goal.}
The solution set is:
\[
G^\star=\{f\in S\mid V(f)=1\}.
\]

\paragraph{Example (Local Constraint Illustration).}
For row 1:
\[
[5,\ 3,\ \_,\ \_,\ 7,\ \_,\ \_,\ \_,\ \_],
\]
cell $(1,3)$ must exclude digits in the same row, same column, and same subgrid.  
Let:
\[
\mathrm{Cand}(1,3)=\mathcal{D}\setminus(\mathrm{Row}(1)\cup \mathrm{Col}(3)\cup \mathrm{Block}(1,3)).
\]
A right move selects \(d\in\mathrm{Cand}(1,3)\); assigning a forbidden digit yields an immediate row/column/block conflict.

\paragraph{Invalid Solutions.}
Typical invalid outcomes include:
\begin{itemize}
    \item duplicate digits in a row or column,
    \item subgrid violations,
    \item mismatch with givens,
    \item incomplete or partially filled grids.
\end{itemize}

\paragraph{Complexity Parameterization.}
We describe complexity via:
\[
\lambda=(q,\delta,\tau),
\]
where:
\begin{itemize}
    \item $q$ = number of empty cells (sparsity),
    \item $\delta$ = constraint tightness (e.g., average candidate-set size),
    \item $\tau$ = approximate backtracking depth induced by a solver.
\end{itemize}
A loose upper bound on the naive search space is:
\[
9^q,
\]
though constraint propagation prunes this substantially.

\paragraph{Complexity and Cognitive Load.}
Difficulty increases with fewer givens and stronger constraint coupling.  
If each empty cell has $b$ viable candidates on average, naive search explores roughly:
\[
O(b^q)
\]
states in the worst case.  
Dense global interactions cause long-range dependencies: a small local mistake can force inconsistencies many steps later.

\paragraph{Relevance for Reasoning Evaluation.}
Sudoku stresses:
\begin{itemize}
    \item global consistency under heavy constraint overlap,
    \item multi-step propagation of row/column/subgrid rules,
    \item long-range dependency management,
    \item avoidance of confident but invalid completions.
\end{itemize}
Failures at higher complexity often appear as duplicated digits, broken subgrid rules, misinterpreted givens, or incomplete final grids, exposing limits in maintaining global constraint satisfaction over large, interdependent structures.

\subsubsection{Rubik's Cube}

\begin{figure}
  \centering
  \includegraphics[width=0.9\columnwidth]{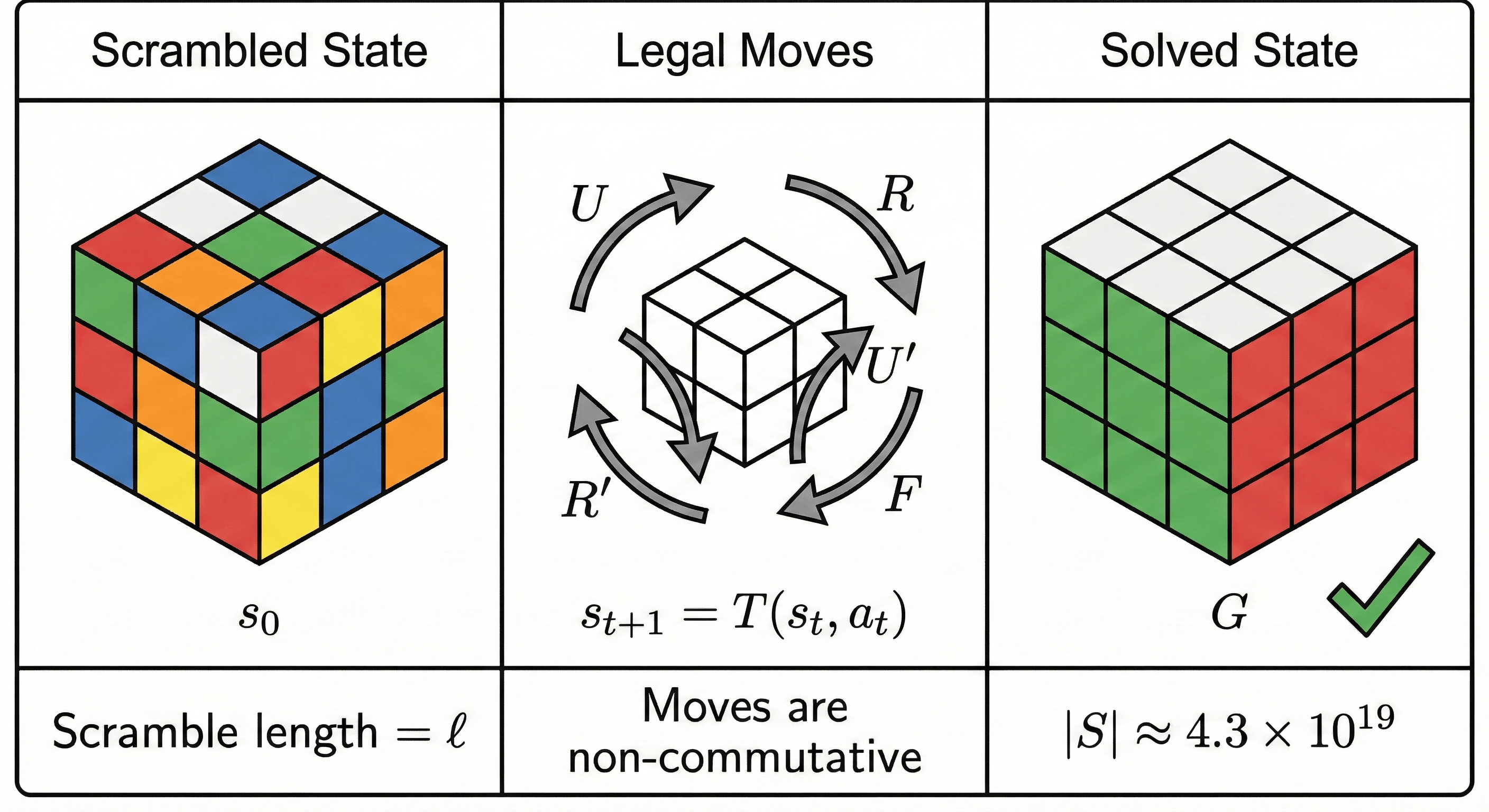}
  \caption{Rubik's Cube representation and face-turn notation used to describe legal moves.}
  \label{fig:rubiks_diagram}
\end{figure}

Rubik's Cube is a spatial planning and state-space search problem defined over a vast but finite discrete space. Starting from a scrambled configuration of the standard $3\times 3\times 3$ cube, the goal is to reach the solved state using a sequence of legal face turns. The task probes complexity-induced limits due to (i) non-commutative transitions, (ii) exponentially increasing horizon with scramble length, and (iii) strict validity constraints on both move syntax and final cube configuration.

\paragraph{Formal Structure.}
A Rubik's Cube instance is:
\[
\mathcal{B} = (S, A, T, V, G),
\]
where $S$ is the state space, $A$ the legal move set, $T$ the deterministic transition function, $V$ the validity predicate, and $G$ the solved-state set.

\paragraph{State Space.}
A state encodes the permutation and orientation of all movable pieces. The reachable state space for the standard cube is:
\[
|S| = 43{,}252{,}003{,}274{,}489{,}856{,}000 
\approx 4.3 \times 10^{19},
\]
large enough that exhaustive search is infeasible without strong algorithmic structure.

\paragraph{Move Set.}
We adopt the standard face-turn metric alphabet:
\[
A = \{U,D,L,R,F,B\} \times \{1,2,3\},
\]
where $1$ denotes a $90^\circ$ clockwise turn, $2$ a $180^\circ$ turn, and $3$ a $90^\circ$ counterclockwise (inverse) turn. Common notation uses primes, e.g., $U'$.

\paragraph{Transition Function.}
Given state $s_t$ and move $a_t \in A$:
\[
s_{t+1}=T(s_t,a_t),
\]
which applies the corresponding piece-permutation and orientation update. Moves are non-commutative:
\[
T(T(s,a),b) \neq T(T(s,b),a),
\]
complicating long-horizon reasoning.

\paragraph{Validity Constraints.}
Two conditions define validity of a move sequence $\pi=(a_1,\dots,a_T)$ from initial state $s_0$:
\[
V(s_0,\pi)=1 \iff 
\Big(a_t\in A\ \forall t\Big)\ \wedge\ T(s_0,\pi)\in G.
\]
Thus sequences must be syntactically legal and must solve the cube.

\paragraph{Goal Condition.}
The goal set $G$ contains one canonical solved state (or all solved orientations). A sequence is accepted if $T(s_0,\pi)\in G$.

\paragraph{Example (Move Inverses).}
Every legal move $a$ has an inverse $a^{-1}$:
\[
T(T(s,a),a^{-1})=s.
\]
For example, $R'$ is the inverse of $R$; $F2$ is self-inverse.

\paragraph{Right and Wrong Moves.}
A right move preserves solvability from the current state and follows legal syntax. Wrong moves include:
\begin{itemize}
    \item illegal tokens (e.g., \texttt{X3}, \texttt{RotateFaceRight}),
    \item syntactically valid sequences that fail to reach $G$,
    \item inconsistent claims of solved states without coherent sequences.
\end{itemize}
Due to non-commutativity and common temporary disordering, correct reasoning requires structured algorithms or accurate state tracking rather than greedy heuristics.

\paragraph{Complexity Parameterization.}
We define instance complexity via scramble length $\ell$ and structural constraints:
\[
\lambda = (\ell, \eta),
\]
where $\ell$ random moves generate the initial state, and $\eta$ characterizes scramble structure (e.g., avoidance of trivial cancellations). The optimal solution length $d(s)$ satisfies $d(s)\le \ell$ for such scrambles, though often $d(s)\ll \ell$ unless trivial inverses are forbidden.

\paragraph{Complexity and Cognitive Load.}
Let $N(\ell)$ denote the number of states within distance $\ell$ of solved. A standard approximation for early depths is:
\[
N(\ell)\approx 1+\sum_{t=1}^{\ell} b^t,
\]
where the effective branching factor $b\approx 13$ in the quarter-turn metric after excluding redundant inverses. Thus:
\[
N(\ell)=O(b^\ell),
\]
indicating exponential growth with scramble length.  
If each reasoning step is correct with probability $p$, then for a solution of length $L$:
\[
P_{\text{success}} \approx p^{\,L},
\]
so even small reasoning errors compound rapidly as $\ell$ increases.

\paragraph{Relevance for Reasoning Evaluation.}
Rubik's Cube stresses:
\begin{itemize}
    \item non-commutative symbolic state tracking,
    \item long-horizon planning under exponential branching,
    \item strict syntactic and semantic move validity,
    \item robust progression toward a single global target state.
\end{itemize}
Failure modes include illegal notation, inconsistent or cyclic move sequences, incorrect state claims, and confidently incorrect final solutions. As such, Rubik's Cube provides a stringent test of reasoning robustness under large-state, high-horizon, non-commutative search dynamics.

\subsection{Rationale for Problem Selection}

The selected puzzle families span complementary forms of discrete reasoning, enabling a unified yet multidimensional evaluation of model robustness under systematically increasing complexity. Each task isolates a distinct structural challenge that contemporary LRMs must overcome to exhibit scalable reasoning.

\paragraph{Recursive and Long-Horizon Planning.}
Tasks such as Tower of Hanoi require precise recursive decomposition and exact state tracking across exponentially long action sequences. These environments stress whether models can maintain internal invariants, propagate intermediate states correctly, and avoid compounding local errors over deep horizons.

\paragraph{Arithmetic and Invariant Preservation.}
The Water Jug problem focuses on deterministic arithmetic transitions governed by strict capacity constraints. Success requires maintaining numeric invariants and computing pour amounts exactly—capabilities that LLMs frequently approximate rather than compute algorithmically. This task probes the fragility of arithmetic reasoning under extended multi-step plans.

\paragraph{Global Constraint Satisfaction.}
Boolean SAT, Cryptarithmetic, Graph Coloring, and Sudoku impose dense, globally interacting logical constraints. These tasks differ in structure—Boolean clauses, carry-coupled digit mappings, adjacency conflicts, and overlapping grid constraints—but share a requirement for globally consistent assignments that cannot be satisfied through local pattern matching alone. They therefore expose weaknesses in global coherence and constraint propagation.

\paragraph{Local Validity Under Interaction.}
Checker Jumping and River Crossing combine local move validity with global feasibility, requiring the model to reason about reversible interactions, symmetric agents, and safety constraints at each step. Because a single invalid intermediate state invalidates the entire reasoning chain, these tasks provide sensitive diagnostics for violations of stepwise legality and state-tracking drift.

\paragraph{High-Dimensional Spatial Search.}
Rubik’s Cube introduces a combinatorially vast, non-commutative state space with structured but complex transition dynamics. Success demands reliable spatial state representation, strict syntactic adherence to move notation, and long-horizon planning under exponential branching—conditions under which LRMs often suffer abrupt failure.

\paragraph{Cross-Task Coverage and Transfer.}
Evaluating LRMs across these heterogeneous reasoning regimes allows us to distinguish genuine algorithmic competence from narrow heuristic adaptation. If models rely on superficial or task-specific heuristics, improvements should fail to transfer across domains with different structural dependencies. Conversely, systematic cross-task performance would provide evidence for more generalizable reasoning strategies.

\subsection{Deterministic Validation}

A core strength of our benchmark is the use of \emph{deterministic validation}, where every model output is checked by a strict, rule-based validator that enforces exact legality and goal conditions for each task. This removes ambiguity in evaluation, eliminates reliance on model-generated explanations or heuristic judgments, and isolates reasoning ability from linguistic fluency. Because solutions must satisfy all structural, arithmetic, or logical constraints under a fixed simulator, models cannot appear correct through persuasive text or partial reasoning; they must produce fully valid outputs. Deterministic validation therefore provides a clean, objective measure of reasoning robustness across complexity levels and enables precise identification of failure modes that would remain hidden under subjective or non-deterministic evaluations.

\subsubsection{Deterministic Validation for Tower of Hanoi}

We implement a deterministic validator that simulates the move sequence on three pegs and enforces strict legality at every step. Tower of Hanoi errors in LRMs typically arise from (i) moving a disk that is not actually at the top of the source peg and (ii) placing a larger disk on a smaller one. Deterministic validation ensures these failures are detected immediately and prevents superficially plausible but impossible trajectories.

\paragraph{Interface.}
For $n$ disks labeled $1$ (smallest) to $n$ (largest), the initial state is:
\[
A=[n,\dots,1],\qquad B=[],\qquad C=[].
\]
A model output is parsed as a list of moves $(d,s,t)$ with $s,t\in\{A,B,C\}$.

\paragraph{Validity Checks.}
A move $(d,s,t)$ is valid iff:
\begin{enumerate}
    \item it has correct format (a three-field tuple);
    \item peg $s$ is non-empty;
    \item $d$ equals the top disk on $s$ (hallucination detector);
    \item if peg $t$ is non-empty, then $\text{top}(t) > d$.
\end{enumerate}

\paragraph{Execution and Goal.}
Valid moves update the state by moving the top disk from $s$ to $t$.  
The solution passes if and only if the final state is:
\[
C=[n,n-1,\dots,1],\qquad A=B=[].
\]

\paragraph{Pseudocode.}

\begin{algorithm}
\caption{Tower of Hanoi Validator}
\begin{algorithmic}[1]
\Require $n$, move list $\pi$
\State $A\!\gets\![n,\dots,1],\ B\!\gets\![],\ C\!\gets\![]$
\For{each $(d,s,t)$ in $\pi$}
    \If{not a 3-tuple} \Return \texttt{Fail} \EndIf
    \If{$s$ empty} \Return \texttt{Fail} \EndIf
    \State $\textit{top}\gets$ top of $s$
    \If{$d\neq \textit{top}$} \Return \texttt{Fail} \EndIf
    \If{$t$ non-empty and top$(t)<d$} \Return \texttt{Fail} \EndIf
    \State pop $d$ from $s$; push $d$ to $t$
\EndFor
\If{$C=[n,\dots,1]$ and $A=B=[]$} \Return \texttt{Pass} \Else \Return \texttt{Fail} \EndIf
\end{algorithmic}
\end{algorithm}

\paragraph{Efficiency.}
Each move is checked and executed in $O(1)$ time; total runtime is $O(T)$ for $T$ moves.

\paragraph{Significance.}
This validator guarantees that only trajectories respecting exact Tower-of-Hanoi mechanics are accepted. It catches the most common LRM failure mode—\emph{disk-identity hallucination}—and ensures that success cannot be achieved through plausible text alone, but only through fully valid state transitions.

\subsubsection{Deterministic Validation for Checker Jumping}

We implement a deterministic validator that simulates Checker Jumping step-by-step and enforces all legality constraints, including direction rules, slide/jump distances, and final-state correctness. This prevents fluent but invalid trajectories—such as backward moves, jumps over empty space, or incorrect terminal boards—from being misclassified as successful reasoning.

\paragraph{Interface.}
For $n$ red and $n$ blue checkers, the board length is $2n+1$ with encoding:
\[
R\!\to\!1,\quad \_\!\to\!0,\quad B\!\to\!2.
\]
Initial and target states are:
\[
s_0=[1^n,0,2^n], \qquad s^\star=[2^n,0,1^n].
\]
A model output is parsed as moves $(c,i,j)$, where $c\in\{R,B\}$ and $i\!\to\!j$ is the attempted displacement.

\paragraph{Validity Constraints.}
A move $(c,i,j)$ is valid iff:
\begin{enumerate}
    \item $i,j$ are in bounds;
    \item $s[i]$ matches the encoded color of $c$;
    \item $s[j]=0$ (destination empty);
    \item motion is forward: $(j-i)\cdot\mathrm{dir}(c)\ge 0$, where $\mathrm{dir}(R)=+1$, $\mathrm{dir}(B)=-1$;
    \item $|j-i|\in\{1,2\}$;
    \item for jumps ($|j-i|=2$), the midpoint cell is non-empty.
\end{enumerate}

\paragraph{Execution and Goal.}
Valid moves update the board by swapping $s[i]\!\to\!s[j]$.  
The sequence passes only if the final board equals $s^\star$ exactly.

\paragraph{Pseudocode.}

\begin{algorithm}
\caption{Checker Jumping Validator}
\begin{algorithmic}[1]
\Require $n$, move list $\pi$
\State $s\gets [1]^n \Vert [0] \Vert [2]^n$
\For{each $(c,i,j)$ in $\pi$}
    \If{$i$ or $j$ out of bounds} \Return \texttt{Fail} \EndIf
    \State $\mathrm{col}\gets 1$ if $c{=}R$ else $2$
    \If{$s[i]\neq \mathrm{col}$} \Return \texttt{Fail} \EndIf
    \If{$s[j]\neq 0$} \Return \texttt{Fail} \EndIf
    \State $\mathrm{dir}\gets +1$ if $\mathrm{col}{=}1$ else $-1$
    \State $\Delta\gets j-i$
    \If{$\Delta\cdot\mathrm{dir}<0$} \Return \texttt{Fail} \EndIf
    \If{$|\Delta|=2$ and $s[(i+j)/2]=0$} \Return \texttt{Fail} \EndIf
    \If{$|\Delta|\notin\{1,2\}$} \Return \texttt{Fail} \EndIf
    \State $s[j]\gets s[i];\ s[i]\gets 0$
\EndFor
\State $s^\star\gets [2]^n \Vert [0] \Vert [1]^n$
\Return \texttt{Pass} if $s=s^\star$ else \texttt{Fail}
\end{algorithmic}
\end{algorithm}

\paragraph{Efficiency.}
Each move validates in $O(1)$; total runtime is $O(T)$ for $T$ steps plus an $O(n)$ final check.

\paragraph{Significance.}
Checker Jumping is prone to subtle but critical errors—backward motion, illegal jumps, mid-sequence drift—that LRMs often overlook. Deterministic validation ensures that only fully legal, goal-reaching trajectories count as correct, enabling accurate measurement of reasoning reliability under increasing instance complexity.

\subsubsection{Deterministic Validation for River Crossing}

We evaluate River Crossing solutions using a deterministic validator that simulates each boat trip and enforces \emph{capacity}, \emph{membership}, and \emph{safety} constraints at every step. This prevents fluent but formally invalid plans—e.g., moving passengers not on the bank, sending an empty boat, overloading capacity, or leaving unsafe configurations—from being counted as correct.

\paragraph{Interface.}
For $N$ couples and boat capacity $k$, individuals are labeled $\{A_i,a_i\}_{i=1}^N$. The initial state is:
\[
L_0=\{A_1,a_1,\dots,A_N,a_N\},\qquad R_0=\emptyset,
\]
with the boat on the left. A model output is parsed as a sequence of passenger sets:
\[
\pi=[P_1,\dots,P_T].
\]

\paragraph{Safety Rule.}
We adopt the classical jealous-husbands constraint:  
a wife $a_i$ may not be in the same location as any other husband unless $A_i$ is also present. For any group $G$,
\[
\mathrm{Safe}(G)=1 \iff \neg\exists i:\ a_i\in G,\ A_i\notin G,\ \mathrm{Men}(G)\neq\emptyset.
\]

\paragraph{Move Eligibility and Transition.}
A trip is legal only if:
\[
1\le |P_t|\le k,\qquad 
P_t \subseteq 
\begin{cases}
L_t & \text{if boat on left},\\
R_t & \text{if boat on right}.
\end{cases}
\]
On departure from the left,
\[
L_{t+1}=L_t\setminus P_t,\qquad R_{t+1}=R_t\cup P_t,
\]
and symmetrically when departing from the right.

\paragraph{Validation Logic.}
Each step must satisfy:
\begin{enumerate}
    \item \textbf{Capacity:} no empty trips, no overloads.
    \item \textbf{Membership:} all passengers originate from the departure bank.
    \item \textbf{Boat safety:} $\mathrm{Safe}(P_t)=1$.
    \item \textbf{Bank safety:} $\mathrm{Safe}(L_{t+1})=\mathrm{Safe}(R_{t+1})=1$.
\end{enumerate}
Any violation yields immediate \texttt{Fail}. The solution passes only if $L_T=\emptyset$.

\paragraph{Pseudocode.}

\begin{algorithm}
\caption{Deterministic River Crossing Verification}
\label{alg:river-verify}
\begin{algorithmic}[1]
\Require $N$, capacity $k$, moves $\pi=[P_1,\dots,P_T]$
\State $L\gets\{A_1,a_1,\dots,A_N,a_N\}$;\quad $R\gets\emptyset$
\State $\textit{side}\gets\texttt{Left}$
\Function{Safe}{$G$}
    \State $\mathrm{Men}\gets\{A_i\in G\}$
    \For{$a_i\in G$}
        \If{$A_i\notin G$ and $|\mathrm{Men}|>0$}
            \Return \texttt{False}
        \EndIf
    \EndFor
    \Return \texttt{True}
\EndFunction
\For{$P$ in $\pi$}
    \If{$|P|<1$ or $|P|>k$} \Return \texttt{Fail} \EndIf
    \State $B\gets L$ if $\textit{side}=\texttt{Left}$ else $R$
    \If{$P\nsubseteq B$} \Return \texttt{Fail} \EndIf
    \If{\Call{Safe}{$P$}=\texttt{False}} \Return \texttt{Fail} \EndIf
    \If{$\textit{side}=\texttt{Left}$}
        \State $L\gets L\setminus P$;\ $R\gets R\cup P$;\ $\textit{side}\gets\texttt{Right}$
    \Else
        \State $R\gets R\setminus P$;\ $L\gets L\cup P$;\ $\textit{side}\gets\texttt{Left}$
    \EndIf
    \If{\Call{Safe}{$L$}=\texttt{False}$\ \lor\ $ \Call{Safe}{$R$}=\texttt{False}}
        \Return \texttt{Fail}
    \EndIf
\EndFor
\Return \texttt{Pass} if $L=\emptyset$ else \texttt{Fail}
\end{algorithmic}
\end{algorithm}

\paragraph{Complexity.}
Each step performs membership checks over at most $k$ passengers and safety checks over groups of size $O(N)$, yielding total complexity:
\[
O(TN).
\]

\paragraph{Significance.}
River Crossing requires global safety preservation at every transition, not just locally valid moves. The deterministic validator filters out solutions that narratively ``sound correct'' but violate constraints mid-trajectory, enabling precise measurement of reasoning reliability under increasing $N$ and restricted capacity $k$.

\subsubsection{Deterministic Validation for Boolean SAT}

Boolean SAT admits a fully deterministic correctness check: a model-proposed assignment is either a complete mapping over all variables or it is not, and it either satisfies every clause or it does not. Our validator exploits this property by enforcing two explicit conditions: \textbf{(i)} completeness of the assignment over all referenced variables, and \textbf{(ii)} exact clause-level satisfaction of the CNF formula. This separates linguistic plausibility from formal correctness and ensures that only semantically valid solutions are counted as success.

\paragraph{Interface.}
A CNF formula is represented as:
\[
\Phi=\bigwedge_{i=1}^{m} C_i,\qquad 
C_i=\bigvee_{\ell\in C_i}\ell,
\]
where each literal $\ell\in\{\pm 1,\dots,\pm n\}$ encodes a variable or its negation, and $\mathrm{var}(\ell)=|\ell|$.  
A model output is parsed into a Boolean assignment:
\[
\alpha:\{1,\dots,n\}\rightarrow\{\texttt{True},\texttt{False}\}.
\]

\paragraph{Literal and Clause Semantics.}
A literal evaluates under $\alpha$ as:
\[
\mathrm{eval}(\ell,\alpha)=
\begin{cases}
\alpha(\mathrm{var}(\ell)) & \ell>0,\\
\neg\alpha(\mathrm{var}(\ell)) & \ell<0.
\end{cases}
\]
A clause is satisfied if any literal is true, and $\Phi$ is satisfied only if all clauses are satisfied:
\[
\mathrm{sat}(C_i,\alpha)=\bigvee_{\ell\in C_i}\mathrm{eval}(\ell,\alpha),\qquad
\mathrm{sat}(\Phi,\alpha)=\bigwedge_{i=1}^{m}\mathrm{sat}(C_i,\alpha).
\]

\paragraph{Validation Logic.}
For each clause, the validator checks:
\begin{enumerate}
    \item \textbf{Completeness:} every variable referenced in the clause appears in $\alpha$; missing variables cause immediate \texttt{Fail}.
    \item \textbf{Clause satisfaction:} if all literals evaluate to false, the validator returns \texttt{Fail} at the first violating clause.
\end{enumerate}
If all clauses pass, the assignment is returned as \texttt{Pass}.

\paragraph{Pseudocode.}

\begin{algorithm}
\caption{Deterministic CNF-SAT Verification}
\label{alg:sat-verify}
\begin{algorithmic}[1]
\Require Clauses $\mathcal{C}=\{C_1,\dots,C_m\}$, assignment $\alpha$
\For{$C_i$ in $\mathcal{C}$}
    \State $\textit{ok}\gets\texttt{False}$
    \For{$\ell$ in $C_i$}
        \State $v\gets|\ell|$
        \If{$v\notin\alpha$} \Return \texttt{Fail} \EndIf
        \State $b\gets\alpha(v)$
        \State $\textit{lit}\gets b$ if $\ell>0$ else $\neg b$
        \If{$\textit{lit}$} $\textit{ok}\gets\texttt{True}$; \textbf{break} \EndIf
    \EndFor
    \If{not $\textit{ok}$} \Return \texttt{Fail} \EndIf
\EndFor
\Return \texttt{Pass}
\end{algorithmic}
\end{algorithm}

\paragraph{Complexity.}
Let $L=\sum_i |C_i|$ be the total number of literals. Verification requires a single pass over all literals with $O(1)$ dictionary lookups, yielding:
\[
O(L),
\]
with frequent early exits in satisfiable instances.

\begin{sloppypar}
\paragraph{Significance.}
SAT's deterministic semantics make it an ideal testbed for explicit validity checking: any deviation—missing variables, contradictory assignments, or unsatisfied clauses—is unambiguously incorrect. The validator ensures that models receive credit only for fully correct satisfying assignments, enabling precise measurement of complexity-driven degradation rather than textual fluency or heuristic plausibility.
\end{sloppypar}

\subsubsection{Deterministic Validation for Cryptarithmetic}

Cryptarithmetic is highly susceptible to \emph{confident but invalid} outputs—digit reuse, missing letters, illegal leading zeros, or arithmetic mismatch. To ensure correctness independent of linguistic fluency, we implement a deterministic validator that parses the equation, enforces all structural constraints, and performs exact base-10 arithmetic on the decoded words.

\paragraph{Interface.}
An instance is an equation of the form:
\[
W_1 + \cdots + W_k = R,
\]
where each $W_i$ and $R$ is a letter-only token.  
A model returns a mapping:
\[
f:\Sigma \rightarrow \{0,\dots,9\},
\]
where $\Sigma$ is the set of distinct letters in the equation.

\paragraph{Validity Constraints.}
A mapping $f$ is accepted only if:

\begin{sloppypar}
\begin{enumerate}
    \item \textbf{Type/range validity:} each key is alphabetic and $f(\ell)\in\{0,\dots,9\}$.
    \item \textbf{Completeness:} every letter in the equation appears in $\mathrm{dom}(f)$.
    \item \textbf{Injectivity:} $f$ is one-to-one (no digit reuse).
    \item \textbf{No leading zeros:} for any word $w$, $f(w[1])\neq 0$.
    \item \textbf{Exact arithmetic equality:}
    \[
    \sum_{i=1}^{k} \mathrm{val}(W_i,f) \;=\; \mathrm{val}(R,f),
    \]
    where $\mathrm{val}(w,f)$ decodes $w$ into its base-10 integer.
\end{enumerate}
\end{sloppypar}

The global predicate $V(f)=1$ iff all constraints hold.

\paragraph{Pseudocode.}

\begin{algorithm}
\caption{Deterministic Cryptarithmetic Verification}
\label{alg:crypto-verify}
\begin{algorithmic}[1]
\Require Equation string $E$, mapping $f$
\State Normalize $E$; split on '=' into LHS and RHS; extract words with regex \texttt{[A-Za-z]+}
\If{RHS does not contain exactly one word} \Return \texttt{Fail} \EndIf
\State $(W_1,\dots,W_k)\gets$ LHS words;\; $R\gets$ RHS word
\ForAll{$(\ell\mapsto d)\in f$}
    \If{$\ell$ not alphabetic or $d\notin\{0,\dots,9\}$} \Return \texttt{Fail} \EndIf
\EndFor
\State $\Sigma \gets$ letters in all words
\If{$\Sigma\nsubseteq\mathrm{dom}(f)$} \Return \texttt{Fail} \EndIf
\If{$f$ not injective} \Return \texttt{Fail} \EndIf
\For{$w$ in $\{W_1,\dots,W_k,R\}$}
    \If{$f(w[1])=0$} \Return \texttt{Fail} \EndIf
\EndFor
\Function{WordToNum}{$w,f$}
    \State \Return integer from concatenation of $f$ applied to letters of $w$
\EndFunction
\State $S \gets \sum_i \Call{WordToNum}{W_i,f}$;\quad $T\gets\Call{WordToNum}{R,f}$
\If{$S=T$} \Return \texttt{Pass} \Else \Return \texttt{Fail} \EndIf
\end{algorithmic}
\end{algorithm}

\paragraph{Complexity.}
Let $L$ be the total number of letter occurrences. Validation requires $O(L + |\Sigma|)$ time, dominated by injectivity checks and digit-string conversion.

\paragraph{Significance.}
Cryptarithmetic solutions often appear coherent while violating a single structural rule or carry constraint. By enforcing injectivity, completeness, leading-digit legality, and exact arithmetic, this validator ensures that only fully correct mappings are counted as success—enabling precise measurement of reasoning degradation as symbol count, word length, and carry depth increase.

\subsubsection{Deterministic Validation for Graph Coloring}

Graph Coloring outputs frequently appear plausible while violating adjacency constraints in subtle ways—e.g., missing vertices, assigning colors to unknown nodes, or introducing a single edge conflict. To ensure correctness independent of linguistic fluency, we use a deterministic validator that enforces exact coverage and checks every edge for color conflicts.

\paragraph{Interface.}
A graph instance is $G=(V,E)$ with vertices $V$ and undirected edges $E\subseteq V\times V$.  
A model produces a coloring:
\[
f:V\rightarrow\mathcal{C},
\]
mapping each vertex to a color label (typically integers or symbols).  
The validator returns \texttt{Pass} or \texttt{Fail}, together with a list of issues.

\paragraph{Validity Constraints.}
A coloring is accepted only if:

\begin{enumerate}
    \item \textbf{Completeness:} every vertex has a color:
    \[
    V\subseteq\mathrm{dom}(f).
    \]
    \item \textbf{No extraneous assignments:} $\mathrm{dom}(f)\subseteq V$.
    \item \textbf{Adjacency legality:}
    \[
    \forall (u,v)\in E,\quad f(u)\neq f(v).
    \]
    \item \textbf{Edge sanity:} all edge endpoints must lie in $V$; unknown endpoints are recorded as errors.
    \item \textbf{Self-loop check (optional):} if self-loops are disallowed, any $(u,u)\in E$ triggers failure.
\end{enumerate}

Optional experimental constraints on color usage can be enforced:
\[
|\{f(v)\mid v\in V\}| \le k_{\max}, \qquad
|\{f(v)\mid v\in V\}| = k_{\mathrm{exact}}.
\]

\paragraph{Pseudocode.}

\begin{algorithm}
\caption{Deterministic Graph Coloring Verification}
\label{alg:graphcolor-verify}
\begin{algorithmic}[1]
\Require Vertices $V$, edges $E$, coloring $f$, optional $(k_{\max},k_{\mathrm{exact}})$
\State $\textit{issues}\gets[\,]$
\State $\textit{missing}\gets V\setminus\mathrm{dom}(f)$
\If{$\textit{missing}\neq\emptyset$} add ``Missing vertices'' to \textit{issues} \EndIf
\State $\textit{extra}\gets \mathrm{dom}(f)\setminus V$
\If{$\textit{extra}\neq\emptyset$} add ``Unknown vertices'' to \textit{issues} \EndIf
\State $\mathcal{C}_{\textit{used}}\gets\{f(v)\mid v\in V\cap\mathrm{dom}(f)\}$
\If{$k_{\max}$ set and $|\mathcal{C}_{\textit{used}}|>k_{\max}$} add ``Too many colors'' \EndIf
\If{$k_{\mathrm{exact}}$ set and $|\mathcal{C}_{\textit{used}}|\neq k_{\mathrm{exact}}$} add ``Wrong number of colors'' \EndIf
\ForAll{$(u,v)\in E$}
    \If{$u\notin V$ or $v\notin V$}
        \State add ``Edge references unknown vertex''; \textbf{continue}
    \EndIf
    \If{$u=v$ and self-loops disallowed} add ``Self-loop not allowed''; \textbf{continue} \EndIf
    \If{$u,v\in \mathrm{dom}(f)$ and $f(u)=f(v)$}
        \State add ``Conflict on edge $(u,v)$''
    \EndIf
\EndFor
\State \Return (\texttt{Pass} iff $\textit{issues}=\emptyset$, \textit{issues})
\end{algorithmic}
\end{algorithm}

\paragraph{Complexity.}
Coverage and color-usage checks take $O(n)$, and edge validation takes $O(m)$ for $n=|V|$ and $m=|E|$, giving total runtime:
\[
O(n+m).
\]

\begin{sloppypar}
\paragraph{Significance.}
Because Graph Coloring involves dense global constraints, errors may be small (a single conflicting edge) yet invalidate the entire solution. This validator ensures that only complete, conflict-free colorings are counted as success, enabling precise measurement of reasoning robustness as graph size and constraint density increase.
\end{sloppypar}

\subsubsection{Deterministic Validation for Water Jug}

Water Jug solutions often contain subtle errors—invalid operators, incorrect pour arithmetic, or final states that do not meet the goal despite fluent reasoning. To ensure correctness independent of narrative explanations, we use a deterministic validator that executes the model’s action sequence on an exact two-jug simulator, enforcing strict operator legality and state transitions.

\paragraph{Interface.}
A problem instance is defined by jug capacities and a target:
\[
c_A,\, c_B,\, g \in \mathbb{N}.
\]
A state is $(x,y)$ with
\[
0 \le x \le c_A,\qquad 0 \le y \le c_B,
\]
and the initial state is $(0,0)$.  
The model output is parsed as a list of single-token steps:
\[
\pi = [[a_1],\dots,[a_T]],
\]
where each $a_t$ must be a valid operator.

\paragraph{Legal Actions.}

The validator recognizes the fixed operator set:
\begin{align*}
\mathcal{A} = \{&\texttt{Fill A}, \texttt{Fill B}, \texttt{Empty A}, \\
                &\texttt{Empty B}, \texttt{Pour A to B}, \texttt{Pour B to A}\}.
\end{align*}
Any token outside $\mathcal{A}$ triggers immediate failure.

\paragraph{Deterministic Transitions.}
For $s_t=(x,y)$:
\[
\begin{aligned}
&\texttt{Fill A}: (c_A,y),\qquad
\texttt{Fill B}: (x,c_B),\\
&\texttt{Empty A}: (0,y),\qquad
\texttt{Empty B}: (x,0),\\
&\texttt{Pour A to B}: \delta = \min(x,c_B-y),\ (x-\delta,\,y+\delta),\\
&\texttt{Pour B to A}: \delta = \min(y,c_A-x),\ (x+\delta,\,y-\delta).
\end{aligned}
\]

\paragraph{Goal Check.}
After executing all actions, the solution is accepted iff:
\[
x_T = g \quad\text{or}\quad y_T = g.
\]

\begin{sloppypar}
\paragraph{Common Failure Modes.}
The validator rejects:
(i) malformed steps,  
(ii) unknown operators,  
(iii) incorrect pour arithmetic (avoided by deterministic updates),  
(iv) final states not reaching the target.
\end{sloppypar}

\paragraph{Pseudocode.}

\begin{algorithm}
\caption{Deterministic Water Jug Verification}
\label{alg:waterjug-verify}
\begin{algorithmic}[1]
\Require Capacities $(c_A,c_B)$, goal $g$, steps $\pi=[[a_t]]_{t=1}^T$
\State $(x,y)\gets(0,0)$
\For{$t=1$ to $T$}
    \If{$\pi[t]$ not a singleton list} \Return \texttt{Fail} \EndIf
    \State $a\gets\pi[t][0]$
    \If{$a=\texttt{Fill A}$}         \State $x\gets c_A$
    \ElsIf{$a=\texttt{Fill B}$}      \State $y\gets c_B$
    \ElsIf{$a=\texttt{Empty A}$}     \State $x\gets 0$
    \ElsIf{$a=\texttt{Empty B}$}     \State $y\gets 0$
    \ElsIf{$a=\texttt{Pour A to B}$}
        \State $\delta\gets\min(x,\,c_B-y)$; $x\gets x-\delta$; $y\gets y+\delta$
    \ElsIf{$a=\texttt{Pour B to A}$}
        \State $\delta\gets\min(y,\,c_A-x)$; $y\gets y-\delta$; $x\gets x+\delta$
    \Else
        \Return \texttt{Fail} \Comment{Invalid operator}
    \EndIf
\EndFor
\If{$x=g$ or $y=g$} \Return \texttt{Pass} \Else \Return \texttt{Fail} \EndIf
\end{algorithmic}
\end{algorithm}

\paragraph{Complexity.}
Each action is processed in constant time, giving total runtime:
\[
O(T).
\]

\paragraph{Significance.}
Because Water Jug requires exact arithmetic transitions, models frequently produce sequences that look correct but violate operator semantics or reach the wrong final state. Deterministic validation ensures that only physically realizable trajectories achieving the exact target are counted as success, enabling precise measurement of reasoning robustness as capacities and goal difficulty increase.

\subsubsection{Deterministic Validation for Sudoku (General $N\times N$, Jigsaw, and Rectangular Regions)}

Sudoku outputs often appear plausible while containing subtle structural errors—mis-sized grids, invalid digit ranges, altered clues, or malformed region definitions that invalidate the entire solution. To enforce correctness independently of natural-language reasoning, we implement a deterministic validator that (i) checks input well-formedness, (ii) constructs the region partition (standard blocks or arbitrary jigsaw regions), (iii) enforces the valid digit domain (with optional blanks), (iv) verifies clue consistency, and (v) detects duplicates across rows, columns, and regions. This unified framework supports arbitrary grid sizes and region systems, enabling rigorous evaluation far beyond the standard $9\times 9$ setting.

\paragraph{Interface.}
An instance consists of:
\begin{align*}
\texttt{grid} &\in \mathbb{Z}^{N\times N}, \\
\texttt{puzzle} &\in \mathbb{Z}^{N\times N}\ (\text{optional}), \\
\texttt{regions} &\in \mathbb{Z}^{N\times N}\ (\text{optional}), \\
\texttt{block\_shape} &= (b_r,b_c)\ (\text{optional})
\end{align*}

The validator returns \texttt{Pass} or \texttt{Fail} together with a list of issues.

\paragraph{Value Domain.}
Digits must lie in
\[
\mathcal{D}=\{1,\dots,N\},
\]
optionally extended with $0$ to denote blanks. If \texttt{require\_complete} is enabled, blanks are forbidden.

\paragraph{Region Systems.}
A valid Sudoku requires a partition of the $N^2$ cells into exactly $N$ regions of size $N$. The validator supports:
(i) \textbf{Jigsaw regions} via an explicit region-ID grid;
(ii) \textbf{Rectangular blocks} via \texttt{block\_shape} with $b_r b_c = N$;
(iii) \textbf{Inferred square blocks} when $N$ is a perfect square and no region system is provided.
Any discrepancy in region count, size, or shape triggers immediate failure.

\paragraph{Clue Consistency.}
If a puzzle grid is provided, every nonzero clue must be preserved:
\[
\texttt{puzzle}[r,c]\neq 0 \;\Rightarrow\; \texttt{grid}[r,c]=\texttt{puzzle}[r,c].
\]

\paragraph{Duplicate Checks.}
Rows, columns, and regions must each contain no repeated digits from $\mathcal{D}$; blanks ($0$) are ignored. Thus the validator examines:
\[
\text{rows }1..N,\quad
\text{columns }1..N,\quad
\text{regions }1..N.
\]

\paragraph{Algorithm Summary.}
Table~\ref{tab:sudoku-any-validator-steps} summarizes the deterministic pipeline.

\begin{table}[!htbp]
\caption{General Sudoku validator pipeline.}
\label{tab:sudoku-any-validator-steps}
\renewcommand{\arraystretch}{1.1}
\begin{tabular*}{\linewidth}{@{} l p{0.85\linewidth} @{}}
\toprule
\textbf{Step} & \textbf{Check / Action} \\
\midrule
0 & Validate shapes of \texttt{grid}, \texttt{puzzle}, \texttt{regions} \\
1 & Build region partition (jigsaw, rectangular, or inferred square blocks) \\
2 & Verify region count $=N$ and each region has exactly $N$ cells \\
3 & Validate digit domain; enforce completeness if required \\
4 & Enforce puzzle clues if provided \\
5 & Check each row for duplicates (ignore zeros) \\
6 & Check each column for duplicates (ignore zeros) \\
7 & Check each region for duplicates (ignore zeros) \\
\bottomrule
\end{tabular*}
\end{table}

\paragraph{Pseudocode.}
A concise validator is given below.

\begin{algorithm}
\caption{Deterministic Sudoku Verification (Any $N$, Blocks or Jigsaw)}
\label{alg:sudoku-any-verify-short}
\begin{algorithmic}[1]
\Require \texttt{grid}, optional \texttt{puzzle}, \texttt{regions}, \texttt{block\_shape}; flags (\texttt{allow\_zeros}, \texttt{require\_complete})
\State Verify \texttt{grid} is $N\times N$; check shapes of optional inputs
\State Construct region partition:
\begin{itemize}
    \item \texttt{regions}: group by IDs; require $N$ regions of size $N$
    \item \texttt{block\_shape}: tile into $b_r\times b_c$ blocks with $b_r b_c=N$
    \item else: infer $\sqrt{N}\times\sqrt{N}$ blocks if $N$ is a perfect square
\end{itemize}
\State For each cell: value $\in\{1..N\}$ (and $0$ if allowed); reject blanks if \texttt{require\_complete}
\If{\texttt{puzzle} provided} enforce all nonzero clues
\EndIf
\Function{CheckUnit}{$U$} ignore zeros, fail if any digit repeats \EndFunction
\For{each row, column, region} \Call{CheckUnit}{unit}
\EndFor
\State \Return \texttt{Pass} if no issues; otherwise \texttt{Fail}
\end{algorithmic}
\end{algorithm}

\paragraph{Complexity.}
All steps operate over $N^2$ cells or $N$ units of size $N$, giving overall time complexity:
\[
O(N^2).
\]

\paragraph{Significance.}
Generalized Sudoku introduces structural pitfalls absent in the standard $9\times 9$ case, including invalid region systems, incorrect digit domains, and subtle clue violations. By explicitly validating region partitions, digit ranges, and all uniqueness constraints, this deterministic validator ensures that only fully correct solutions are accepted—crucial for evaluating reasoning robustness under increasing puzzle size and constraint density.

\subsubsection{Deterministic Validation for Rubik's Cube}

Rubik’s Cube solutions often contain superficially coherent but formally invalid elements—malformed cube states, illegal move tokens, or sequences that do not actually transform the start state into the target. To separate genuine correctness from linguistic plausibility, we implement a deterministic validator that (i) checks cube-state sanity, (ii) enforces a strict move alphabet, (iii) simulates the cube under exact face-turn permutations, and (iv) verifies that the final configuration matches the declared target state.

\paragraph{Input/Output Interface.}
An instance includes a cube size $N$ (typically $3$) and two facelet strings:
\[
s_0,\ s^\star \in \Sigma^{6N^2},
\]
where $\Sigma$ is the face-color alphabet (e.g., $\{W,R,G,Y,O,B\}$).  
A model output is a whitespace-separated move string:
\[
\pi = \texttt{"R U R' U'"}.
\]
The validator returns \texttt{Pass} or \texttt{Fail} along with a list of issues.

\paragraph{Move Alphabet and Parsing.}
We adopt the standard quarter-turn metric:
\[
\mathcal{M}=\{f,\ f',\ f2 \mid f\in\{U,D,L,R,F,B\}\}.
\]
The validator splits the move string on whitespace; any token not in $\mathcal{M}$ causes immediate failure.

\paragraph{State Validity.}
Before simulation, we enforce:
\begin{enumerate}
    \item \textbf{Length:} $|s_0|=|s^\star|=6N^2$.
    \item \textbf{Alphabet:} all characters lie in $\Sigma$.
    \item \textbf{Color counts (optional):} each color appears exactly $N^2$ times in both states.
\end{enumerate}
These checks detect malformed or impossible cube configurations.

\paragraph{Deterministic Transition Semantics.}
Each move token is mapped to a fixed permutation of the $6N^2$ facelets. For $N{=}3$, these are the standard quarter-turn permutations applied as:
\begin{itemize}
    \item rotate the turned face $90^\circ$,
    \item cycle the affected edge strips on adjacent faces,
    \item flatten back to a $6N^2$ string.
\end{itemize}
Let $T(s,m)$ denote this transition. Inverses and double turns are implemented as:
\[
T(s,f') = T^{-1}(s,f), \qquad
T(s,f2)=T(T(s,f),f).
\]

\paragraph{Goal Condition.}
After simulating all moves:
\[
s_T = s^\star
\]
is required for acceptance; otherwise, the validator reports a final-state mismatch.

\paragraph{Common Failure Modes.}
The validator detects:
\begin{itemize}
    \item invalid cube lengths or non-alphabet characters,
    \item inconsistent color counts (impossible cube states),
    \item illegal move tokens (e.g., \texttt{X}, \texttt{R3}),
    \item syntactically correct move traces that do not reach the target state.
\end{itemize}

\paragraph{Algorithm Summary.}
Table~\ref{tab:rubik-validator-steps} shows the validation pipeline.

\begin{table}
\centering
\caption{Rubik's Cube validator pipeline (deterministic simulation).}
\label{tab:rubik-validator-steps}
\begin{tabular}{p{1.4cm} p{5.8cm}}
\hline
\textbf{Step} & \textbf{Check / Action} \\
\hline
0 & Verify $N$ and lengths $|s_0|=|s^\star|=6N^2$ \\
1 & Alphabet check on $s_0$, $s^\star$ \\
2 & (Optional) Verify color counts ($N^2$ per color) \\
3 & Tokenize move string into $m_1\dots m_T$ \\
4 & Enforce each $m_t\in\mathcal{M}$ \\
5 & Simulate $s\leftarrow T(s,m_t)$ for all $t$ \\
6 & Accept iff $s=s^\star$ \\
\hline
\end{tabular}
\end{table}

\paragraph{Pseudocode.}
\begin{algorithm}
\caption{Deterministic Rubik's Cube Verification}
\label{alg:rubik-verify}
\begin{algorithmic}[1]
\Require $N$, start state $s_0$, move string $M$, target $s^\star$, colors $\Sigma$
\If{$|s_0|\neq 6N^2$ or $|s^\star|\neq 6N^2$} \Return \texttt{Fail} \EndIf
\If{any character in $s_0$ or $s^\star$ $\notin\Sigma$} \Return \texttt{Fail} \EndIf
\If{(optional) any color count $\neq N^2$} \Return \texttt{Fail} \EndIf
\State $\pi\gets$ tokenize $M$
\For{each $m\in\pi$}
    \If{$m\notin\mathcal{M}$} \Return \texttt{Fail} \EndIf
\EndFor
\State $s\gets s_0$
\For{each $m\in\pi$}
    \State $s\gets T(s,m)$
\EndFor
\If{$s=s^\star$} \Return \texttt{Pass} \Else \Return \texttt{Fail} \EndIf
\end{algorithmic}
\end{algorithm}

\paragraph{Computational Complexity.}
Each face turn affects $O(N^2)$ facelets, giving:
\[
O(TN^2),
\]
which is effectively $O(T)$ for the standard $3\times 3\times 3$ cube.

\paragraph{Significance.}
Rubik’s Cube is highly sensitive to syntactic and semantic errors: one invalid token or incorrect state update invalidates the entire solution. By performing strict token validation and exact simulation of face-turn mechanics, our validator ensures that a model receives credit only when its move sequence is \emph{formally correct} and genuinely transforms $s_0$ into $s^\star$. This is essential for isolating true reasoning failures as scramble complexity and solution length increase.

\subsection{Complexity Regime Design}

Our complexity regime defines a unified difficulty ladder across all puzzle environments, using levels $L_1$–$L_5$ (and $L_{6+}$ for selected tasks) as summarized in below. Each level increases reasoning demand while preserving task semantics, enabling controlled, cross-task comparison. At the low end ($L_1$–$L_2$), instances require short reasoning chains and minimal constraint interaction, serving primarily as sanity checks. Mid-range levels ($L_3$–$L_4$) introduce multi-step dependencies, longer planning horizons, and denser constraint propagation—forming the core regime in which systematic reasoning failures begin to emerge. High-complexity levels ($L_5$ and $L_{6+}$) exceed typical LRM capacity, involving deep recursion, large branching factors, or long search horizons that frequently induce abrupt collapse. Although individual tasks parameterize complexity through different controls (e.g., disks, clauses, letters, graph size, jug capacities, Sudoku sparsity, scramble length), these levels are aligned to represent comparable increases in cognitive load across domains.

\begin{description}
    \item[$L_1$] \textbf{Minimal complexity:} Short reasoning chains; rule-understanding and formatting sanity check.
    \item[$L_2$] \textbf{Low complexity:} Small search space; shallow state tracking; limited constraint interactions.
    \item[$L_3$] \textbf{Intermediate complexity:} Multi-step global reasoning; moderate constraint coupling; systematic errors begin to appear.
    \item[$L_4$] \textbf{High intermediate complexity:} Long-horizon planning; dense constraint propagation; requires consistent multi-step reasoning.
    \item[$L_5$] \textbf{High complexity:} Beyond typical LRM capacity; deep recursion, large branching factors, long search horizons; abrupt collapse modes common.
    \item[$L_{6+}$] \textbf{Extreme regime:} Used selectively; exceeds feasible symbolic reasoning depth; used primarily for failure-mode diagnostics.
\end{description}

\section{Experimental Protocol}

This section describes the experimental setup used to evaluate reasoning behavior across models and tasks. We detail the models evaluated, prompting strategy, decoding and budget constraints, validation methodology, and evaluation criteria. Our protocol is designed to ensure fairness, reproducibility, and task-agnostic comparison across increasing complexity levels.

\subsection{Models Evaluated}

We evaluate a broad spectrum of contemporary large language models spanning both \emph{closed-source} frontier systems and state-of-the-art \emph{open-source} releases. Closed-source models include commercial reasoning-enabled variants (e.g., OpenAI, Anthropic, Google) accessed through their official APIs, while open-source models include leading community-driven LLMs such as LLaMA, DeepSeek, Mixtral, Qwen, and other publicly released checkpoints. When available, we invoke each model’s designated ``thinking'' or reasoning-enhanced mode to ensure consistent evaluation of explicit reasoning capabilities. All models are queried independently under identical prompts and complexity settings, without fine-tuning or task-specific adaptation. Thus, all reported results reflect strictly zero-shot general-purpose reasoning performance across the benchmark.

\subsection{Prompting Strategy}

All puzzle environments are evaluated using structured, zero-shot prompts that explicitly specify (i) the problem rules, (ii) the input structure, and (iii) the required machine-parseable output format. Although models are instructed to reason step-by-step, correctness is determined solely from the final structured output consumed by the deterministic validators. This removes ambiguity and ensures that reasoning traces do not influence correctness judgments.

Prompts are \emph{complexity-invariant}: each task family uses a single prompt template across all complexity regimes ($L_1$–$L_5$ and beyond). Only instance-specific parameters (e.g., number of disks, graph edges, clause sets, scramble sequence) are substituted per test case. This design prevents confounding effects from prompt rewording, reduces benchmark contamination, and ensures that variation in performance is attributable only to changes in task complexity.

Each prompt enforces a strict output schema (e.g., lists of triples, boolean dictionaries, boat-passenger lists, coordinate pairs, or WCA move sequences). Any syntactic deviation—malformed lists, missing fields, invalid tokens—results in immediate validator failure. This creates a uniform evaluation environment across all models and prevents subjective interpretation of natural-language responses.

Below we show the standardized prompt templates used for each puzzle environment. Each template is displayed alongside a corresponding illustrative figure.

\begin{figure}
\centering
\includegraphics[width=0.95\columnwidth]{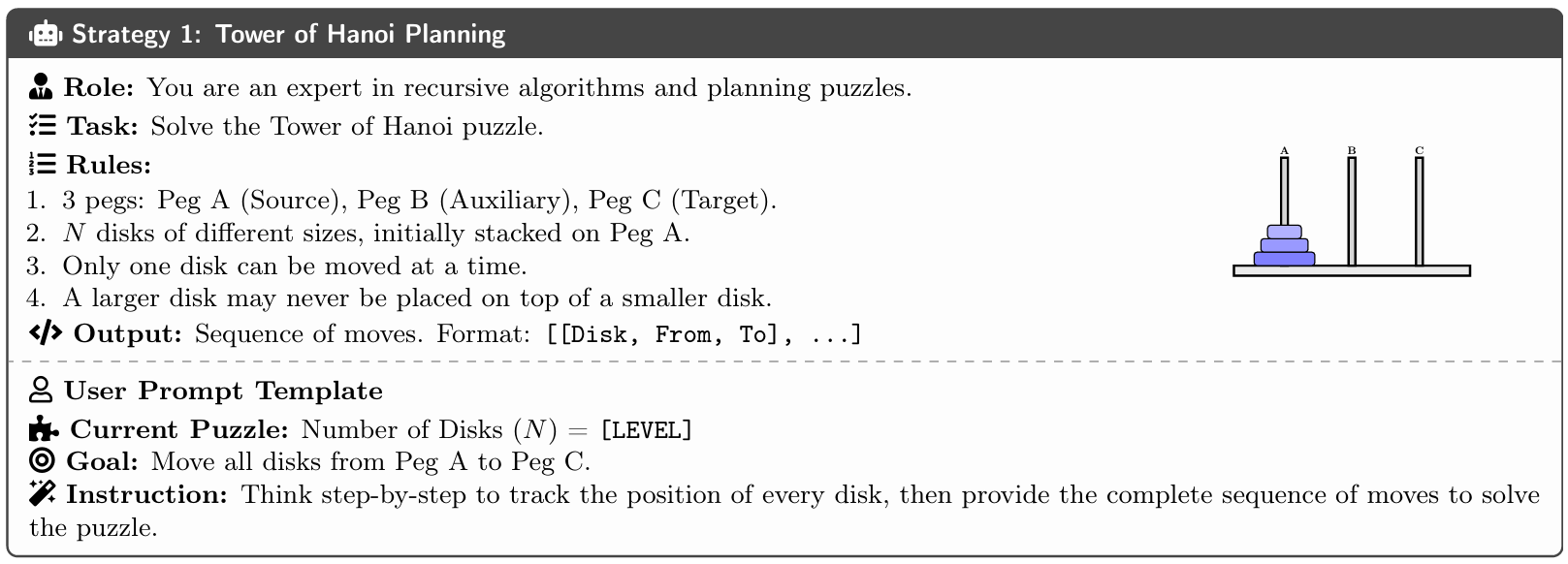}
\caption{Prompt template for the Tower of Hanoi puzzle, including role specification, rule definitions, and required move-list output format.}
\label{fig:prompt-hanoi}
\end{figure}

\begin{figure}
\centering
\includegraphics[width=0.95\columnwidth]{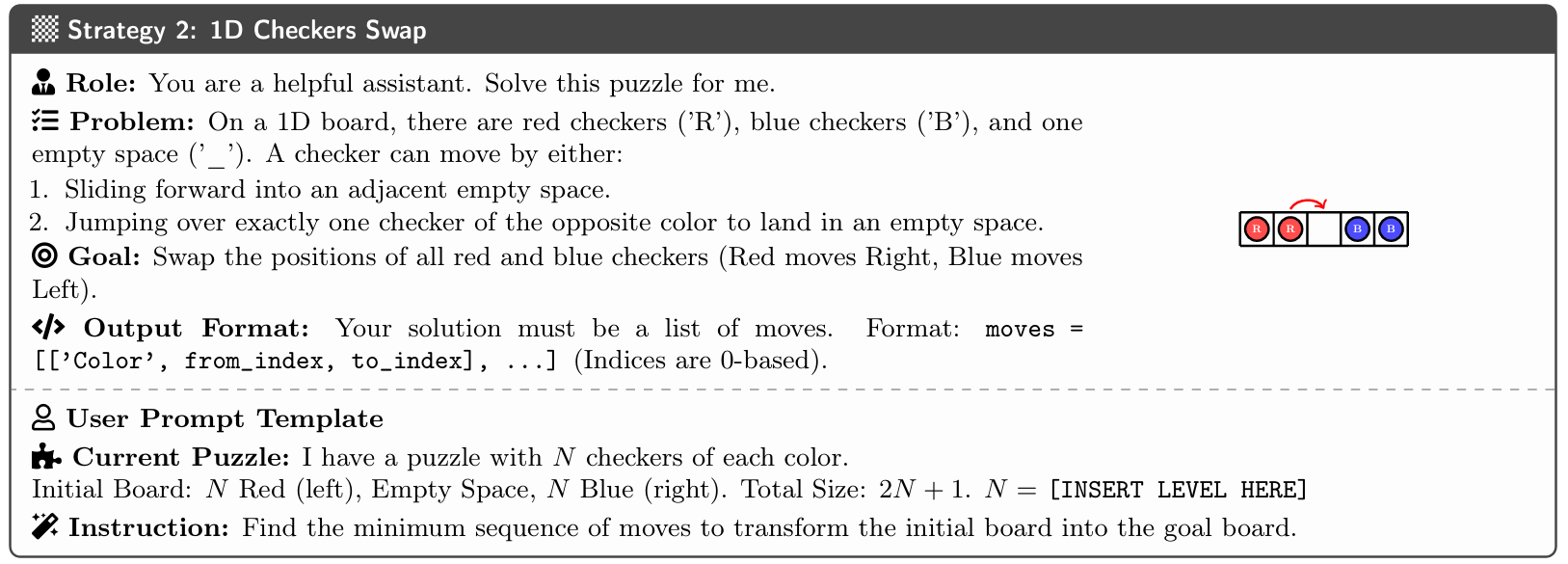}
\caption{Prompt template for the 1D Checkers Swap puzzle, specifying legal moves and an index-based output schema.}
\label{fig:prompt-checkers}
\end{figure}

\begin{figure}
\centering
\includegraphics[width=0.95\columnwidth]{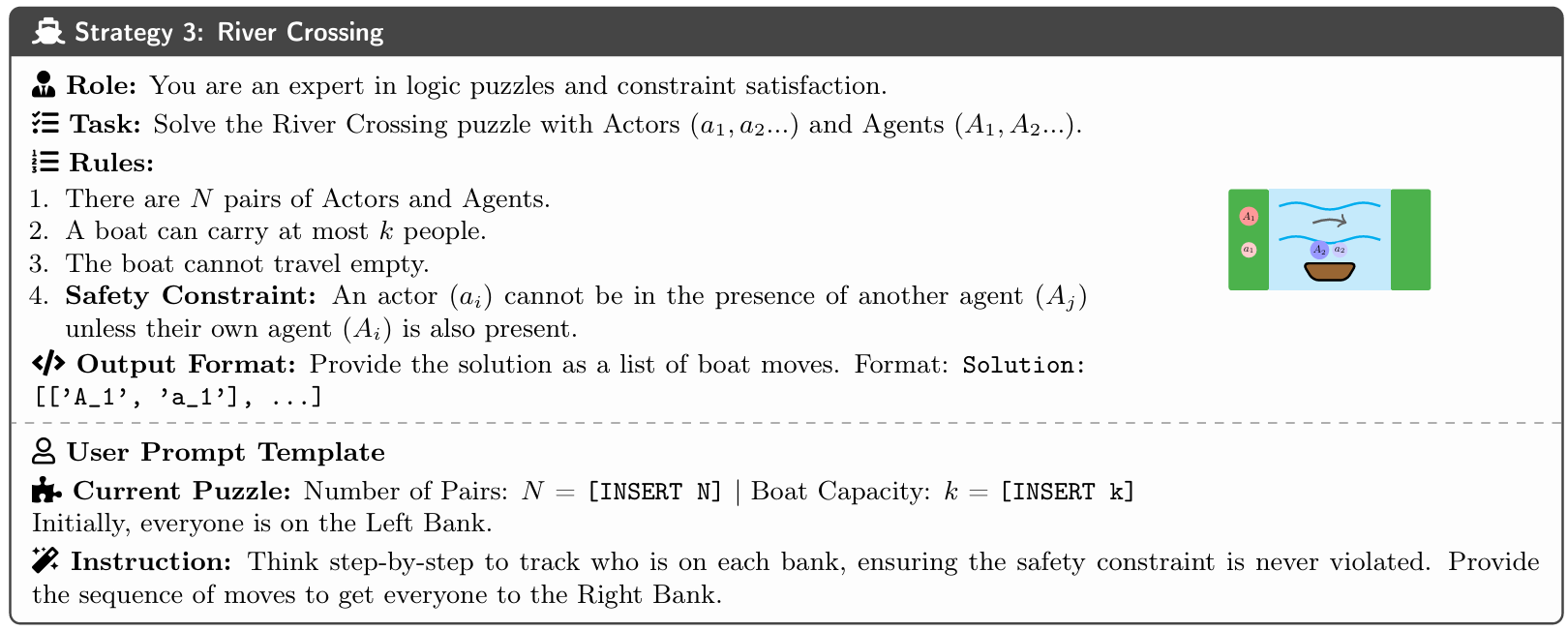}
\caption{Prompt structure for the River Crossing puzzle, emphasizing capacity limits and the jealous-husbands safety constraint.}
\label{fig:prompt-river}
\end{figure}

\begin{figure}
\centering
\includegraphics[width=0.95\columnwidth]{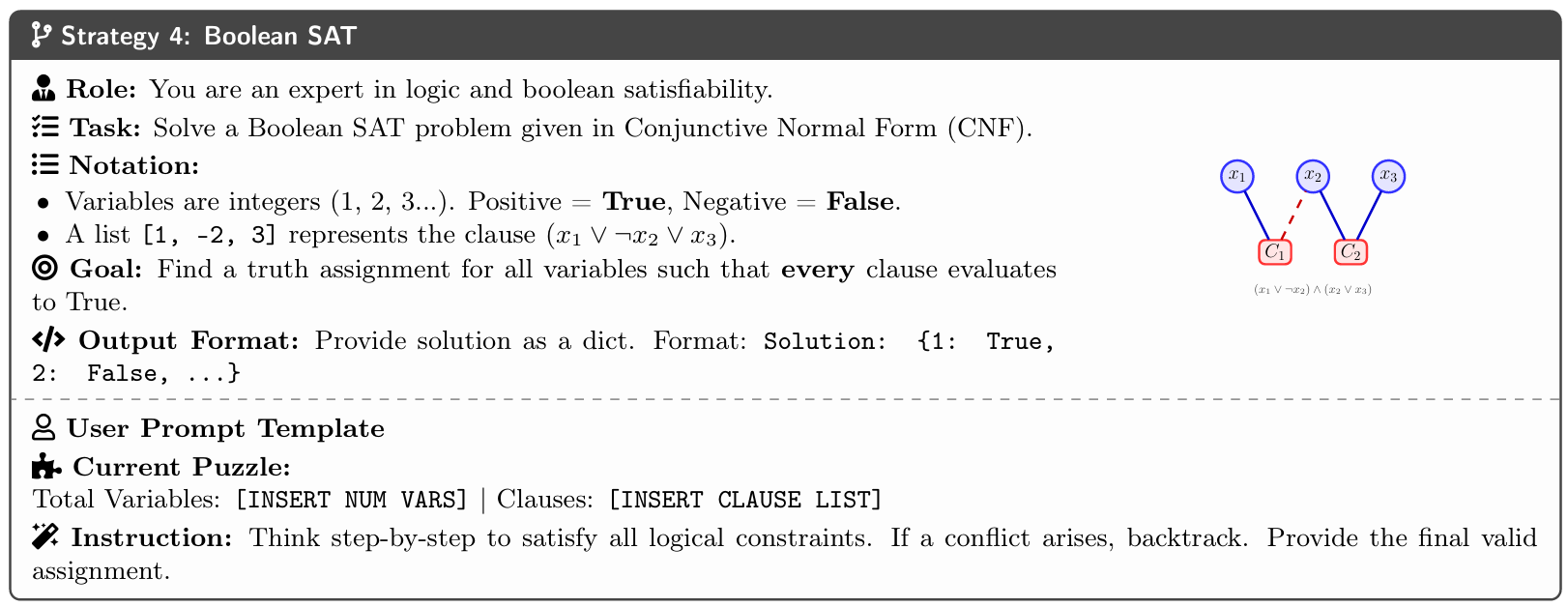}
\caption{Prompt template for Boolean SAT, showing the CNF array encoding and dictionary-based assignment output.}
\label{fig:prompt-sat}
\end{figure}

\begin{figure}
\centering
\includegraphics[width=0.95\columnwidth]{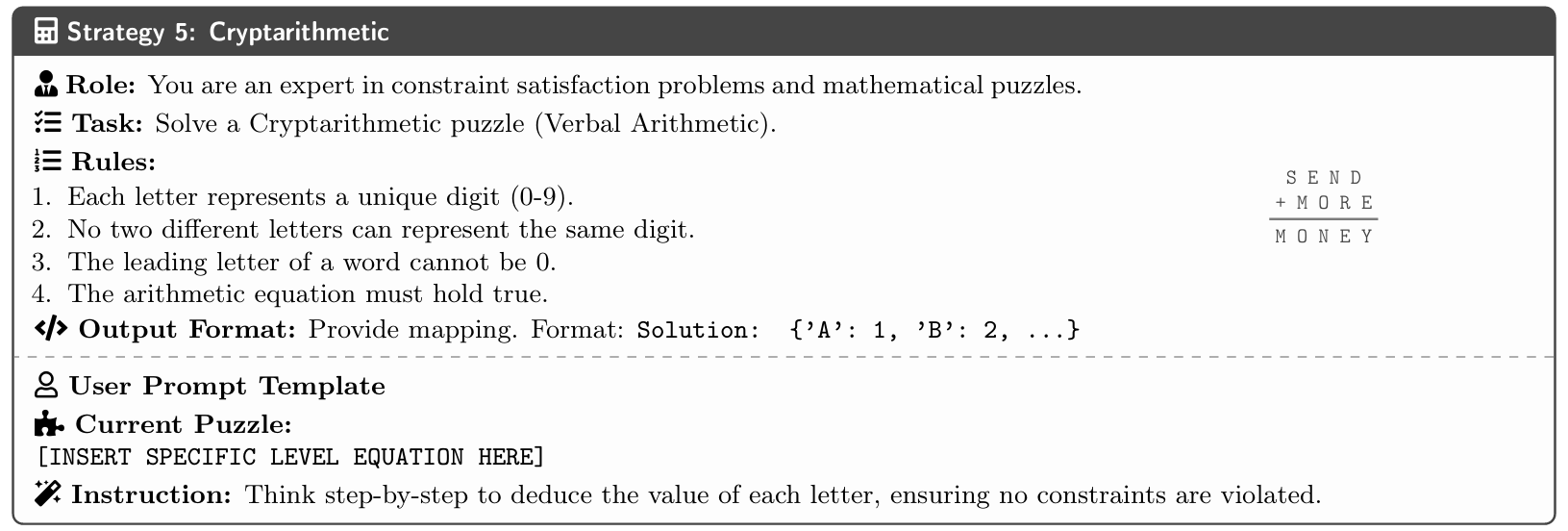}
\caption{Prompt template for Cryptarithmetic puzzles, outlining digit-uniqueness, leading-zero constraints, and mapping output format.}
\label{fig:prompt-crypto}
\end{figure}

\begin{figure}
\centering
\includegraphics[width=0.95\columnwidth]{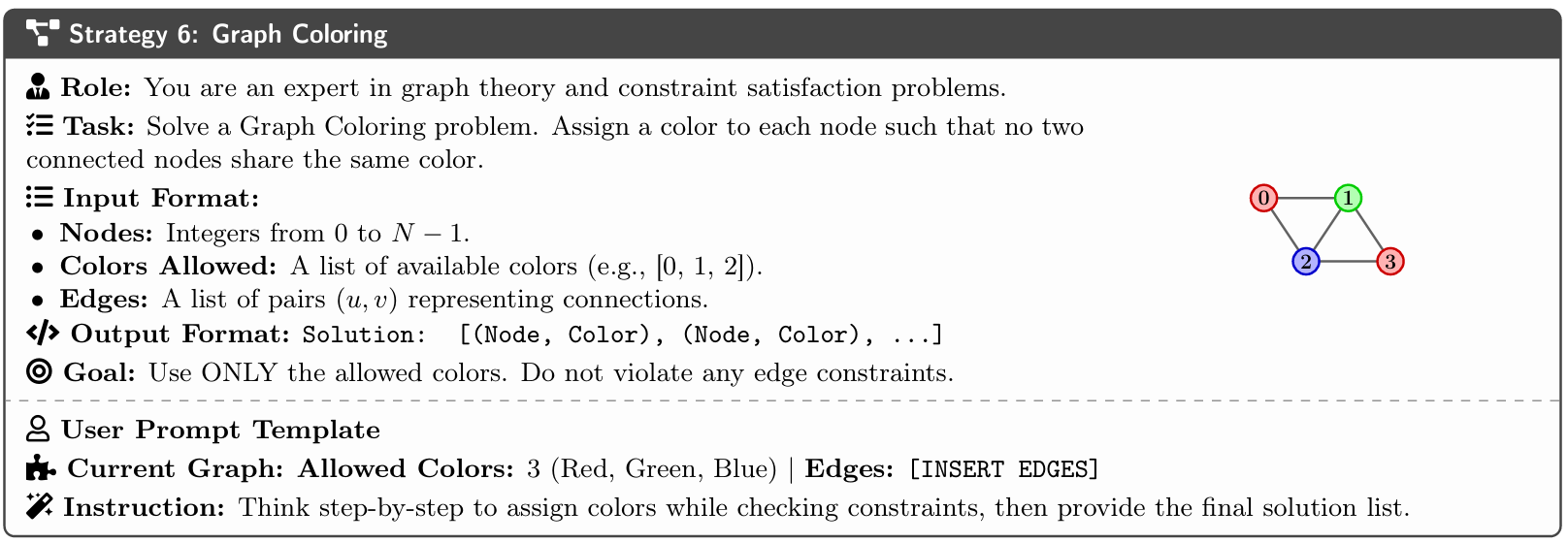}
\caption{Prompt template for Graph Coloring, specifying allowed colors, node sets, and adjacency constraints.}
\label{fig:prompt-graphcolor}
\end{figure}

\begin{figure}
\centering
\includegraphics[width=0.95\columnwidth]{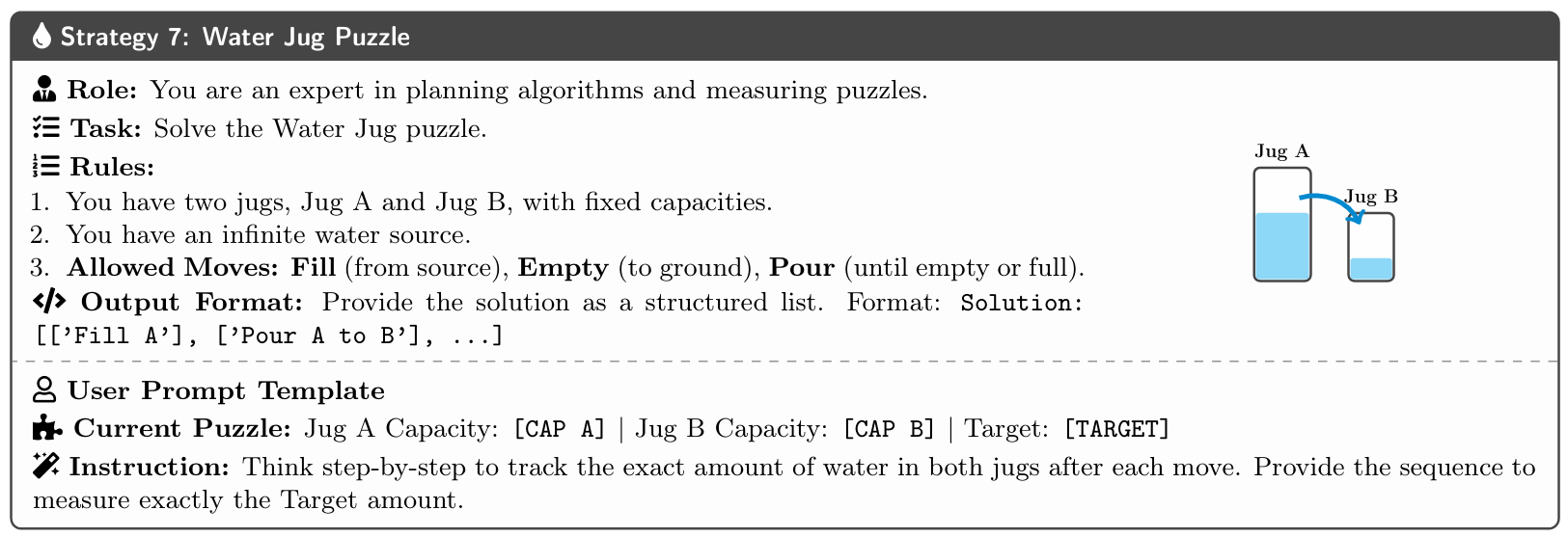}
\caption{Prompt template for the Water Jug puzzle, including the fixed action space and structured move-list format.}
\label{fig:prompt-waterjug}
\end{figure}

\begin{figure}
\centering
\includegraphics[width=0.95\columnwidth]{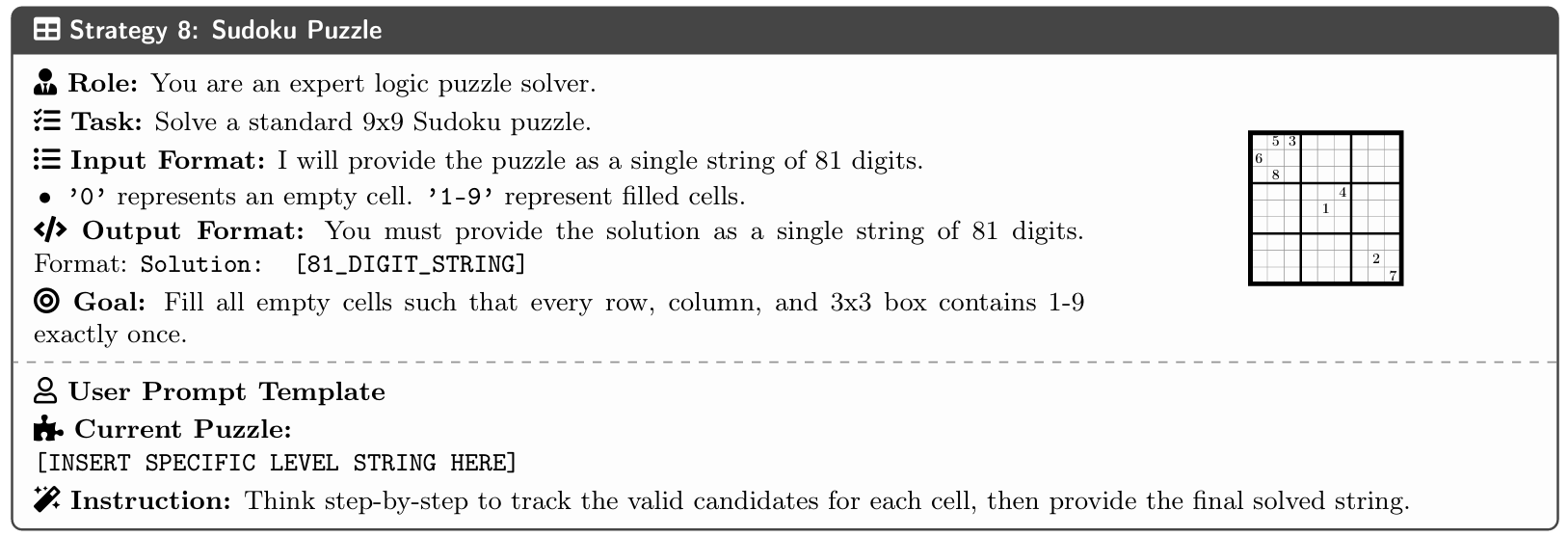}
\caption{Prompt template for Sudoku, using the 81-character string representation for puzzle input and output.}
\label{fig:prompt-sudoku}
\end{figure}

\begin{figure}
\centering
\includegraphics[width=0.95\columnwidth]{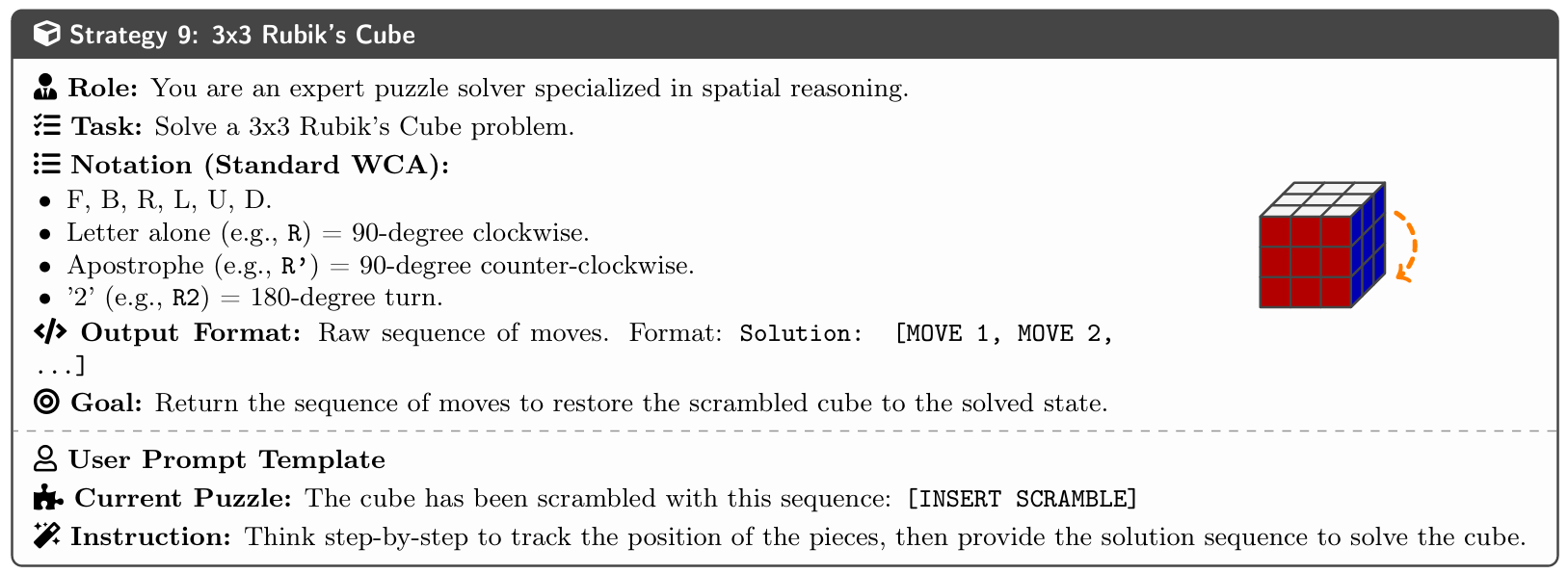}
\caption{Prompt template for the 3$\times$3 Rubik's Cube, using WCA notation for all legal face-turns.}
\label{fig:prompt-rubik}
\end{figure}

\subsection{Decoding and Budget Constraints}

Models are evaluated under default decoding settings recommended by their respective providers, with temperature fixed across runs. Maximum token budgets are set sufficiently high to avoid premature truncation at lower complexity levels. Importantly, no adaptive increase in token budget is applied at higher complexity levels; this allows us to observe whether models voluntarily reduce reasoning effort or collapse despite available capacity, consistent with prior observations on reasoning degradation.

\subsection{Validation and Correctness Checking}

Each task is paired with a deterministic, task-specific validator implemented as executable code. Model outputs are parsed into structured representations such as assignments, action sequences, or final states. A solution is marked as \texttt{Pass} if it satisfies all formal task constraints, \texttt{Fail} if a valid output is produced but violates constraints, and \texttt{Collapse} if the model fails to produce a parseable or complete solution within the allocated budget.

This validation protocol decouples correctness from linguistic plausibility and prevents models from receiving implicit credit for fluent but invalid reasoning traces.

\subsection{Evaluation Metrics}

For each model, task, and complexity level, we record binary correctness (Pass/Fail), collapse incidence, token usage, and execution time where available. In addition to aggregate success rates, we analyze qualitative changes in reasoning traces across complexity levels, including trace length, internal consistency, repetition, and premature termination. These metrics enable identification of reasoning collapse thresholds and characterization of failure modes beyond raw accuracy.

\subsection{Task Prompt and I/O Specification}

Each puzzle environment is paired with a fixed zero-shot prompt template and a deterministic I/O schema that jointly define the contract between the model and the validator. Prompts enumerate all task rules, admissible operations, and the required output format, while validators enforce strict structural and semantic correctness independently of the model’s reasoning trace. Table~\ref{tab:task_protocol} summarizes the prompt structure, required input parameters, and expected machine-parseable outputs for all tasks, following the standardized ordering used throughout the experimental pipeline. This unified interface ensures that performance differences arise from task complexity and model reasoning ability rather than from prompt variation or subjective interpretation.

\begin{table}
\centering
\small
\caption{Prompt structure, inputs, and required outputs for all benchmark tasks (ordered consistently with the experimental pipeline).}
\label{tab:task_protocol}
\renewcommand{\arraystretch}{1.15}
\resizebox{\columnwidth}{!}{
\begin{tabular}{lccc}
\hline
\textbf{Task} & \textbf{Structure} & \textbf{Inputs} & \textbf{Required Output} \\
\hline
Tower of Hanoi        & Recursion             & Disks $n$                  & Move list \\
1D Checkers Swap      & Planning              & Board, $N$ per color       & Move list \\
River Crossing        & Planning              & Pairs $n$, boat cap.\ $k$  & Transfers \\
Boolean SAT           & Constraints           & CNF clauses                & Boolean assignment \\
Cryptarithmetic       & Symbolic CSP          & Letter–digit equation      & Letter$\rightarrow$digit map \\
Graph Coloring        & Constraints           & Nodes, edges, colors       & Node$\rightarrow$color map \\
Water Jug             & Arithmetic/Planning   & Capacities, target goal    & Action sequence \\
Sudoku                & Constraint Grid       & $N \times N$ puzzle        & Completed grid \\
Rubik’s Cube (3×3)    & Search/Spatial        & Scramble sequence          & Move sequence \\
\hline
\end{tabular}
}
\end{table}

\section{Results Across the Complexity Regime}

This section analyzes model behavior across the full difficulty spectrum of the benchmark tasks. 
We evaluate each model's stability, reasoning fidelity, and robustness under escalating combinatorial complexity. 

\subsection{Task-Wise Model Performance}
\subsubsection{Tower of Hanoi}
Across the recursive planning hierarchy, all models solve Levels~1 and~2 reliably, indicating competence with short-horizon symbolic manipulation. 
Performance begins to diverge at Level~3, where weaker models fail to maintain the recursive invariant required for multi-step disk transfer planning. 
At Level~4, every model except \emph{Gemini~3 Pro} and \emph{DeepSeek~V3.2} collapses entirely, revealing sensitivity to depth-based reasoning chains and error-compounding dynamics.

\begin{figure*}[t]
\centering
\includegraphics[width=0.95\textwidth]{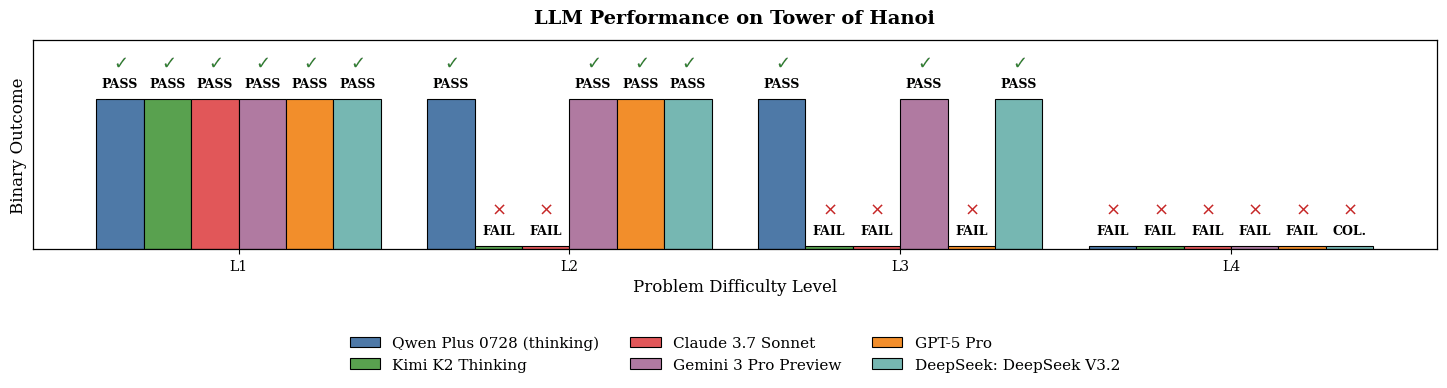}
\caption{LLM task performance across Tower of Hanoi complexity levels.}
\label{fig:hanoi}
\end{figure*}

\subsubsection{Water Jug}
All models demonstrate perfect performance through Level~3, confirming that simple Diophantine-style pouring operations remain tractable.
A sharp phase transition appears at Level~4: only \emph{Gemini~3 Pro} succeeds, while others fail or collapse due to inconsistent evaluation of reachable water states. 
Level~5 becomes universally challenging, with all models failing or collapsing. This confirms an abrupt brittleness to deeper multi-step search in continuous-state puzzles.

\begin{figure*}
\centering
\includegraphics[width=0.95\textwidth]{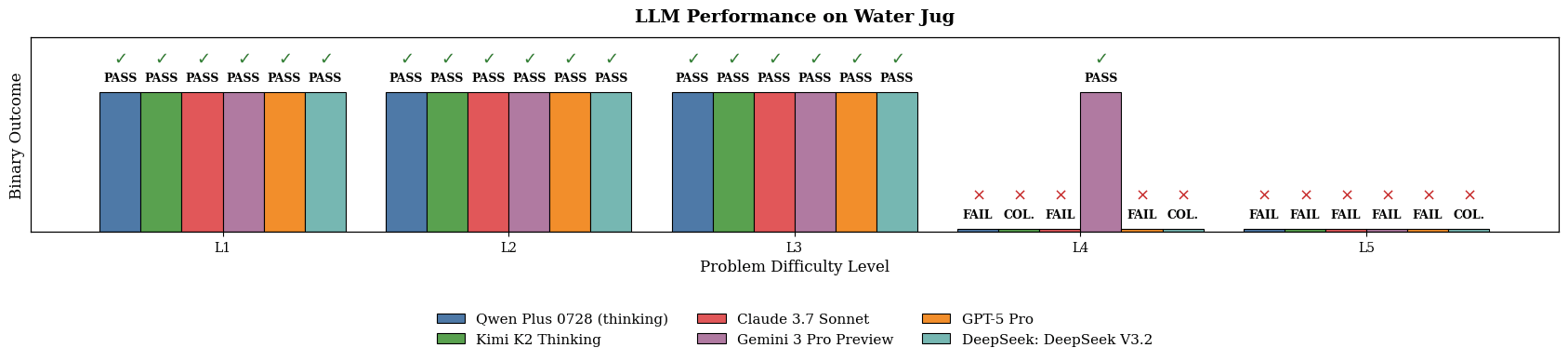}
\caption{LLM outcomes on the Water Jug task across increasing difficulty.}
\label{fig:waterjug}
\end{figure*}

\subsubsection{Boolean Satisfiability (SAT)}
Performance on SAT exposes pronounced differences in symbolic-logic robustness. 
While all models easily solve Levels~1--2, the majority fail or collapse from Level~3 onward. 
\emph{GPT-5 Pro} is a strong outlier, maintaining perfect pass rates up to Level~6, showcasing superior clause-level reasoning and search consistency. 
Other models exhibit early collapse signatures, suggesting instability in maintaining assignment consistency across clauses.

\begin{figure*}
\centering
\includegraphics[width=0.95\textwidth]{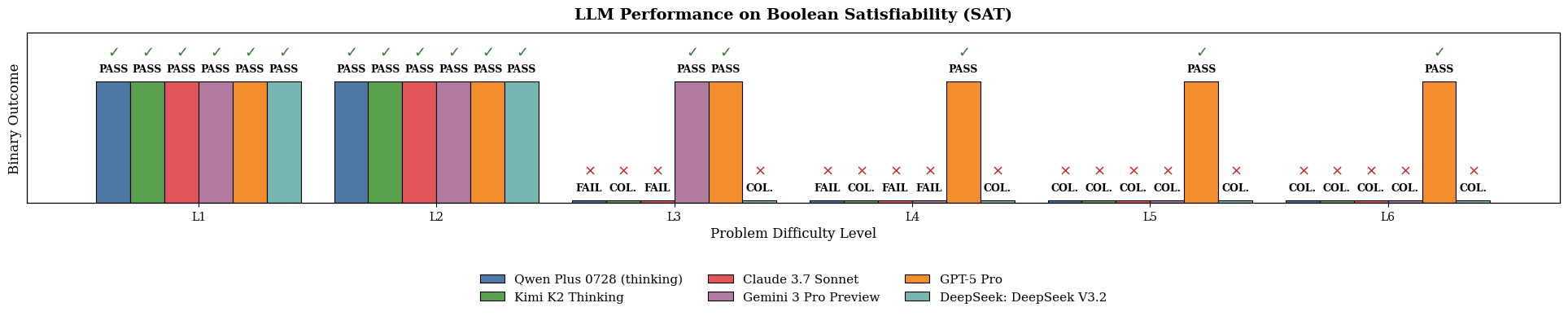}
\caption{Model robustness on Boolean SAT across clause-complexity levels.}
\label{fig:sat}
\end{figure*}

\subsubsection{Checker Jumping}
This task reveals the steepest degradation among all domains. 
While initial levels are solved by most models, catastrophic collapse emerges as early as Level~2 for several models and becomes universal by Levels~5--6. 
\emph{GPT-5 Pro} remains an exception, sustaining correct jump-sequence reasoning longer than others. 
The collapse behavior is directly correlated with models’ inability to handle combinatorial branching.

\begin{figure*}
\centering
\includegraphics[width=0.95\textwidth]{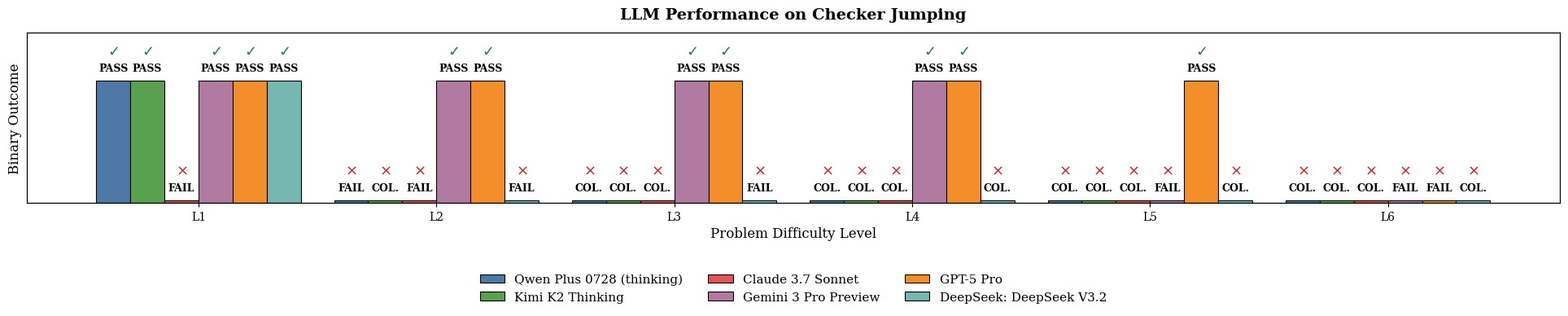}
\caption{Performance trajectories on the Checker Jumping puzzle.}
\label{fig:checker}
\end{figure*}

\subsubsection{Graph Coloring}
Models exhibit strong performance for Levels~1--3, implying competence with simple constraint-satisfaction reasoning. 
However, stability deteriorates sharply at Levels~4--5, where incorrect color propagation and early collapse patterns dominate. 
Only \emph{Gemini~3 Pro} and \emph{GPT-5 Pro} maintain high accuracy through mid-range complexity, indicating better global-constraint reasoning.

\begin{figure*}
\centering
\includegraphics[width=0.95\textwidth]{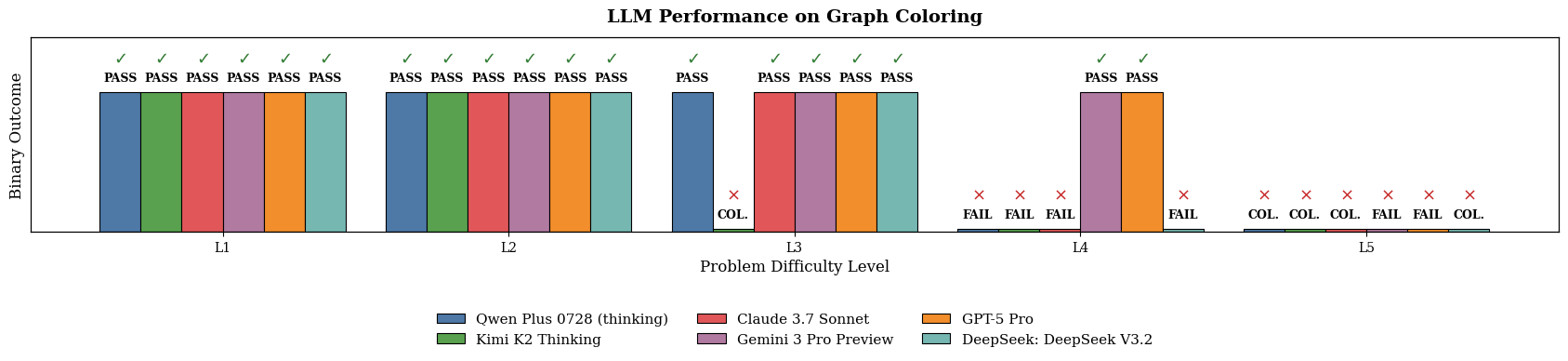}
\caption{Graph Coloring pass/fail/collapse profiles across complexity levels.}
\label{fig:graphcolor}
\end{figure*}

\subsubsection{River Crossing}
The classic relational-constraint puzzle exposes significant weaknesses in multi-agent state modeling. 
With the exception of \emph{Gemini~3 Pro} and \emph{GPT-5 Pro}, models frequently collapse from Levels~2--6 due to inconsistent evaluation of unsafe crossings. 
Only stronger models maintain stable reasoning throughout, indicating improved internal consistency for rule-based abductive inference.

\begin{figure*}
\centering
\includegraphics[width=0.95\textwidth]{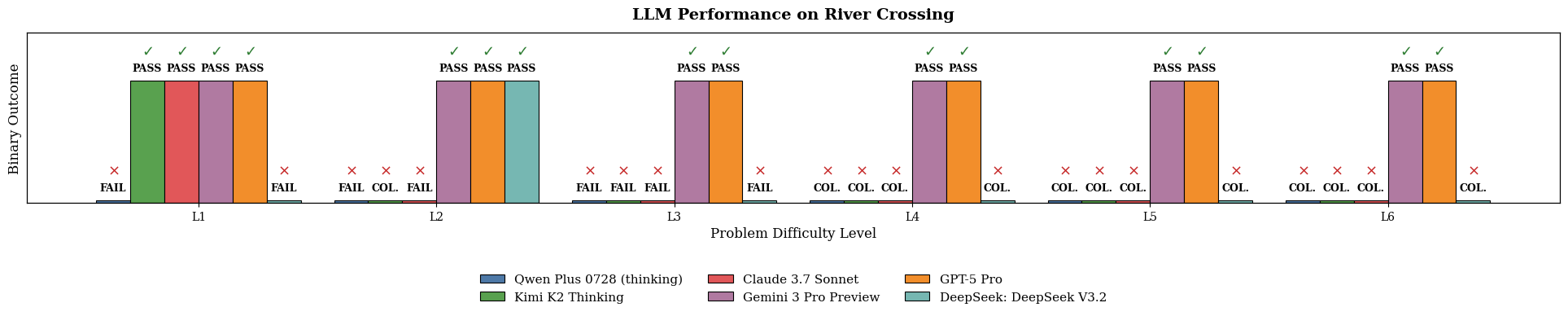}
\caption{LLM performance on River Crossing under escalating multi-agent constraints.}
\label{fig:river}
\end{figure*}

\subsubsection{Rubik's Cube}
All models solve Levels~1--3 with high stability, demonstrating competence with basic cube-state operations. 
Failures begin appearing at Level~4, although several models—including \emph{Gemini~3 Pro} and \emph{DeepSeek~V3.2}—continue to perform strongly. 
Remarkably, \emph{DeepSeek~V3.2} is the only model to solve Level~6, indicating exceptional resilience in high-dimensional permutation reasoning.

\begin{figure*}
\centering
\includegraphics[width=0.95\textwidth]{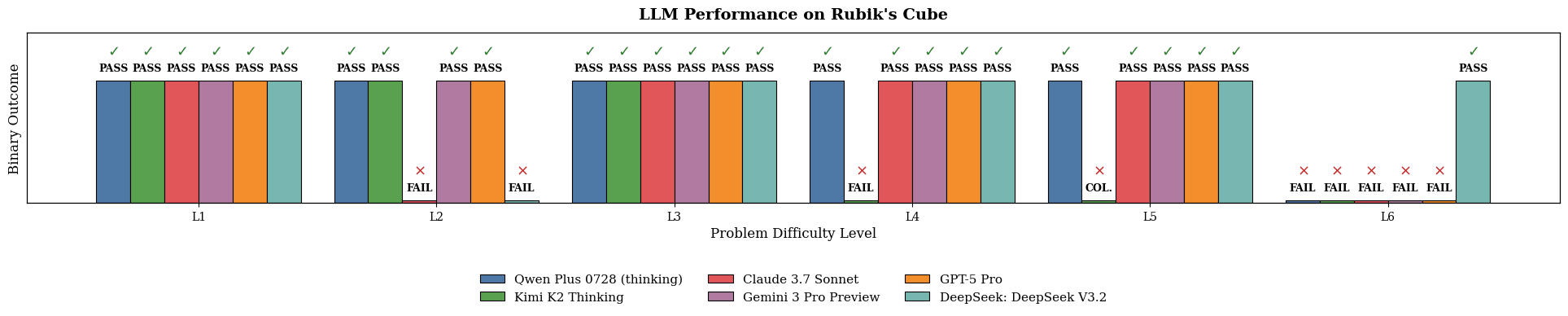}
\caption{Rubik's Cube performance across permutation-complexity levels.}
\label{fig:rubik}
\end{figure*}

\subsubsection{Sudoku}
This task exposes the most fragile reasoning patterns: nearly all models fail on every level except for isolated success by \emph{Kimi K2 Thinking} and \emph{GPT-5 Pro} at Level~1.
Higher levels result in universal collapse, indicating substantial difficulty in maintaining grid-wide consistency constraints under symbolic noise.

\begin{figure*}
\centering
\includegraphics[width=0.65\textwidth]{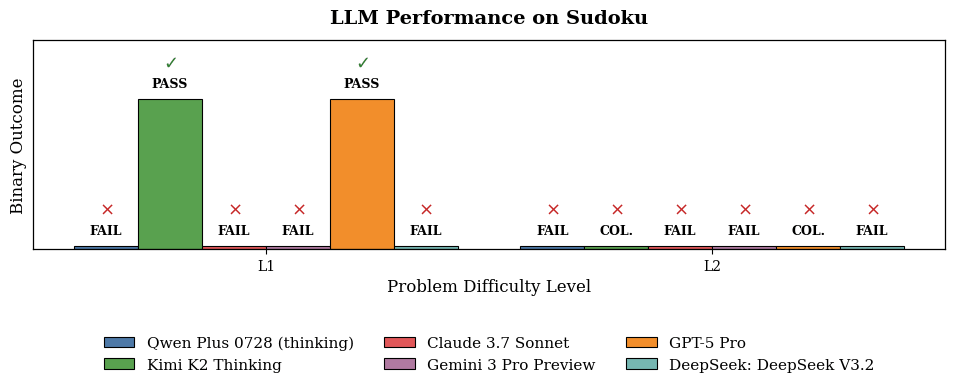}
\caption{Sudoku performance, illustrating extreme sensitivity to symbolic-consistency violations.}
\label{fig:sudoku}
\end{figure*}

\subsection{Task-Wise Difficulty Structure}

\begin{figure}
    \centering
    \includegraphics[width=0.9\columnwidth]{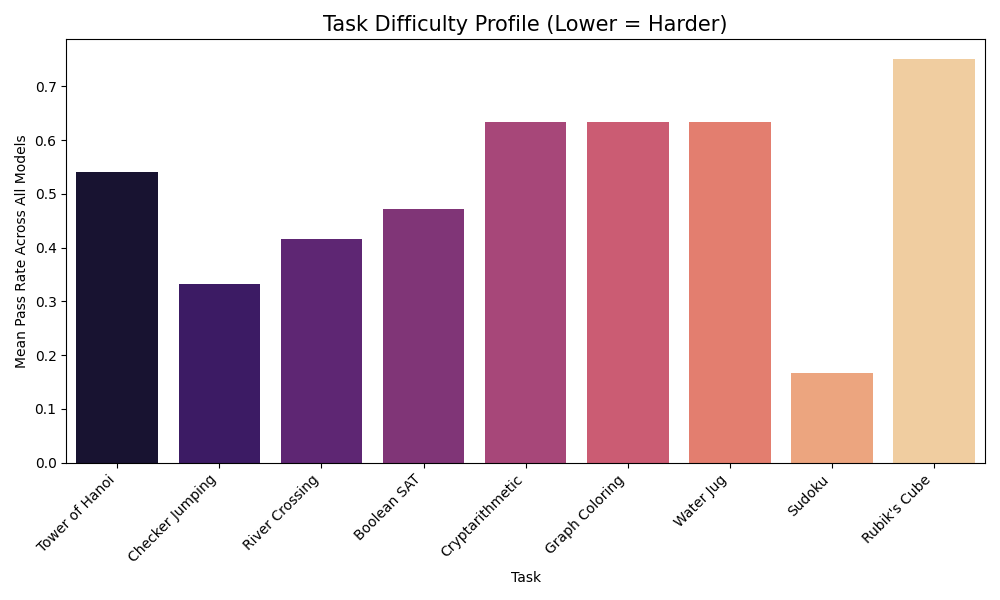}
    \caption{Mean pass rate across tasks (lower mean indicates higher difficulty).}
    \label{fig:task-difficulty}
\end{figure}

Figure~\ref{fig:task-difficulty} summarizes the average pass rate of all six models across the nine benchmark tasks. The difficulty profile exhibits a strongly non-uniform structure. Tasks with deterministic or recursive solution templates---such as \emph{Tower of Hanoi}, \emph{Graph Coloring}, \emph{Cryptarithmetic}, and \emph{Water Jug}---show moderate to high average pass rates, reflecting that models can leverage pattern imitation or short-horizon consistency. In contrast, tasks requiring multi-step symbolic tracking or strict global constraints---\emph{Checker Jumping}, \emph{River Crossing}, and especially \emph{Sudoku}---show substantially reduced success rates. \emph{Rubik’s Cube} stands as an outlier: its high average pass rate is driven by low-level instances, with aggregate performance masking sharp failure onset at higher levels.

\subsection{Global Scaling With Complexity}

\begin{figure}
    \centering
    \includegraphics[width=0.9\columnwidth]{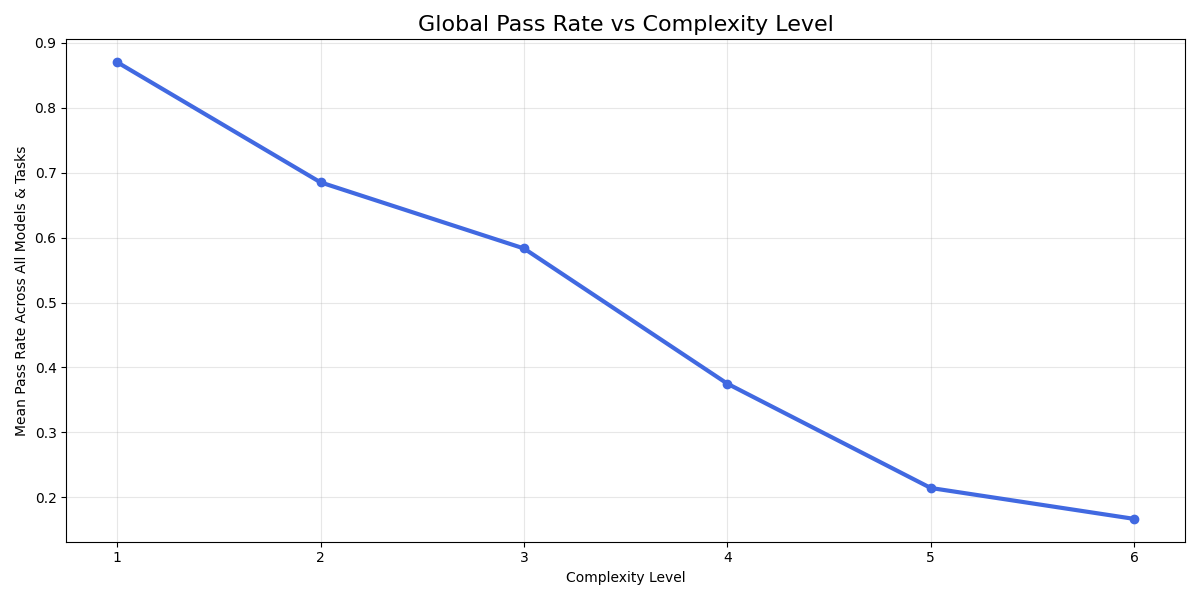}
    \caption{Global mean pass rate as a function of complexity level.}
    \label{fig:global-scaling}
\end{figure}

Figure~\ref{fig:global-scaling} highlights the global decline in mean pass rate as a function of problem complexity. Performance remains high at $L_{1}$, begins to diverge across models at $L_{2}$--$L_{3}$, and collapses sharply beyond $L_{4}$. The monotonic decay indicates that increased reasoning horizon, constraint density, and state-space branching systematically stress the underlying autoregressive architectures. Importantly, the steepest drop occurs between $L_{3}$ and $L_{4}$, aligning with the onset of multi-step consistency requirements where single-step local reasoning no longer suffices.

\subsection{Global Error Distribution Across Models}

\begin{figure}
    \centering
    \includegraphics[width=0.9\columnwidth]{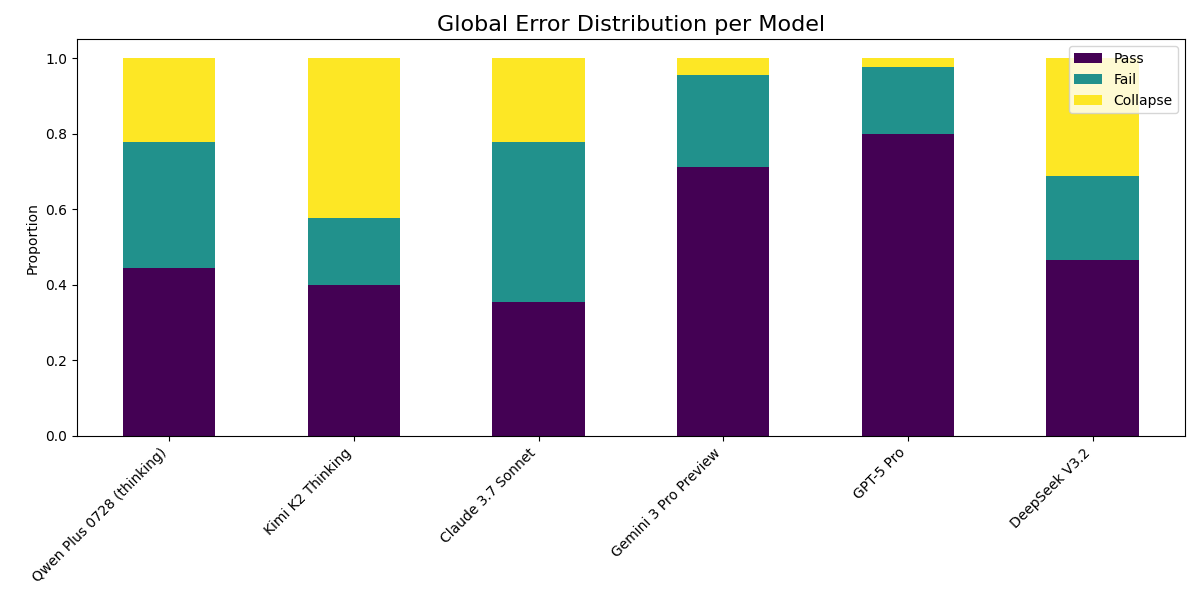}
    \caption{Global error distribution per model, showing proportions of correct outputs, incorrect outputs, and collapses across all tasks and complexity levels.}
    \label{fig:global-error-dist}
\end{figure}

The global error distribution presented in Fig.~\ref{fig:global-error-dist} reveals substantial heterogeneity in how different LLMs allocate their failures between \emph{incorrect} reasoning and outright \emph{collapse}. GPT\textendash5 Pro exhibits the strongest robustness profile, achieving approximately 80\% successful completions with almost no collapses, indicating stable reasoning trajectories even under higher task difficulty. Gemini~3 Pro Preview shows similarly resilient behavior, with over 70\% passes and relatively few collapses, suggesting effective long-horizon error control. In contrast, Qwen Plus 0728 and Claude~3.7 Sonnet demonstrate only moderate stability, splitting their errors between incorrect outputs and structural breakdowns—an indication of vulnerability to constraint-dense or sequentially dependent tasks. Kimi K2 Thinking displays the weakest reliability, with collapse events comprising nearly half of all outcomes, signaling brittle planning fidelity and poor state maintenance at increasing complexity. DeepSeek~V3.2 presents a mixed failure profile: although moderate in overall pass rate, it exhibits a disproportionately high collapse fraction, implying that while the model can occasionally produce correct solutions, its reasoning trajectory is unstable and prone to degeneration under multi-constraint or adversarial workloads. Collectively, these patterns show that stronger models tend to fail ``gracefully,'' producing structured but incorrect outputs, whereas weaker models exhibit catastrophic collapses, highlighting fundamental architectural differences in reasoning robustness.

\subsection{Global Collapse Thresholds Across Tasks and Models}

\begin{figure}
    \centering
    \includegraphics[width=0.9\columnwidth]{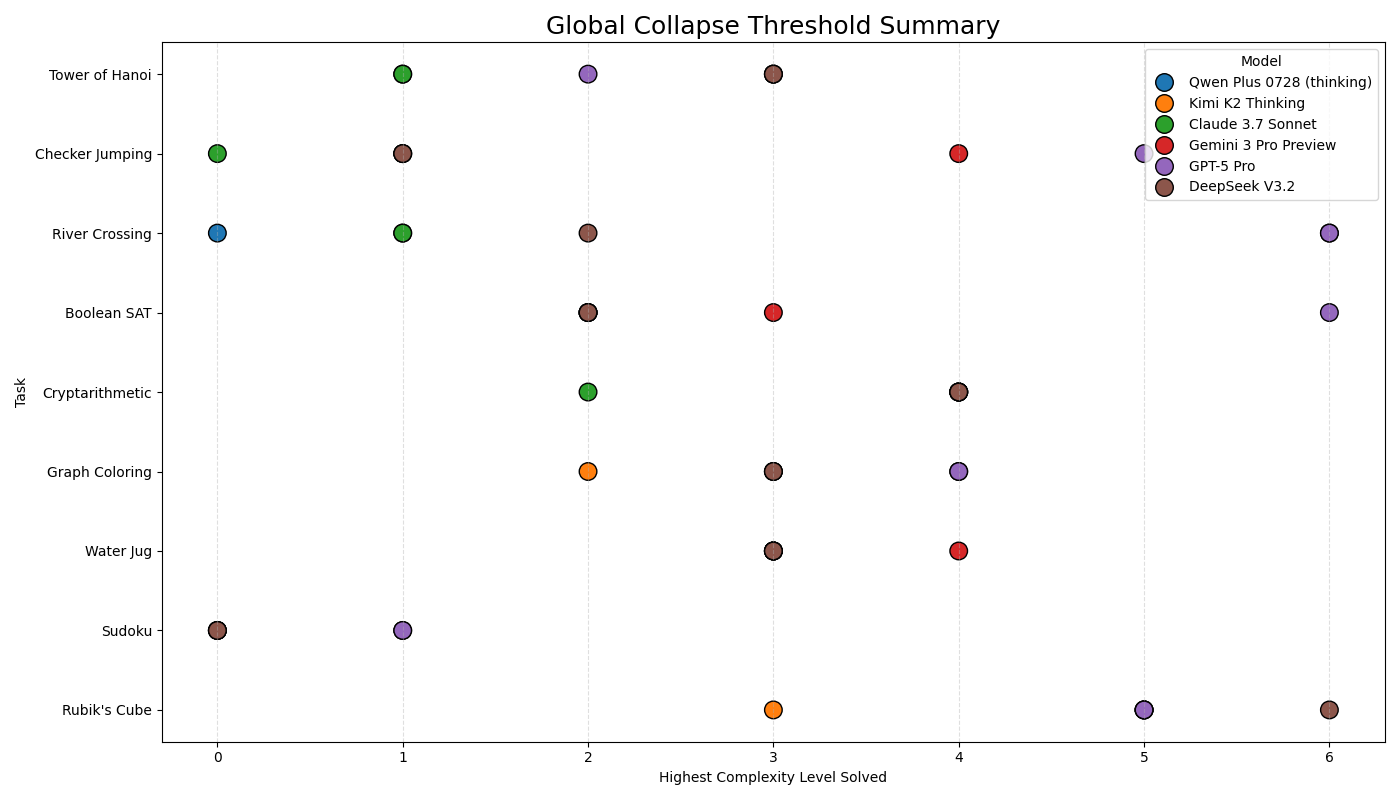}
    \caption{Global collapse threshold summary showing, for each model, the highest complexity level $L_k$ at which valid solutions are reliably produced across all nine reasoning tasks.}
    \label{fig:collapse-summary}
\end{figure}

Figure~\ref{fig:collapse-summary} synthesizes the collapse behaviour of all models across the full benchmark, revealing a consistent hierarchy of robustness and a sharp task-structure dependence. GPT--5~Pro demonstrates the broadest stability envelope, solving up to $L_5$--$L_6$ on most tasks before degradation, followed closely by Gemini~3~Pro~Preview, which maintains reliability through mid- to high-complexity regimes. In contrast, Qwen Plus 0728, Kimi K2 Thinking, and Claude 3.7 Sonnet exhibit substantially earlier breakdowns, typically collapsing between $L_1$ and $L_3$ on constraint-dense or multi-step planning tasks. DeepSeek~V3.2 displays a mixed profile: although capable of high performance on several tasks, it collapses early on others, indicating unstable long-horizon state maintenance.

Across models, tasks with strong recursive invariants (e.g., \emph{Tower of Hanoi}) support the deepest scaling, whereas constraint-dense tasks (e.g., \emph{Boolean SAT}, \emph{Graph Coloring}) and high-branching combinatorial tasks (e.g., \emph{Checker Jumping}, \emph{Rubik's Cube}) induce earlier collapse. This global trend underscores that collapse thresholds are governed jointly by model architecture and task structure: stronger models tend to fail \emph{later} and \emph{gracefully}, whereas weaker models collapse rapidly once the latent state or constraint load exceeds their representational capacity.

\subsection{Cross-Task Performance Landscape and Global Failure Behavior}

\begin{figure*}
    \centering
    \includegraphics[width=0.95\textwidth]{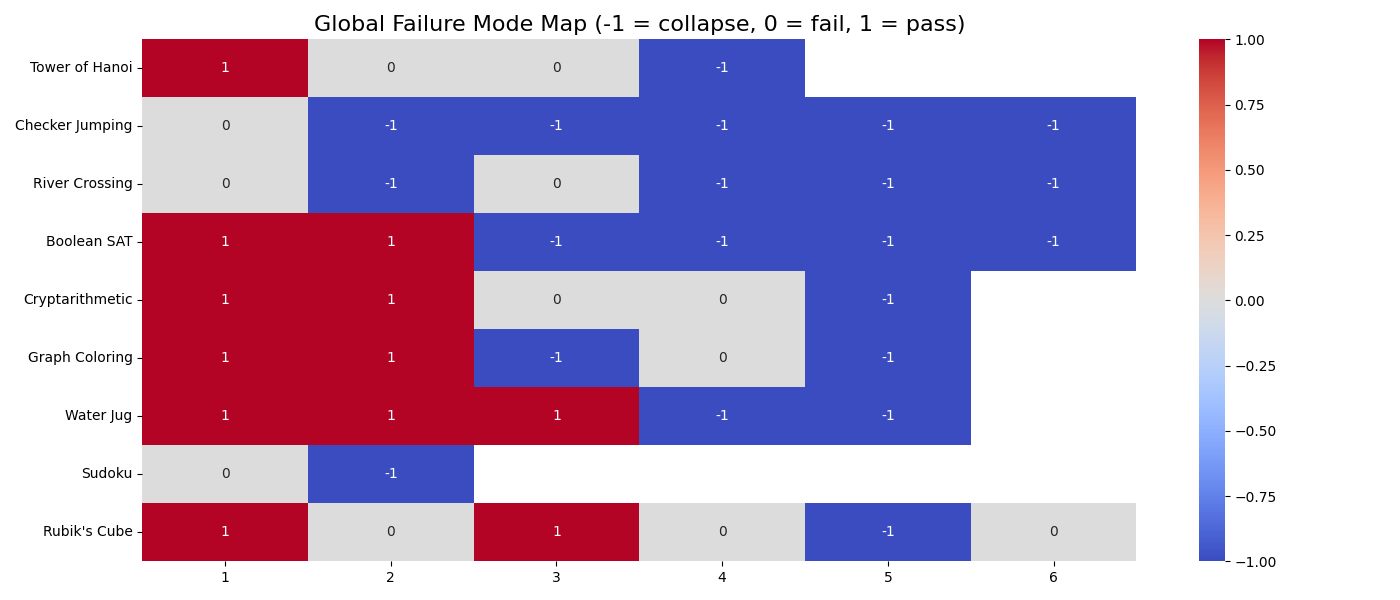}
    \caption{Global Failure Mode Map across tasks and difficulty levels. Outcomes are encoded as: $1$ = pass, $0$ = fail, and $-1$ = collapse.}
    \label{fig:global-failuremap}
\end{figure*}

\begin{figure*}
    \centering
    \includegraphics[width=0.9\textwidth]{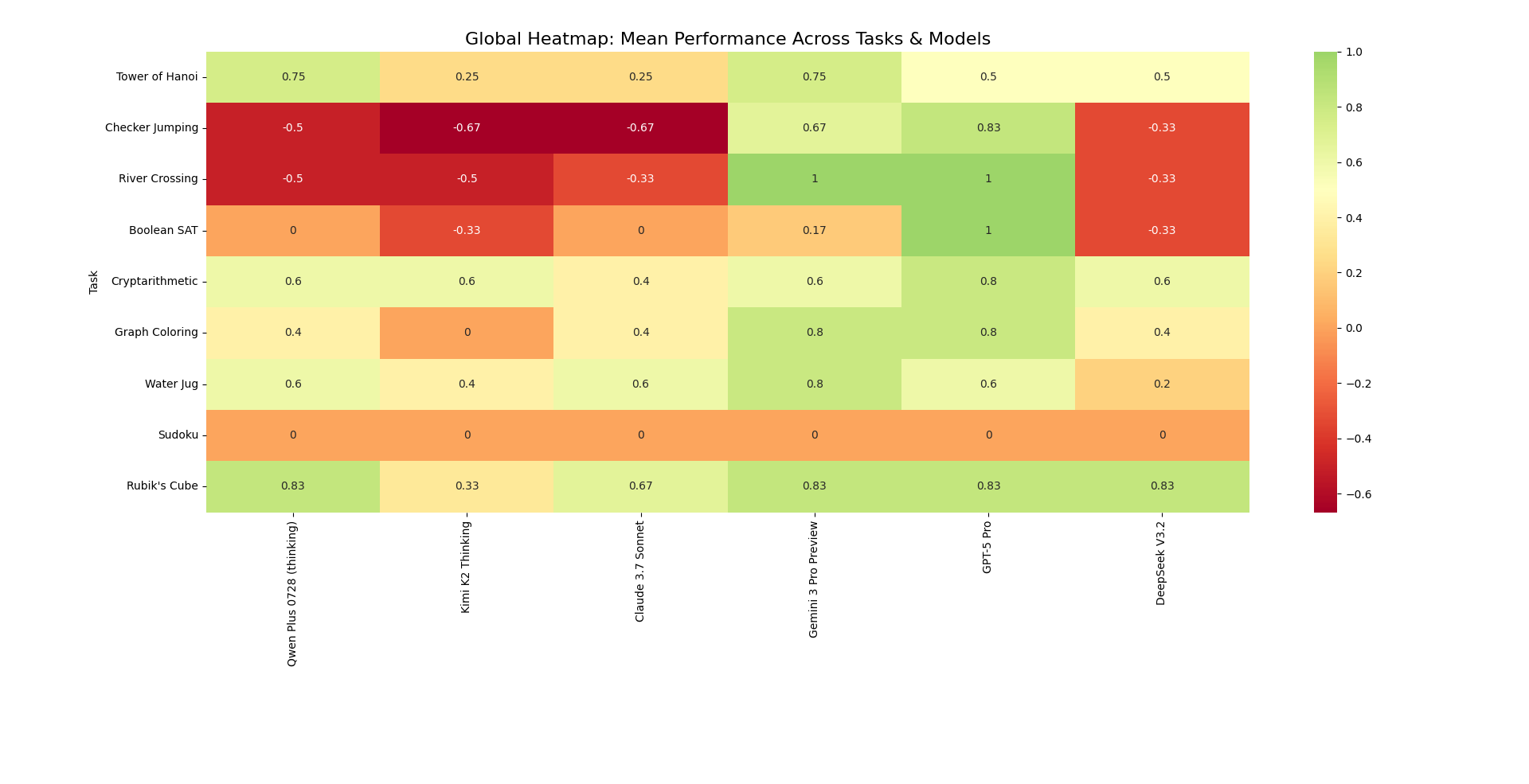}
    \caption{Global mean-performance heatmap aggregating pass rates across all models and tasks. Higher values indicate stronger and more consistent reasoning ability.}
    \label{fig:global-heatmap}
\end{figure*}

\begin{figure*}
    \centering
    \includegraphics[width=0.95\textwidth]{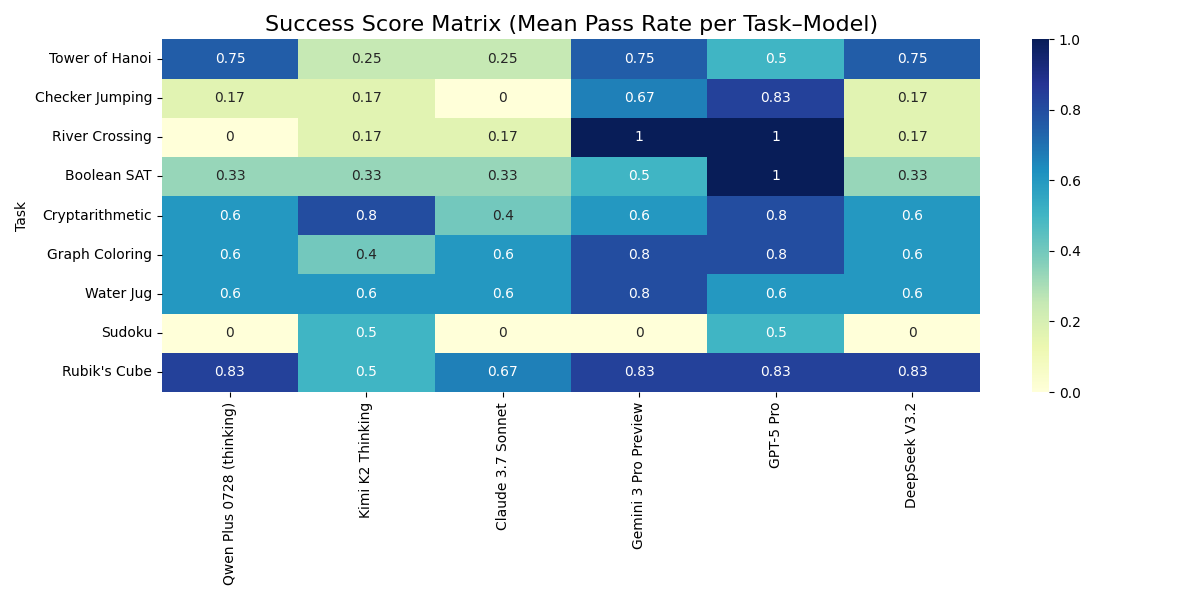}
    \caption{Success Score Matrix showing normalized mean pass rates per task--model pair.}
    \label{fig:success-matrix}
\end{figure*}

To characterize the broader behavioral tendencies of large reasoning models (LRMs) across heterogeneous puzzle families, we complement task-specific analyses with three global perspectives (Figs.~\ref{fig:global-failuremap}--\ref{fig:success-matrix}). These views jointly expose how failure types, aggregate competence, and cross-model variance emerge as task structure and difficulty intensify.

The \emph{Global Failure Mode Map} (Fig.~\ref{fig:global-failuremap}) illustrates that collapse events concentrate at higher difficulty tiers ($L_4$--$L_6$), especially for \emph{Checker Jumping}, \emph{River Crossing}, and \emph{Boolean SAT}. These patterns indicate that autoregressive inference becomes unstable under long-horizon reasoning or densely coupled constraints. In contrast, structurally regular tasks such as \emph{Tower of Hanoi}, \emph{Cryptarithmetic}, and \emph{Water Jug} show more predictable transitions from success to failure before collapse, suggesting that latent pattern reuse partially compensates for missing algorithmic generalization.

The \emph{Global Mean-Performance Heatmap} (Fig.~\ref{fig:global-heatmap}) reveals large competence disparities across models. GPT-5 Pro consistently forms the top-performing tier, followed by Gemini 3~Pro Preview, which shows strong reliability on both combinatorial and procedural domains. Mid-tier models such as Claude~3.7~Sonnet and Kimi K2 Thinking exhibit domain-specific strengths---performing well on symbolic arithmetic but degrading rapidly on relational or multi-agent puzzles. Qwen Plus (thinking) and DeepSeek~V3.2 show the greatest variability: while they occasionally excel (e.g., \emph{Rubik's Cube}), they also display early collapse on tasks with dense constraints.

The \emph{Success Score Matrix} (Fig.~\ref{fig:success-matrix}) quantifies these disparities more precisely. GPT-5 Pro demonstrates near-uniform competence across tasks, while Gemini 3~Pro Preview shows similarly strong but slightly more selective performance. The remaining models exhibit marked specialization: for example, Kimi K2 Thinking performs competitively on \emph{Cryptarithmetic} yet collapses early on \emph{Checker Jumping} and \emph{SAT}. DeepSeek~V3.2 displays a highly heterogeneous profile, with strong success on \emph{Rubik’s Cube} but frequent collapse on high-constraint tasks. Overall, these patterns indicate that reasoning ability in LRMs is not uniformly general but closely tied to structural task properties and model-specific inductive biases.

Taken together, these global analyses show that LRMs do not fail randomly; rather, they exhibit structured, domain-dependent breakdown patterns. Models with stronger implicit search priors (GPT-5 Pro, Gemini 3~Pro Preview) remain stable deeper into the complexity spectrum, whereas others rely more heavily on recognizable structural templates. This heterogeneity underscores the necessity of multi-task and multi-signal evaluation when assessing claims of ``general'' reasoning competence.

\subsection{Aggregate Model Competence}

\begin{figure}
    \centering
    \includegraphics[width=0.9\columnwidth]{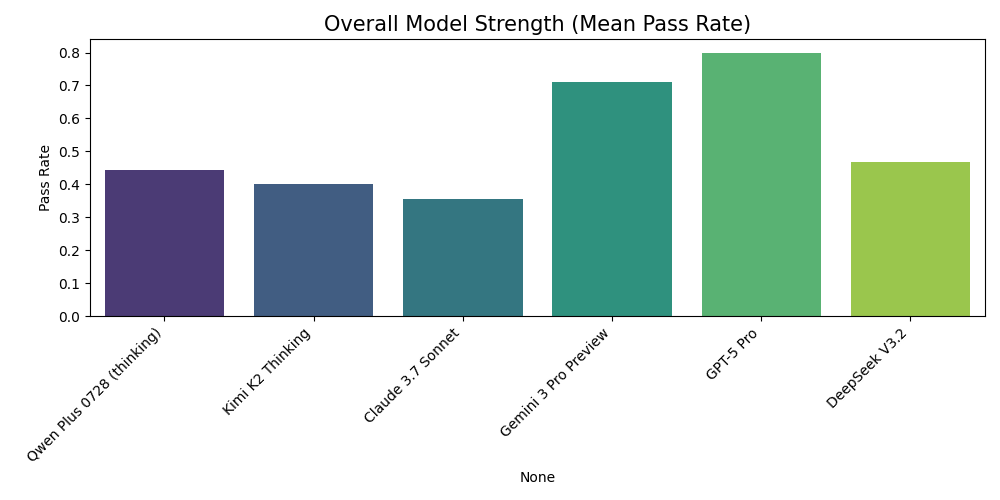}
    \caption{Aggregate model strength, measured as overall mean pass rate across all tasks and complexity levels.}
    \label{fig:model-strength}
\end{figure}

The overall mean pass rate per model, shown in Figure~\ref{fig:model-strength}, reveals a wide separation in effective reasoning robustness. GPT-5 Pro leads by a significant margin, followed by Gemini 3 Pro. These models exhibit extended resistance to collapse and maintain high pass rates even under increased constraint density. Qwen Plus and DeepSeek achieve moderate robustness, whereas Claude 3.7 Sonnet and Kimi K2 Thinking show early breakdowns under compositional or sequential stress. This separation aligns with architectural differences: stronger search-like inductive biases and longer-context stability correlate with improved reasoning longevity.

\section{Discussion}

Large reasoning models (LRMs) exhibit compelling yet fragile reasoning behavior under controlled increases in problem complexity. Across the nine heterogeneous tasks evaluated in this study, a consistent qualitative trajectory emerges: strong and reliable performance at low complexity ($L_1$--$L_2$), unstable and task-dependent behavior at intermediate complexity ($L_3$--$L_4$), and sharp degradation beyond task-specific thresholds at high complexity ($L_5+$). This section synthesizes these findings, clarifies the dominant failure mechanisms revealed by deterministic validation, and discusses structural drivers, cross-model regularities, implications, and limitations.

\subsection{Collapse Phenomena Under Controlled Complexity}

A key observation is that performance typically does not degrade smoothly with increasing complexity. Instead, for most tasks, models exhibit phase-transition-like behavior: a relatively stable regime at low complexity followed by a rapid drop in success as the complexity parameter crosses a task-specific threshold. In the low-complexity regime ($L_1$--$L_2$), models frequently generate coherent traces that follow recognizable solution templates and pass deterministic validation. In the intermediate regime ($L_3$--$L_4$), partial failures become common: models may correctly state rules but violate them in execution, drift from the implied intermediate state, or terminate prematurely. At the highest levels ($L_5+$), outputs are often invalid, incomplete, or structurally corrupted despite large token budgets.

Interpreting this behavior requires caution: these results do not imply that LRMs ``stop reasoning'' at high complexity, but rather that their reasoning process becomes unreliable under increased cognitive load. Under our formal definition of reasoning (state-consistent and constraint-valid transformations), collapse corresponds to a sharp increase in intermediate invalidity and a steep drop in end-to-end success. This suggests that current LRMs operate with a finite effective capacity for maintaining structured state, enforcing interacting constraints, and planning over long horizons, such that performance can deteriorate abruptly once that capacity is exceeded.

\subsection{Mechanisms of Failure Revealed by Deterministic Validation}

Deterministic validators allow us to move beyond surface plausibility and attribute failures to explicit violations of task mechanics. Across tasks, four recurring mechanisms explain most validator failures. Table~\ref{tab:failure_mechanisms} summarizes them; below we provide operational clarifications and task-level interpretation.

\paragraph{(1) Long-horizon state degradation.}
In long planning tasks, failures frequently arise after many steps that appear locally plausible. Operationally, this corresponds to an increasing rate of illegal transitions as trajectory length grows: moves reference nonexistent state elements (e.g., moving a disk not on top), undo progress through cyclic sequences, or drift into unreachable configurations. This pattern is most evident in long-horizon domains with strict step validity (e.g., Tower of Hanoi, Rubik’s Cube). Importantly, this behavior is \emph{consistent with} limited effective state retention under autoregressive decoding, where earlier commitments must be implicitly maintained across extended generations.

\paragraph{(2) Constraint propagation failure.}
In constraint-dense tasks, models often articulate the constraints correctly but fail to jointly enforce them. Operationally, validators detect globally inconsistent assignments: repeated digits in Sudoku, violated clauses in SAT, or adjacency conflicts in graph coloring. These failures indicate that constraint satisfaction does not emerge simply from verbalizing constraints. Instead, increasing density and coupling of constraints appears to overwhelm the model’s ability to preserve global consistency across many interacting commitments.

\paragraph{(3) Unstable feasibility planning under safety constraints.}
For tasks such as River Crossing and Checker Jumping, correctness requires maintaining feasibility at every intermediate step. Operationally, failures occur when a model proposes an action that violates safety predicates (unsafe bank configurations) or performs locally legal moves that lead to dead-ends or incomplete plans. This suggests that under branching action spaces with global feasibility invariants, LRMs often rely on shallow heuristics that do not reliably anticipate downstream feasibility.

\paragraph{(4) Non-scaling of reasoning with token length.}
A striking pattern is that longer traces do not reliably improve correctness at higher complexity. Operationally, at $L_3$--$L_5$, we frequently observe (i) longer traces that still end in invalid solutions, and (ii) degraded traces that become repetitive, truncated, or internally inconsistent. This indicates that the limiting factor is not simply token budget; rather, additional generation does not guarantee better state tracking or constraint enforcement once the underlying difficulty exceeds the model’s effective capacity.

\begin{table*}
\centering
\small
\caption{Core mechanisms of reasoning failure in LRMs, identified through validator-grounded analysis across tasks.}
\label{tab:failure_mechanisms}
\renewcommand{\arraystretch}{1.25}
\begin{tabular}{p{3.2cm} p{10.4cm}}
\hline
\textbf{Failure Mechanism} & \textbf{Description and Typical Manifestation} \\
\hline

\textbf{Long-Horizon State Degradation} 
& Breakdown of state tracking over extended sequences: illegal transitions, cyclic regressions, or references to nonexistent state elements (e.g., Tower of Hanoi, Rubik’s Cube). This pattern is consistent with limited effective state retention under autoregressive generation. \\

\textbf{Constraint Propagation Failure} 
& Inability to maintain global consistency across interacting constraints: models describe rules correctly yet output contradictory or invalid assignments (e.g., Sudoku duplicates, violated SAT clauses, graph coloring conflicts). \\

\textbf{Unstable Feasibility Planning} 
& Failure to preserve feasibility under sequential or safety constraints: unsafe intermediate states, deadlocks, or incomplete plans (e.g., River Crossing, Checker Jumping), indicating difficulty coordinating branching actions with global invariants. \\

\textbf{Non-Scaling of Reasoning with Token Length} 
& Increased verbosity does not yield improved correctness: at higher complexity, traces may become longer but remain invalid, or degrade into repetition/truncation despite large budgets, suggesting limits beyond token availability. \\
\hline
\end{tabular}
\end{table*}

Taken together, these mechanisms indicate that many LRM failures are structural: they arise from breakdowns in state maintenance, global constraint enforcement, and feasibility reasoning under complexity—rather than from superficial formatting issues or missing rule knowledge.

\subsection{Structural Drivers of Complexity Scaling}

Collapse thresholds differ by task family, but they correlate strongly with the structural properties that determine cognitive load. Table~\ref{tab:structural_drivers} summarizes the dominant regimes.

\paragraph{Recursive structure and template-like solutions.}
Tasks with strong self-similar templates (e.g., Tower of Hanoi) tend to support deeper scaling. One plausible interpretation is that LRMs can imitate recursive templates more reliably than they can perform explicit search: when the solution exhibits a repetitive, rule-driven structure, the model can extend patterns longer before state drift becomes catastrophic. Nevertheless, recursion does not guarantee indefinite scaling; as horizons grow, errors still accumulate and eventually trigger failure.

\paragraph{Sequential deterministic dynamics.}
Tasks with low branching and deterministic arithmetic transitions (e.g., Water Jug) often scale moderately. Here, local correctness is easier to maintain, but long-range dependencies still accumulate. Once plans require many interdependent operations, small arithmetic or state errors propagate and invalidate the final outcome.

\paragraph{Constraint-dense compositionality.}
Constraint-heavy tasks (e.g., Sudoku, SAT, graph coloring) commonly induce earlier collapse. These domains require simultaneous satisfaction of many interacting constraints, and errors may be subtle yet globally fatal. This indicates that the ability to \emph{state} constraints does not translate into the ability to \emph{enforce} them under high coupling.

\paragraph{High branching and high-dimensional state spaces.}
Tasks characterized by large branching factors, safety constraints, or entangled spatial state (e.g., Rubik’s Cube, River Crossing) often exhibit the earliest collapse. Here, error compounding is accelerated by the size of the action space and the need to maintain consistency across many coupled degrees of freedom.

\begin{table*}
\centering
\small
\caption{Structural factors governing reasoning scalability and collapse thresholds across tasks.}
\label{tab:structural_drivers}
\renewcommand{\arraystretch}{1.25}
\begin{tabular}{p{3.4cm} p{10.2cm}}
\hline
\textbf{Structural Regime} & \textbf{Impact on Scaling and Collapse Behavior} \\
\hline

\textbf{Recursive Structure}  
\newline (e.g., Tower of Hanoi) 
& Often supports deeper scaling relative to other families: self-similar, rule-driven templates are easier to extend before catastrophic drift, though horizons still eventually exceed reliable state tracking. \\

\textbf{Sequential Deterministic Dynamics}  
\newline (e.g., Water Jug) 
& Enables moderate scaling when branching is low and transitions are predictable, but failures emerge as long-range dependencies accumulate and arithmetic/state drift compounds. \\

\textbf{Constraint-Dense Compositionality}  
\newline (e.g., Sudoku, SAT, Graph Coloring) 
& Tends to collapse earlier: models can verbalize constraints yet struggle to jointly enforce global consistency under increasing density and coupling, leading to subtle but fatal violations. \\

\textbf{High-Branching / High-Dimensional State Spaces}  
\newline (e.g., Rubik’s Cube, River Crossing) 
& Often collapses earliest: large action spaces, safety predicates, and entangled state representations amplify error propagation and overwhelm reliable state maintenance. \\
\hline
\end{tabular}
\end{table*}

These regimes highlight that ``reasoning'' is not a monolithic capability: scalability depends strongly on the structural demands of the domain.

\subsection{Cross-Model Regularities}

Despite differences in model scale, training data, and ``thinking'' modes, we observe convergent behavioral regularities across models. Table~\ref{tab:cross_model_regularities} summarizes the dominant patterns.

\begin{table*}
\centering
\small
\caption{Cross-model regularities observed across evaluated LRMs. Convergent patterns across tasks and regimes suggest shared structural constraints rather than model-specific idiosyncrasies.}
\label{tab:cross_model_regularities}
\renewcommand{\arraystretch}{1.18}
\begin{tabular}{p{3.4cm} p{10.4cm}}
\hline
\textbf{Regularity} & \textbf{Description} \\
\hline

\textbf{Uniform Three-Phase Trajectory} 
& Most models exhibit reliable success at $L_1$--$L_2$, unstable reasoning at $L_3$--$L_4$, and sharp degradation at $L_5+$, consistent with shared limits in state tracking and constraint enforcement. \\[4pt]

\textbf{Robustness on Template-Based Tasks} 
& Tasks with strong templates (e.g., Tower of Hanoi; to some extent Checker Jumping) show more stable behavior, consistent with pattern imitation outperforming open-ended search under complexity. \\[4pt]

\textbf{Early Collapse on Dense or High-Dimensional Tasks}
& Constraint-dense and high-dimensional domains (e.g., Sudoku/SAT/Graph Coloring; Rubik's Cube/River Crossing) induce similar collapse regimes across models, suggesting that structure dominates over model-specific differences. \\[4pt]

\textbf{Token Length Not Predictive of Success}
& Additional chain-of-thought length does not reliably improve correctness at higher complexity; verbosity varies, but validity does not track it consistently once tasks exceed effective capacity. \\[4pt]

\textbf{Convergent Failure Modes}
& Similar structural errors recur across architectures: contradictory assignments (SAT), repeated digits (Sudoku), illegal transitions or unsafe states (planning tasks), indicating shared bottlenecks. \\
\hline
\end{tabular}
\end{table*}

Overall, these regularities suggest that scaling model size or encouraging longer traces alone is insufficient to guarantee robustness under controlled complexity increases.

\subsection{Implications for Benchmarking and Model Design}

Our findings have several implications.

\paragraph{Benchmarking should be complexity-aware and validator-grounded.}
Static benchmark accuracy can mask failure dynamics. Controlled complexity scaling exposes the transition from reliable behavior to instability and collapse, while deterministic validation prevents fluent but invalid reasoning from being misclassified as success. Future evaluations should report performance as a function of complexity (e.g., accuracy--complexity curves) and include intermediate validity measures when trajectories are required.

\paragraph{Reasoning traces are not reliable evidence of correctness.}
At higher complexity, models frequently produce plausible narratives that violate task constraints. This underscores that interpretability via chain-of-thought is limited unless paired with external validation, and that ``more tokens'' does not guarantee stronger reasoning.

\paragraph{Robust reasoning likely requires explicit structure beyond pure generation.}
The dominant failure mechanisms point toward missing capabilities: reliable state representation, constraint propagation, and search/planning under branching. Tool use (e.g., external solvers), structured decoding, explicit memory, or neuro-symbolic hybrids may be necessary to maintain correctness under high complexity. Our framework provides a testbed for evaluating whether such augmentations shift collapse thresholds and reduce intermediate invalidity.

\subsection{Limitations and Future Directions}

This study has several limitations that should be considered when interpreting results.

\paragraph{Prompt sensitivity and evaluation protocol.}
We primarily evaluate models under a standardized, zero-shot setting to probe intrinsic reasoning under controlled complexity. However, results may vary under different prompting strategies (few-shot exemplars, decomposition prompts, self-consistency, or tool-augmented prompting). A systematic prompt-sensitivity analysis would clarify the extent to which collapse thresholds are robust to prompt design.

\paragraph{Discrete complexity levels.}
Complexity is discretized into task-specific levels ($L_1$--$L_5+$). While sufficient to expose sharp transitions, finer-grained (or continuous) difficulty schedules could reveal more nuanced dynamics, such as gradual degradation within intermediate regimes or multiple breakpoints.

\paragraph{Task coverage and ecological validity.}
Although the nine puzzle families span major forms of discrete reasoning (planning, constraint satisfaction, combinatorial search), they remain synthetic and may not fully represent real-world reasoning, where external knowledge, noisy inputs, and underspecified goals are common. Extending controlled complexity paradigms to more naturalistic settings remains an important direction.

\paragraph{Instance generation and potential leakage.}
Even with controlled puzzles, some instances (or close variants) may exist in pretraining corpora, especially for canonical problems. While our emphasis is on complexity scaling and deterministic validation rather than absolute accuracy on specific instances, future work should further reduce leakage risk via procedurally generated instances with documented seeds and distributions.

\paragraph{Model coverage.}
Our evaluated model set spans multiple open and proprietary LRMs, but does not exhaust the space of architectures and augmentation strategies (e.g., retrieval-augmented generation, planner-tool loops, or multimodal agents). Testing these systems under the same validators is necessary to determine whether augmentation shifts collapse thresholds or merely changes failure modes.

\paragraph{Future directions.}
Future work should (i) introduce tool-augmented and memory-augmented baselines to test whether explicit structure reduces collapse; (ii) expand task families to include additional modalities (e.g., inductive logic, geometry, probabilistic reasoning); (iii) develop trace-level metrics (e.g., intermediate validity rate, contradiction rate) as standard reporting; and (iv) explore formal breakpoint estimation methods to quantify collapse thresholds with uncertainty (e.g., segmented regression over accuracy--complexity curves).
\section{Conclusion}

This study delivers a systematic, controlled evaluation of large reasoning models (LRMs) across nine structurally diverse tasks and graded complexity regimes, revealing a consistent and architecture-agnostic pattern: strong apparent reasoning at low complexity, unstable performance at intermediate levels, and abrupt, task-dependent collapse once structural demands surpass model capacity. Through zero-shot prompting paired with deterministic validators, we isolate genuine reasoning ability from surface-level fluency and show that failures arise from fundamental limitations in long-horizon state tracking, global constraint propagation, and multi-step planning. Collapse thresholds correlate strongly with task structure—models scale furthest on recursive or low-branching problems, degrade earlier on constraint-dense tasks, and fail most rapidly on high-dimensional or safety-constrained environments—highlighting that LRMs lack transferable reasoning abstractions and rely instead on brittle, domain-specific heuristics. Cross-model regularities further suggest shared computational bottlenecks rather than idiosyncratic weaknesses. Overall, these findings underscore that increasing token budgets, reasoning verbosity, or model size alone does not yield scalable algorithmic competence; progress toward truly robust high-complexity reasoning will likely require architectural innovations that support persistent state, explicit constraint enforcement, symbolic abstraction, or hybrid neuro-symbolic integration, all of which are essential for deploying LRMs in demanding scientific, planning, and decision-making settings.

\nocite{*}
\bibliographystyle{unsrtnat}
\bibliography{references}

\end{document}